\documentclass{article}

\usepackage[preprint]{neurips_2025}

\usepackage[utf8]{inputenc} 
\usepackage[T1]{fontenc}    
\usepackage{hyperref}       
\usepackage{url}            
\usepackage{booktabs}       
\usepackage{amsfonts}       
\usepackage{nicefrac}       
\usepackage{microtype}      
\usepackage{kotex} 

\usepackage{comment} 
\usepackage{xcolor} 
\usepackage{graphicx}
\usepackage{color} 
\usepackage{pifont} 
\usepackage{xspace} 
\usepackage{wrapfig,lipsum} 
\usepackage{colortbl} 
\usepackage{multirow}
\usepackage{amsthm} 
\usepackage{amsmath} 
\usepackage{float} 
\usepackage{comment} 
\usepackage{adjustbox} 
\usepackage{wrapfig,lipsum,booktabs} 
\usepackage{caption} 
\usepackage[most]{tcolorbox}
\usepackage{xr}
\usepackage{subcaption} 
\usepackage{dblfloatfix} 

\newtheorem{proposition}{Proposition}

\newcommand{\cmarkg}{\textcolor{green}{\ding{51}}\xspace}%
\newcommand{\xmarkg}{\textcolor{red}{\ding{55}}\xspace}%

\newcommand{\rednum}[1]{\cellcolor[HTML]{FFB2B2}{\textcolor{black}{#1}}}

\newcommand{\orangenum}[1]{\cellcolor[HTML]{FFD9B2}{\textcolor{black}{#1}}}

\newcommand{\lightorangenum}[1]{\cellcolor[HTML]{FFFFB2}{\textcolor{black}{#1}}}

\newcommand{\tabincell}[2]{\begin{tabular}{@{}#1@{}}#2\end{tabular}} 

\newcommand{\etal}{\textit{et al.}} 

\theoremstyle{definition}

\theoremstyle{remark}
\newtheorem{remark}{Remark}
\theoremstyle{plain}

\theoremstyle{plain}

\title{Geometrical Properties of Text Token Embeddings for Strong Semantic Binding in Text-to-Image Generation}

\author{Hoigi Seo$^{1*}$ \quad Junseo Bang$^{1*}$ \quad Haechang Lee$^{1*}$ \\ \textbf{Joohoon Lee$^{2}$ \quad Byung Hyun Lee$^{1}$ \quad Se Young Chun$^{1,2,3,\dagger}$}\\
$^1$Dept. of ECE, \ $^2$IPAI \& $^3$INMC, \ Seoul National University, \ Republic of Korea,\\
{\tt\small \{seohoiki3215,qkdwnstj10,harrylee,joohoonl,ldlqudgus756,sychun\}@snu.ac.kr} 
}

\begin{document}

\maketitle
\begin{abstract}
Text-to-image (T2I) models often suffer from text-image misalignment in complex scenes involving multiple objects and attributes. Semantic binding has attempted to associate the generated attributes and objects with their corresponding noun phrases (NPs) by text or latent optimizations with the modulation of cross-attention (CA) maps; yet, the factors that influence semantic binding remain underexplored. Here, we investigate the geometrical properties of text token embeddings and their CA maps. We found that the geometrical properties of token embeddings, specifically angular distances and norms, are crucial factors in the differentiation of the CA map. These theoretical findings led to our proposed training-free text-embedding-aware T2I framework, dubbed \textbf{TokeBi}, for strong semantic binding. TokeBi consists of Causality-Aware Projection-Out (CAPO) for distinguishing inter-NP CA maps and Adaptive Token Mixing (ATM) for enhancing inter-NP separation while maintaining intra-NP cohesion in CA maps. Extensive experiments confirm that TokeBi outperforms prior arts across diverse baselines and datasets.
\end{abstract}


\section{Introduction}
\label{sec:intro}
Text-to-image (T2I) synthesis has made significant advances in generating high-fidelity images from text prompts, driven by diffusion models~\cite{ chen2024pixartsigma,chen2024pixart, esser2024scaling,podell2023sdxl, rombach2022high, zhuo2024lumina}.
However, ensuring precise adherence to prompts in image generation remains challenging. The issues such as \romannumeral 1) missing and \romannumeral 2) misbinding of objects and attributes disrupt compositional generation~\cite{feng2022training, rassin2024linguistic} and 
are more pronounced as the number of objects (\textit{i.e.}, multi-object) and attributes (\textit{i.e.}, multi-attribute) increases within a prompt~\cite{chefer2023attend, podell2023sdxl, rassin2022dalle}.
These problems ultimately degrade text-image alignments.

The \emph{semantic binding} task has emerged to address this challenge by aiming to generate objects that correspond to each noun phrase (NP) in the text prompt.
A key observation in this context is that, for effective semantic binding, cross-attention (CA) maps corresponding to text tokens from different NPs (inter-NP) should be separated, while CA maps for tokens within the same NP (intra-NP) should remain clustered together~\cite{marioriyad2024attention, rassin2024linguistic}.
Leveraging this insight, various approaches have been proposed, including latent vector optimization~\cite{chefer2023attend, dahary2024yourself, hu2024token,kim2024text, li2023divide, rassin2024linguistic, sueyoshi2024predicated}, text embedding optimization~\cite{hu2024token}, or text embedding modification~\cite{chen2024cat, feng2022training,zhuang2024magnet} to enhance semantic binding.
However, these approaches often rely on spatial priors or extra user input~\cite{dahary2024yourself,hu2024token}, limiting the compositional diversity of generated images.
Moreover, increased alteration of the text embeddings themselves during optimization~\cite{hu2024token} can lead to distorted token semantics, resulting in inefficiency and suboptimal performance.

We investigated the influence of token embeddings, specifically their geometric properties such as \emph{norm} and \emph{angular distance}, on semantic binding, which has not been thoroughly explored.
Here, we found that when token embeddings with high similarity appear across distinct NPs, significant overlaps in CA maps occur, causing poor semantic binding.
Increasing the embedding norm of either attribute or object tokens alleviates semantic neglect, a phenomenon where the semantics of certain tokens are omitted in the generated image, thereby enhancing semantic binding. A larger embedding norm increases distances between token embeddings, consequently reducing undesirable CA overlaps. Interestingly, our analysis in this work reveals that prior methods \emph{implicitly} leverage this norm-based effect. These findings are supported by our comprehensive empirical and theoretical analyses.

We propose \textbf{TokeBi}, a training-free approach to enhance semantic binding in T2I diffusion models. TokeBi consists of several key components, building upon our theoretical findings. First, to mitigate issues in semantic binding arising from angular distance across text tokens, we propose Causality-Aware Projection Out (CAPO) to enforce orthogonalization.
Given the intrinsic asymmetry in \emph{causal} text encoders (\textit{e.g.}, CLIP~\cite{radford2021learning}) where earlier tokens have no access to latter ones, we adopt asymmetric Schmidt orthogonalization~\cite{leon2013gram}, projecting out preceding token embeddings from later ones.
In contrast, for \emph{non-causal} text encoders (\textit{e.g.}, T5~\cite{raffel2020exploring}), we employ L\"owdin orthogonalization~\cite{mayer2002lowdin} to symmetrically project token embedding pairs.
Second, inspired by our findings of the relationship between token norm and CA map, we introduce Adaptive Token Mixing (ATM) 
to optimize token norms with the mixing matrix within NPs, effectively improving semantic binding.
Our loss design, based on entropy~\cite{shannon1948mathematical} and Bhattacharyya distance~\cite{sen1996anil}, ensures intra-NP compactness and inter-NP distinction in CA maps by optimizing the mixing matrix.

To evaluate TokeBi's semantic binding performance, we conducted extensive experiments comparing it with various SOTA training-free semantic binding methods. The evaluation employed diverse metrics, including BLIP-VQA~\cite{huang2023t2i}, GenEval~\cite{ghosh2023geneval}, CIC~\cite{chen2024cat}, VQAScore~\cite{lin2024evaluating}, VIEScore~\cite{ku2024viescore}, and human evaluation across multiple T2I benchmarks.
The results demonstrate that TokeBi outperforms existing methods across most metrics on various datasets and baselines, setting a new benchmark.

\noindent\textbf{Key contributions:} \textbf{\romannumeral 1)} We \emph{theoretically} and \emph{empirically} analyze the geometrical properties of token embeddings influencing semantic binding, leading to two necessary conditions and enabling to reinterpret existing works in a new perspective. \textbf{\romannumeral 2)} Building on these findings, we propose \textbf{TokeBi}, a training-free method designed to effectively tackle the T2I semantic binding task. TokeBi comprises novel components: CAPO, ATM, and a loss formulation for condensed intra-NP and separated inter-NP.
\textbf{\romannumeral 3)} We extensively evaluate TokeBi across various baselines, datasets, and metrics, demonstrating that it achieves a new state-of-the-art performance for the semantic binding task in T2I generation.

\section{Related Works}
\label{sec:related_work}
\paragraph{Semantic leakage and neglect in T2I generation.}
T2I diffusion models~\cite{chen2024pixartsigma, chen2024pixart, podell2023sdxl, rombach2022high} generate high-quality images aligned with input text by leveraging cross-attention mechanisms between latent vectors and text embeddings. 
However, they often struggle to accurately reflect input prompts, particularly in complex scenarios involving multiple objects and attributes.
This leads to two major issues~\cite{chefer2023attend, podell2023sdxl, rassin2022dalle}: semantic neglect, where certain objects or attributes in the prompt are ignored, and semantic leakage, where attributes of one object mistakenly appear on another. Addressing these issues has led to the emergence of the semantic binding task.

Attend-and-Excite (A\&E)~\cite{chefer2023attend} optimizes latent vectors to ensure all subject tokens activate their corresponding image patches in CA maps, mitigating both semantic neglect and leakage. SynGen~\cite{rassin2024linguistic} optimizes latents to ensure that intra-NP tokens (within the same NP) have similar CA maps, while inter-NP tokens (across different NPs) maintain distinct CA maps. Additionally, Marioriyad \etal~\cite{marioriyad2024attention} quantifies overlap in CA maps using the variance and center-of-mass distance, targeting semantic neglect.
Since textual input alone provides weak conditioning~\cite{stap2020conditional, tewel2024training}, some methods incorporate spatial guidance via annotations such as bounding boxes~\cite{chen2024training, dahary2024yourself,ma2024directed, mao2023training,phung2024grounded, wang2024compositional, xie2023boxdiff} and masks~\cite{kim2023dense, park2024shape}.
Additionally, some studies~\cite{hao2023optimizing, phung2024grounded, zhong2023adapter} introduce multi-modal frameworks that leverage large language models' (LLMs) capabilities.
While these methods are effective, they lack a direct study on the role of text embeddings. Additionally, spatial prior-based methods require extra user input or limit compositional flexibility. 

\vspace{-0.5em}
\paragraph{Token-driven approaches for semantic alignment.}
Recent efforts have explored the role of text token embeddings in T2I synthesis, aiming to enhance semantic binding. 
StructureDiffusion~\cite{feng2022training} enhances semantic binding by using embeddings from hierarchical phrase structures rather than a single textual sequence.
ToMe~\cite{hu2024token} additively merges a set of token embeddings within an NP using fixed coefficients, substitutes the end token, and optimizes both text embeddings and latents to cluster CA maps into fixed positions.
Magnet~\cite{zhuang2024magnet} manipulates token embeddings to bind objects and attributes with a positive and negative dataset. 
Despite advances in semantic binding, these approaches have limitations: performance gains are marginal~\cite{feng2022training}, flexibility is restricted by fixed coefficients (\textit{e.g.}, 1.1 or 1.2) used in the linear summation of text token embeddings and predefined spatial information~\cite{hu2024token}, mitigation of semantic neglect is limited to objects only~\cite{chen2024cat}, and reliance on datasets introduces performance variability depending on the dataset~\cite{zhuang2024magnet}.
None of the methods fully address the relationship between the geometrical properties of text token embeddings and CA maps, nor leverage it in their design.

\section{Analysis on Geometrical Properties of Text Token in Cross-Attention}
\label{sec:analysis}
\subsection{Preliminaries: Cross-Attention in Diffusion Models}
\label{subsec:background}
T2I diffusion frameworks~\cite{chen2024pixartsigma, chen2024pixart, podell2023sdxl, rombach2022high} typically condition the denoising network using text embeddings obtained from a text encoder. This conditioning is achieved through the cross-attention (CA) module within the denoising network. Let the text embeddings be represented as $T\in\mathbb{R}^{L\times d_1}$ where $L$ is the number of tokens and $d_1$ is the dimension of text embedding.
We also denote $H \in \mathbb{R}^{N \times d_2}$ as the latent tokens where $N$ is the number of latent tokens and $d_2$ is their dimension.
Each vector is then projected into the query $Q$, key $K$, and value $V$ via learned linear transformations as $Q = HW_Q$, $K = TW_K$, and $V = TW_V$, where $W_K, W_V \in \mathbb{R}^{d_1 \times d}$ and $W_Q \in \mathbb{R}^{d_2 \times d}$.
The $Q$, $K$, and $V$ are then used in the CA operation as follows:
\begin{equation}
    \text{CrossAttention}(Q, K, V) =
    \text{softmax}\!\Bigl(\tfrac{Q K^\top}{\sqrt{d}}\Bigr) V ,
    \label{eq:attention}
\end{equation}
where softmax operation ensures that, for each text token, the probabilities across all latent tokens sum to 1, yielding CA maps $P \in \mathbb{R}^{N \times L}$. For the $i$-th text token, we denote $A_i$ to the sum-1 normalized distribution of its corresponding CA map $P_i \in \mathbb{R}^N$.

\subsection{Analysis 1: Embedding Distance and Semantic Binding}
\label{subsec:embedding_distance_analy}
\begin{figure}[!t]
\begin{center}
\centerline{\includegraphics[width=\textwidth]{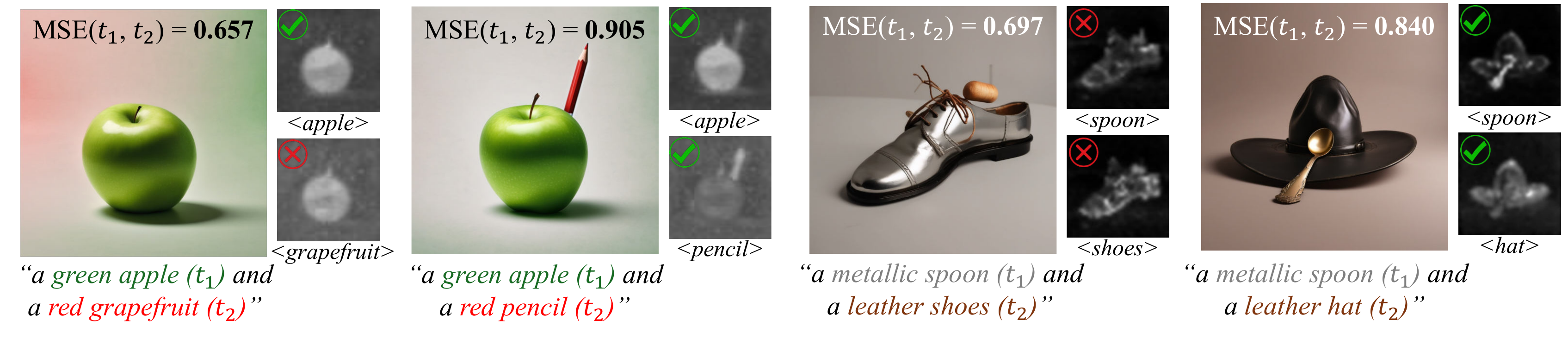}}
\vspace{-0.8em}
\caption{\textbf{Relationship between token embedding mean squared error (MSE) and semantic binding.}
Upper numbers indicate the MSE between noun tokens (t$_1$, t$_2$) within each NP. Stacked images show attention maps of the token below. Lower MSE leads to object neglect (\textit{e.g.}, \textit{``grapefruit''}, \textit{``spoon''}) due to substantial attention overlap, while higher MSE enhances text-image alignment.
Further details on the relationship between MSE and semantic binding can be found in Appendix Sec.~\ref{sup_subsec:additional_analysis_results}.
}
\label{fig:mse}
\vspace{-2.5em}
\end{center}
\end{figure}

Previous studies~\cite{hu2024token, marioriyad2024attention, rassin2024linguistic, zhang2024attention} have shown that semantic binding deteriorates when token attention maps overlap excessively or become too dispersed. However, the influence of token embedding geometries on these overlaps and dispersions remains unexplored. To investigate this, we experiment with several multi-attribute multi-object prompts differing with a fixed seed.
In Fig.~\ref{fig:mse}, we replaced \romannumeral 1) \textit{`grapefruit'} with \textit{`pencil'} in the prompt \textit{``a green apple and a red \underline{grapefruit},''} and \romannumeral 2) \textit{`shoes'} with \textit{`hat'} in the prompt \textit{``a metallic spoon and a leather \underline{shoes}.''}
Then, we measured token embedding mean squared error (MSE) between noun tokens within different noun phrases (NPs). We find that lower embedding MSE between distinct NPs indicates semantic neglect, whereas higher MSE suggests better semantic binding.
This is also evident in the CA maps of the corresponding tokens in Fig.~\ref{fig:mse}.
To further validate this, we measured the MSE and GenEval~\cite{ghosh2023geneval} scores across 100 prompts from the color attribute set. Spearman’s rank correlation test between them revealed a strong positive correlation ($\rho=0.95$, $p$-value 0.005), indicating statistical significace.
Further details of relationship between MSE and semantic binding are in Appendix Sec.~\ref{sup_subsec:additional_analysis_results}.
Analysis shows that lower embedding MSE leads to greater overlap in CA maps, weakening semantic binding, while higher MSE produces more distinct CA maps, strengthening it. This observation motivates the following proposition.

\begin{proposition}[KL Divergence Monotonicity in Distance]\label{prop:mse}
    Let $Q=[q_1; \cdots; q_N]$ where $q_a$ is the $a$-th query token corresponds to $a$-th token embedding $t_a$. Then, increasing the distance between text token embeddings $\|t_i - t_j\|^2$ monotonically increases the Kullback-Leibler divergences $D_{KL}(A_i\|A_j)$.   
\end{proposition}

\noindent Proof is provided in Appendix Sec.~\ref{sup_subsec:proof_proposition_1}.
Proposition~\ref{prop:mse} states that as the distance between two token embeddings increases, their corresponding CA maps become more distinct.
Our empirical and theoretical analyses suggest that increasing token embedding distance across different NPs is crucial for preventing CA overlap, thereby mitigating object neglect and preserving semantics. Notably:
\vspace{-0.5mm}
\begin{tcolorbox}[before skip=2mm, after skip=0.0cm, boxsep=0.0cm, middle=0.0cm, top=0.1cm, bottom=0.1cm]
To increase the distance of text token embeddings, one can either \romannumeral 1) increase the \emph{angular distance} between the embeddings or \romannumeral 2) enlarge their \emph{norms}..
\end{tcolorbox}
\vspace*{1mm}
A remedy for increasing the angle is in Sec.~\ref{subsec:capo}. Analysis for enlarging the norms follows next.

\subsection{Analysis 2: Embedding Norm and Semantic Binding}
\label{subsec:embedding_norm_analy}
\paragraph{Relationship between the token embeddings norm
and semantic binding.}
Existing semantic binding approaches~\cite{hertz2023prompt, hu2024token, zhuang2024magnet} adjust text embeddings to control CAs but do not explicitly link token embedding norm scaling to attribute and object activations within the CA maps.

\begin{figure}[!t]
\begin{center}
\centerline{\includegraphics[width=\linewidth]{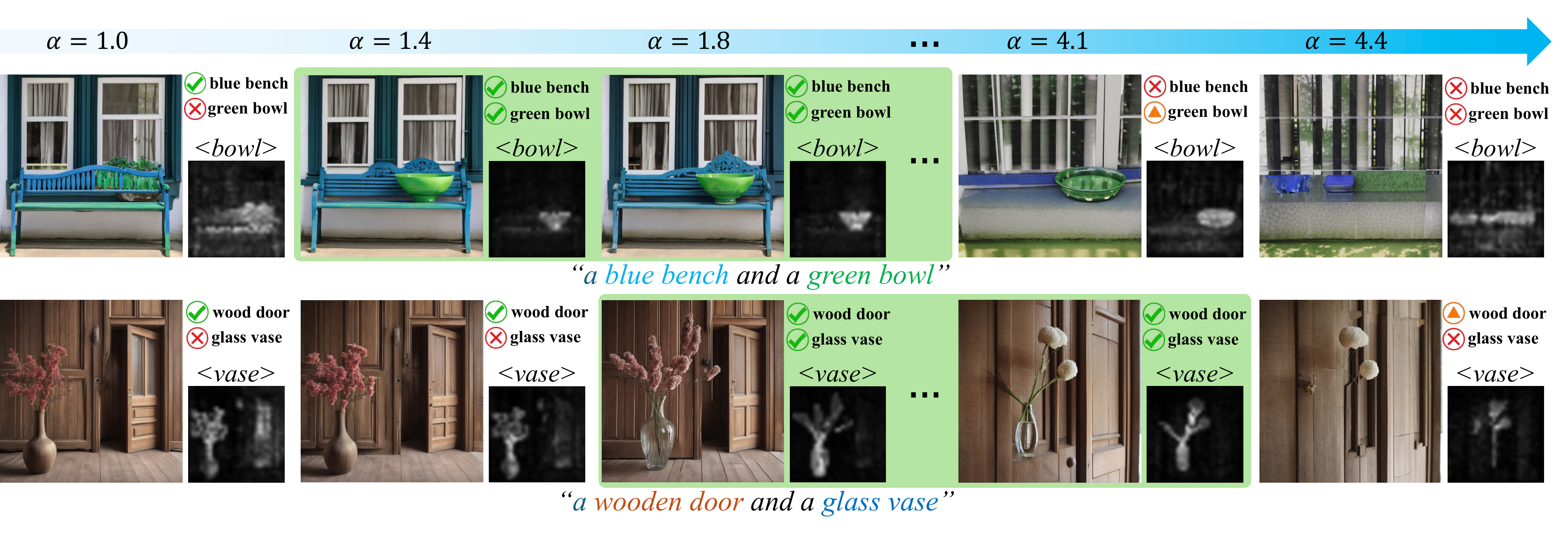}}
\vspace{-1.0em}
\caption{\textbf{Semantic binding under token embedding scaling.} We generate images by scaling the text token embeddings of under-represented objects using a parameter $\alpha$, and visualize CA maps via SDXL. Moderate scaling enhances the visibility of neglected objects like ``\textit{green bowl}'' and ``\textit{glass vase}'', producing condensed attention. Excessive scaling, however, harms semantic alignment. The optimal $\alpha$ depends on prompt and random seed, suggesting the need for adaptive scaling.}
\label{fig:norm_scaling}
\vspace{-2.5em}
\end{center}
\end{figure}
As illustrated in Fig.~\ref{fig:norm_scaling}, scaling the norm of text embeddings in T2I synthesis model improves the visibility of underrepresented semantics.The optimal scaling factor varies with prompt and seed but remains consistent across attributes and objects, highlighting the need for adaptive scaling. The proposition below explains how increased embedding norms enhance semantic binding.

\begin{proposition}
Let two token embeddings $t_i,\;t_j\in \mathbb{R}^d$ follow the assumptions 1) and 2) in Appendix Sec.~\ref{sup_subsec:proof_proposition_2}. With scalar values $\lambda_i,\;\lambda_j > 1$, the following holds.
\begin{equation}
    \|t_i-\;t_j\|_2^2 < \|\lambda_i t_i-\lambda_jt_j\|_2^2.
\end{equation}
\label{prop:norm_scaling}
\vspace{-1.5em}
\end{proposition}
\noindent The proof, assumptions and verifications are provided in Appendix (Sec.~\ref{sup_subsec:proof_proposition_2} and Sec.~\ref{sup_subsec:verification_of_assumptions}). Proposition~\ref{prop:norm_scaling} suggests that, under certain conditions, scaling the norm of token embeddings increases the distance between two vectors. This leads to a more distinct CA map, as supported by Proposition~\ref{prop:mse}.
A remedy for effectively increasing the norm of token embeddings is provided in Sec.~\ref{subsec:token_mixing}.

\vspace{-0.5em}
\paragraph{Re-interpreting existing token-driven methods based on our norm scaling.}
Recent T2I diffusion models rely on text embeddings to condition the image synthesis. These embeddings, extracted from a text encoder, play a crucial role in guiding the generative process. Assuming that token embedding vectors exhibit Gaussian properties~\cite{qian2021conceptualized, vilnis2015word}, their behavior can be analyzed as following.
\begin{remark}[Norm of token embedding add \& sub]
Suppose that two token embeddings \( t_i, t_j \in \mathbb{R}^d \) 
follow \( t_i, t_j \sim \mathcal{N}(\mu_d, \Sigma_d) \) and they satisfy $\|\mu_d\|_2 \ll \|t_i\|_2, \|t_j\|_2$. Then:
\begin{equation}
\mathbb{E}[\|t_i\|_2],\ \mathbb{E}[\|t_j\|_2] \le \mathbb{E}[\|t_i\pm t_j\|_2].
\end{equation}
\label{remark:add_norm}
\end{remark}
\vspace{-1em}
The proof and assumptions are in Appendix (Sec.~\ref{sup_subsec:proof_remark_1} and \ref{sup_subsec:verification_of_assumptions}).
Remark~\ref{remark:add_norm} implies that adding or subtracting two token embeddings increases the norm of the resulting embedding.

Based on this Remark~\ref{remark:add_norm}, we re-interpret Prompt-to-Prompt~\cite{hertz2023prompt} as scaling attention maps by adjusting text embedding norms, thereby modulating semantic influence. Similarly, Magnet~\cite{zhuang2024magnet} strengthens or weakens prompts by adding or subtracting their embeddings, respectively, which increases the embedding norm according to Remark~\ref{remark:add_norm}.
More analysis is in Appendix~\ref{sup_subsec:additional_analysis_results}.

\begin{figure}[!t]
\centering
  \begin{subfigure}{0.49\linewidth} 
    \includegraphics[width=1\linewidth]{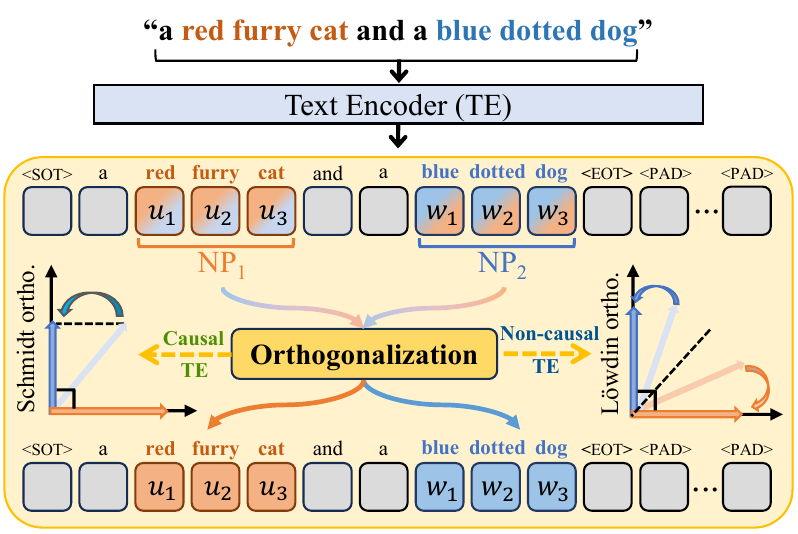}
    \caption{Causality-Aware Projection Out (CAPO).}
    \label{fig:overall_capo}
  \end{subfigure}
  \hfill
  \begin{subfigure}{0.49\linewidth} 
    \includegraphics[width=1\linewidth]{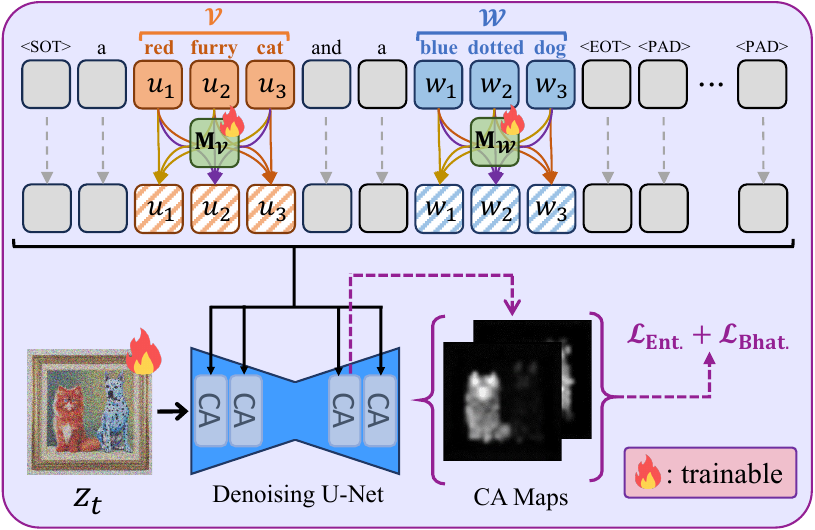}
    \caption{Adaptive Token Mixing (ATM).}
    \label{fig:overall_atm}
  \end{subfigure}
\vspace{-0.5em}
\caption{
\textbf{Overview of TokeBi.} TokeBi comprises two core components: Causality-Aware Projection Out (CAPO) and Adaptive Token Mixing (ATM).
CAPO increases the angular distances between text token embeddings in inter-noun phrases (NPs) through orthogonalization, guided by the text encoder’s causality: Schmidt orthogonalization is used for causal encoders, and L\"owdin for non-causal ones.
ATM adjusts embedding magnitudes by applying a learnable transformation matrix $\mathbf{M}$ to mix tokens within intra-NPs, thereby strengthening semantic binding.
Both $\mathbf{M}$ and the latent vectors $z_t$ are optimized using our loss design, $\mathcal{L}_\text{Ent.} + \mathcal{L}_\text{Bhat.}$, derived from the cross-attention maps.
}\label{fig:method}
\vspace{-1.0em}
\end{figure}

\section{TokeBi: Reflecting Geometrical Properties of Text Token Embeddings}

Sec.~\ref{sec:analysis} highlights the importance of text token embedding distances, particularly \emph{angular distance} and \emph{norm}, in semantic binding.
We propose Causality-Aware Projection Out (CAPO) to increase inter-NP angular distances by orthogonalizing embeddings.
Adaptive Token Mixing (ATM) then enables intra-NP norm scaling without semantic information loss.
Optimized with entropy and Bhattacharyya losses, ATM condenses intra-NP and separates inter-NP CA maps.

\subsection{Causality-Aware Projection Out} 
\label{subsec:capo}
Drawing from the insights in Proposition~\ref{prop:mse}, we apply orthogonalization to increase angular distances between token embeddings from distinct noun phrases (NPs), thus reducing dependency. Considering text encoder causality, we note that causal encoders employ attention masks, restricting tokens to attend only to preceding positions, whereas non-causal encoders allow unrestricted mutual attention. Therefore, we propose \emph{Causality-Aware Projection Out (CAPO)}, aiming to preserve maximal token information while increasing inter-NP distances. Defining preceding NP tokens as $ \mathcal{U} = \{u_1, u_2, \dots, u_m\} $ and succeeding NP tokens as $ \mathcal{W} = \{w_1, w_2, \dots, w_n\}$, CAPO uses Schmidt orthogonalization~\cite{leon2013gram} in causal encoders, projecting each $w_i\in\mathcal{W}$ orthogonally against tokens $u_j\in\mathcal{U}$, yielding $w_i'$ as follows:
\vspace{-0.5em}
\begin{equation}
    w_i' = w_i - \sum_{j=1}^{m} \frac{\langle w_i, u_j \rangle}{\langle u_j, u_j \rangle} u_j,
\end{equation}
where $ \langle u_i, w_j \rangle $ represents the inner product between $ u_i $ and $ w_j $.
This transformation ensures that embeddings in $\mathcal{W}$,  embeddings in the latter NP, remain orthogonal to all embeddings in $ \mathcal{U} $, eliminating redundant dependencies while preserving the semantic integrity of reference embeddings in the former NP. For a non-causal text encoder with bidirectional dependencies, we apply a symmetric approach using L\"owdin orthogonalization~\cite{mayer2002lowdin}, ensuring equitable decorrelation of token embeddings across NPs.
Given a token embedding matrix $X=[u_i,w_j]$ spanning different NPs, we compute the orthogonalized pair $X'=[u_i',w_j']$ as follows:
\begin{equation}
    X'=X(X^\top X)^{-1/2}.
\end{equation}
The overall process of CAPO is illustrated in Fig.~\ref{fig:overall_capo}. Leveraging CAPO allows for attributes and objects across NPs to stay separate, which improves semantic binding and maintains compositional consistency during the T2I generation process that involves multiple objects and attributes.

\subsection{Adaptive Token Mixing} 
\label{subsec:token_mixing}
Building on the insights from Proposition~\ref{prop:norm_scaling} and Fig.~\ref{fig:norm_scaling}, we propose \emph{Adaptive Token Mixing (ATM)}, which adaptively enhances the distinctiveness of CA maps for each NP. Instead of solely controlling token norms, ATM employs a learnable mixing matrix to blend token embedding semantics. This approach not only adjusts token norms but also keeps their intended semantics in NP, thereby improving semantic binding performance, which is illustrated in Fig.~\ref{fig:overall_atm}.

For a given token embedding set of an NP, $\mathcal{V} = \{v_1, v_2, \dots, v_n\}$, we define the matrix $\mathbf{V} \in \mathbb{R}^{n \times d}$, where each token embedding vector is stacked along the row dimension and $d$ denotes the embedding dimension. We then initialize a learnable \emph{mixing matrix} $\mathbf{M}_{\mathcal{V}} \in \mathbb{R}^{n \times n}$. The mixing operation is performed as Eq.~\ref{eq:atm} to generate a new representation $\mathbf{V^*}\in\mathbb{R}^{n\times d}$, which is subsequently utilized in both the optimization step and the image synthesis process:
\begin{equation}\label{eq:atm}
    \mathbf{V^*} = \mathbf{M}_{\mathcal{V}}\mathbf{V}.
\end{equation}
For example, given an NP with one object with two attributes, such as ``red furry cat'', we can construct a token embedding set $\mathcal{V}=\{v_{\text{red}}, v_{\text{furry}},v_{\text{cat}}\}$, build a token matrix $\mathbf{V}\in\mathbb{R}^{3\times d}$ with $\mathcal{V}$, and initialize $\mathbf{M}_{\mathcal{V}} \in \mathbb{R}^{3 \times 3}$ as follows:
\vspace{-0.5em}
\begin{equation}\label{eq:atm2}
    \mathbf{M_{\mathcal{V}}} =
    \begin{bmatrix}
    a_{11} & a_{12} & a_{13} \\
    a_{21} & a_{22} & a_{23} \\
    a_{31} & a_{32} & a_{33}
    \end{bmatrix}
    ,\quad 
    \mathbf{V} =
    \begin{bmatrix}
     v_{\text{red}}\\
    v_{\text{furry}}\\
    v_{\text{cat}}
    \end{bmatrix},
\end{equation}
The updated token embeddings are computed as $\mathcal{V^*}=\{v^*_{\text{red}}, v^*_{\text{furry}},v^*_{\text{cat}}\}$ using Eq.~\ref{eq:atm} and~\ref{eq:atm2}.
Our approach adaptively adjusts token norms to capture semantics in NP while enhancing individual token representation. By mixing tokens within an NP, it strengthens entailment relationships, fostering cohesive meaning integration and reinforcing semantic consistency for more effective semantic binding.
Our ATM dynamically optimizes mixing and scaling coefficients, guided by a unique loss design detailed in Sec.\ref{subsec:loss_design}, offering greater flexibility and substantially improving semantic binding.

\subsection{Loss-Guided Cross-Attention Condensation and Separation} 
\label{subsec:loss_design}
Recall that effective semantic binding requires condensed and distinct CA maps. We previously introduced necessary conditions and the \emph{ATM}, which leverage the conditions. This structure is optimized via an objective explicitly designed to achieve condensed and distinct CA maps.

\begin{wrapfigure}[11]{r}{0.48\textwidth}
    \vspace{-0.6em}
    \centering
    \begin{subfigure}{0.45\linewidth}
        \includegraphics[width=1\linewidth]{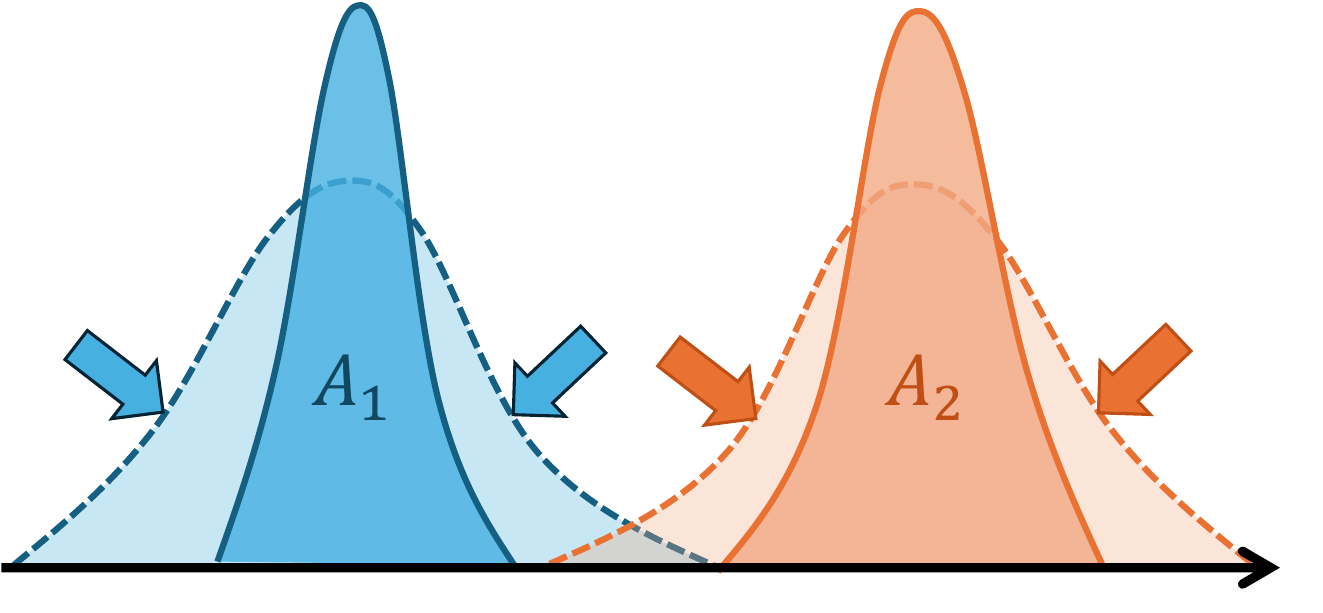}
        \caption{Effect of $\mathcal{L}_{\text{Ent.}}$.}
        \label{fig:ent_loss}
    \end{subfigure}
    \begin{subfigure}{0.45\linewidth}
        \includegraphics[width=1\linewidth]{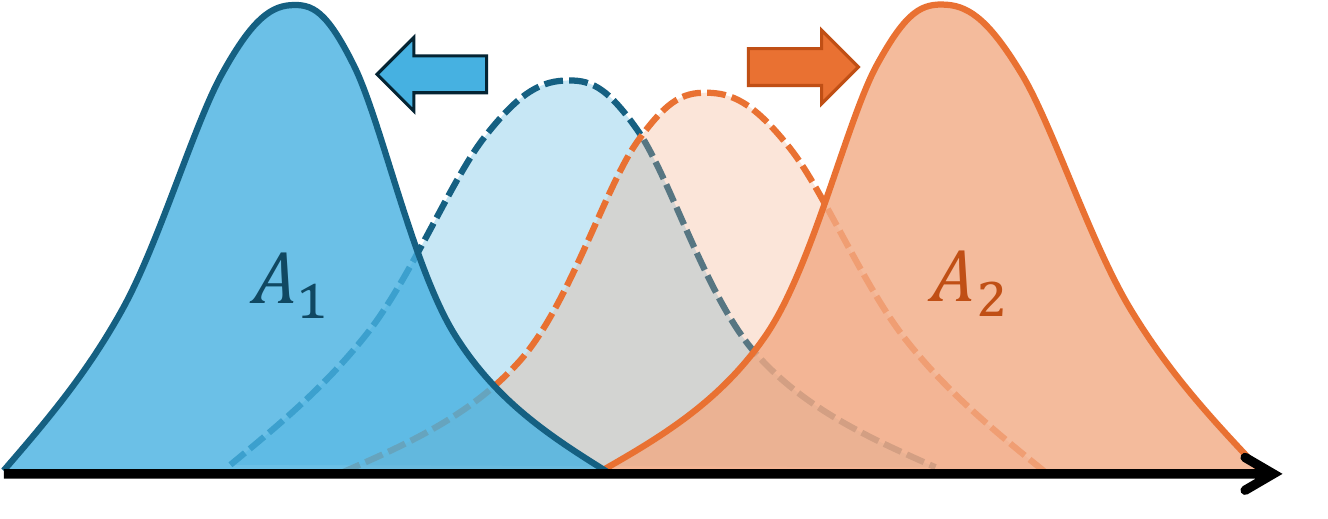}
        \caption{Effect of $\mathcal{L}_{\text{Bhat.}}$.}
        \label{fig:baha_loss}
    \end{subfigure}
    \vspace{-0.2em}
    \caption{\textbf{Illustration of the effects of $\mathcal{L}_{\text{Ent.}}$ and $\mathcal{L}_{\text{Bhat.}}$ on two probability distributions.} \textbf{(a)} As $\mathcal{L}_{\text{Ent.}}$ decreases, each distribution, $A_1$ and $A_2$, becomes more condensed. \textbf{(b)} As $\mathcal{L}_{\text{Bhat.}}$ decreases, the overlap between $A_1$ and $A_2$ is reduced.}
    \label{fig:loss}
    \vspace{-1em}
\end{wrapfigure}

For the condensed CA maps, we employ a Shannon entropy~\cite{shannon1948mathematical} loss that encourages attention condensation by minimizing entropy. Given the probability $ p_i $ over an $ A_k $ of $k$-th token, the entropy loss $\mathcal{L}_{\text{Ent.}}$ is formulated as:
\begin{equation}
    \mathcal{L}_{\text{Ent.}} = -\sum_{k\in \mathcal{K}}\sum_{p_i \in A_k} p_i\log p_i,
\end{equation}
where $\mathcal{K}$ denotes the set of object token indices. 
As shown in Fig.~\ref{fig:ent_loss}, this objective facilitates the formation of localized CA maps, effectively preserving the object-attribute relationships.

To make the inter-NP CA map more distinct, we introduce the Bhattacharyya distance~\cite{sen1996anil} loss, which measures the overlap between probability distributions, encouraging more distinct attention across phrases.
It is formulated as:
\begin{equation}
\mathcal{L}_{\text{Bhat.}} = \sum_{(m ,n)\in\mathcal{S}}\Bigg(\sum_{i=1}^N\Bigg(\sum_{p_i\in A_m,\ q_i\in A_n}\sqrt{p_iq_i}\Bigg)\Bigg),
\end{equation}
where $\mathcal{S}$ is the set of object token index pairs from different NPs.
As shown in Fig.~\ref{fig:baha_loss}, $\mathcal{L}_{\text{Bhat.}}$ promotes mutually exclusive CA maps by reducing overlap, thereby enhancing semantic binding.

By combining the two previously introduced objectives, $\mathcal{L}_{\text{Ent.}}$ and $\mathcal{L}_{\text{Bhat.}}$, we formulate the final loss function $\mathcal{L}_{\text{total}} = \mathcal{L}_{\text{Ent.}} + \lambda\mathcal{L}_{\text{Bhat.}}$, serving as the optimization objective.
where $ \lambda$ is a coefficient that balances intra-NP condensation and inter-NP separation.
With $\mathcal{L}_{\text{total}}$, we optimized latent vectors and mixing matrix parameter set $\theta$ via backpropagation with step size $\eta$ as $\theta '=\theta - \eta\cdot\nabla_{\theta}\mathcal{L}_{\text{Total}}$.
Our loss uses no spatial priors and relies solely on two key insights: CA map condensation and separation. Its effectiveness in semantic binding is empirically validated in the following section.


\begin{table}[t]
\caption{\textbf{Comparison of training-free semantic binding methods for T2I generation.} 
Across various benchmark datasets and metrics, 
TokeBi shows superior performance without requiring additional training, achieving a \textbf{120.7\%} improvement over the SDXL in GenEval.
The user study for human evaluation on perceptual quality and semantic alignment accuracy is in Appendix Sec.~\ref{sup_subsec:add_quanti}.}
\label{tab:2x2_quant_table}
\centering
\renewcommand{\arraystretch}{1.0} 
\setlength{\tabcolsep}{0.55em} 
\resizebox{0.99\textwidth}{!}
{%
\begin{tabular}{c|ccc|c|c|c|c}
\toprule
\multirow{2}{*}{Method} & \multicolumn{3}{c|}{BLIP-VQA~\cite{huang2023t2i} $\uparrow$} & \multirow{2}{*}{\tabincell{c}{GenEval~\cite{ghosh2023geneval} $\uparrow$\\Color attribute}} & \multirow{2}{*}{\tabincell{c}{CIC~\cite{chen2024cat} $\uparrow$\\`2 Obj. exist'}} & \multirow{2}{*}{\tabincell{c}{VQAScore~\cite{lin2024evaluating} $\uparrow$}} & \multirow{2}{*}{\tabincell{c}{VIEScore~\cite{ku2024viescore} $\uparrow$}} \\
& Color & Texture & Shape & & & \\ 
\midrule
SDXL~\cite{podell2023sdxl} & 0.6121 & 0.5568 & 0.4982 & 0.2050 & 0.518 & 0.7327 & 7.380 \\ 
\midrule
StructureDiff$_{XL}$~\cite{feng2022training} & 0.5092 & 0.4996 & 0.4416 & 0.1375 & 0.436 & 0.6924 & 7.148 \\ 
A\&E$_{XL}$~\cite{chefer2023attend} & 0.6309 & 0.5550 & 0.5075 & 0.2325 & 0.536 & 0.7333 & 7.560 \\ 
SynGen$_{XL}$~\cite{rassin2024linguistic} & 0.6549 & \lightorangenum{0.5734} & \lightorangenum{0.5092} & 0.1550 & 0.533 & 0.7502 & 7.804  \\
Magnet~\cite{zhuang2024magnet} & \lightorangenum{0.6839} & 0.5686 & 0.5056 & \lightorangenum{0.2875} & \orangenum{0.555} & \lightorangenum{0.7870} & \lightorangenum{7.996} \\
ToMe~\cite{hu2024token} & \orangenum{0.7013} & \orangenum{0.6490} & \orangenum{0.5455} & \orangenum{0.4001} & \lightorangenum{0.546} & \orangenum{0.8078} & \orangenum{8.201} \\
TokeBi \textbf{(Ours)} & \rednum{0.7610} & \rednum{0.6581}  & \rednum{0.5472} & \rednum{0.4525} & \rednum{0.639} & \rednum{0.8219} & \rednum{8.328} \\
\bottomrule
\end{tabular}
}
\end{table}

\section{Experiments}
\label{sec:experiments}

\subsection{Experimental Setups}
\label{subsec:experimental_setups}

\vspace{-0.2em}
\paragraph{Baselines.}
\begin{wraptable}{r}{0.55\textwidth}
\vspace{-1.2em}
\caption{\textbf{Comparison of training-free semantic binding methods on the challenging 3$\times$3 prompts.}  
While other semantic binding methods either underperform compared to na\"ive SDXL or achieve only marginal improvements in the $3\times3$ setting, TokeBi outperforms them by a large margin, achieving gains of up to \textbf{44.6\%}.}
\label{tab:3x3_quant_table}
\centering
\renewcommand{\arraystretch}{0.9} 
\setlength{\tabcolsep}{0.2em} 
\resizebox{0.55\textwidth}{!}{%
\begin{tabular}{c|c|c|c}
\toprule
\multirow{2}{*}{Method} & \multirow{2}{*}{\tabincell{c}{CIC~\cite{chen2024cat} $\uparrow$\\ `2 Obj. exist'}} & \multirow{2}{*}{\tabincell{c}{VQAScore~\cite{lin2024evaluating}} $\uparrow$} & \multirow{2}{*}{\tabincell{c}{VIEScore~\cite{ku2024viescore}} $\uparrow$} \\
& & & \\
\midrule
SDXL~\cite{podell2023sdxl} & 0.260 & 0.6886 & 6.120 \\
\midrule
StructureDiff$_{XL}$~\cite{feng2022training} & 0.204 & 0.6240 & 5.792 \\
A\&E$_{XL}$~\cite{chefer2023attend} & 0.275 & 0.6805 & \lightorangenum{6.248} \\ 
SynGen$_{XL}$~\cite{rassin2024linguistic} & 0.280 & \orangenum{0.6997} & \orangenum{6.408} \\ 
Magnet~\cite{zhuang2024magnet} & \lightorangenum{0.304} & 0.6907 & 6.240 \\
ToMe~\cite{hu2024token} & \orangenum{0.340} & \lightorangenum{0.6919} & 6.133 \\
{TokeBi \textbf{(Ours)}} & \rednum{0.376} & \rednum{0.7422} & \rednum{7.020}\\
\bottomrule
\end{tabular}
}
\vspace{-1em}
\end{wraptable}

To assess TokeBi's performance, we compare it with several baselines: ToMe~\cite{hu2024token}, StructureDiffusion~\cite{feng2022training} (StructureDiff), Attend-and-Excite~\cite{chefer2023attend} (A\&E), SynGen~\cite{rassin2024linguistic}, and Magnet~\cite{zhuang2024magnet}.
Quantitative comparisons employ SDXL~\cite{podell2023sdxl}, while qualitative analyses additionally use PlayGround-v2 (PlayG-v2)~\cite{liplayground} and PixArt-$\Sigma$~\cite{chen2024pixartsigma}. 
Notably, SDXL and PlayG-v2 use CLIP~\cite{radford2021learning}, a causal text encoder, whereas PixArt-$\Sigma$ utilizes T5~\cite{raffel2020exploring}, a non-causal text encoder.
None of the baselines require extra training.
For automatic NP extraction, we utilize SpaCy~\cite{honnibal2017spacy} for syntactic parsing.
To ensure fairness, we use official implementations for each method, adapting them when necessary for unsupported base models. Implementation details are in Appendix (Sec.~\ref{sup_sec:implementation_details}).

\begin{figure*}[!t]
\begin{center}
\centerline{\includegraphics[width=\textwidth]{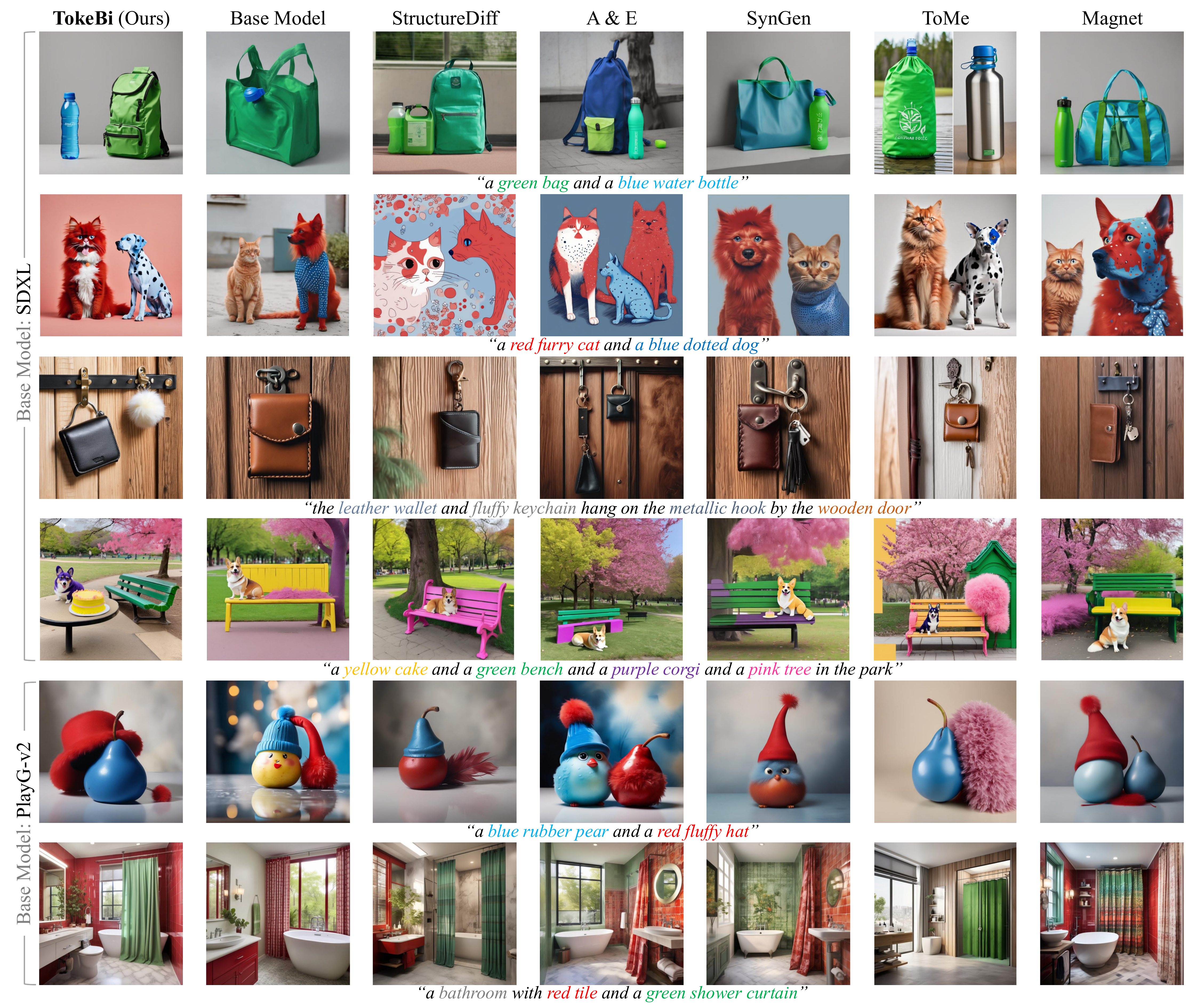}}
\vspace{-0.5em}
\caption{\textbf{Qualitative comparison of training-free semantic binding methods.} 
TokeBi generates images with improved text alignment for both simple and complex prompts.
Additional qualitative results, including \emph{uncurated samples}, are provided in Appendix Sec.~\ref{sup_subsec:add_quali}-\ref{sup_subsec:uncurated}.}
\label{fig:qualitative_comparison}
\vspace{-2.0em}
\end{center}
\end{figure*}

\begin{figure*}[!t]
\begin{center}
\centerline{\includegraphics[width=1\textwidth]{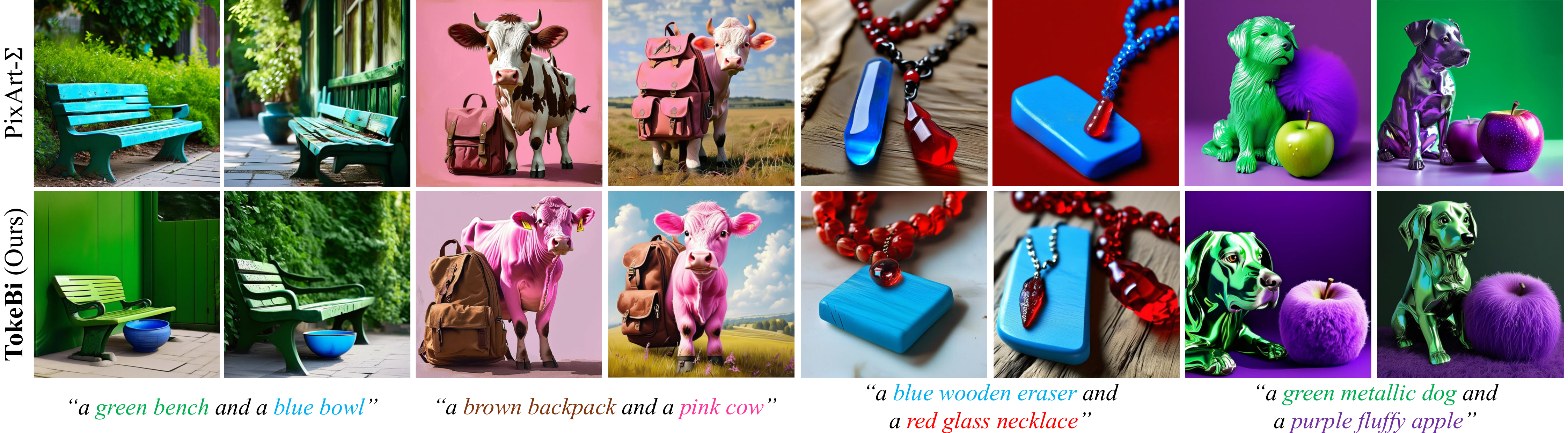}}
\vspace{-0.5em}
\caption{\textbf{Qualitative results of TokeBi with Pixart-$\mathbf{\Sigma}$ as its base model}. PixArt-$\Sigma$ (\emph{first row}) exhibits difficulties in semantic binding despite its use of a \emph{non-causal} large text encoder (T5-XXL) and advanced denoising architecture (DiT). 
TokeBi (\emph{second row}) effectively enhances semantic binding in prompts with multiple attributes per object, as exemplified by the 3$\times$3 scenario.
}
\label{fig:quali_pixart}
\vspace{-3.0em}
\end{center}
\end{figure*}

\vspace{-0.5em}
\paragraph{Datasets.}
For quantitative evaluation, we employ T2I-CompBench~\cite{huang2023t2i}, GenEval~\cite{ghosh2023geneval} color subset, and our 3$\times$3 dataset.
T2I-CompBench consists of three subsets—color, texture, and shape—each with 300 prompts.
GenEval's color attribute subset contains 100 prompts describing objects with distinct colors.
To further examine TokeBi's robustness under \emph{multiple attribute scenarios}, we construct the 3$\times$3 dataset, consisting of structured prompts with two objects, each assigned two attributes, following the template: ``a/an \{adj.\} \{adj.\} \{noun\} and a/an \{adj.\} \{adj.\} \{noun\}.''
Attributes and nouns from T2I-CompBench are used to generate 200 prompts via an LLM, resulting in 1,200 evaluation prompts. Dataset details are in Appendix Sec.~\ref{sup_sec:dataset}.

\vspace{-0.5em}
\paragraph{Metrics.} 
We evaluate semantic binding performance using five metrics: BLIP-VQA~\cite{huang2023t2i}, GenEval~\cite{ghosh2023geneval}, CIC~\cite{chen2024cat}, VQAScore~\cite{lin2024evaluating}, and VIEScore~\cite{ku2024viescore}. BLIP-VQA follows established protocols~\cite{feng2024ranni, hu2024token, hu2024ella, jiang2024comat} on the T2I-CompBench~\cite{huang2023t2i}.
GenEval assesses object-color relationships.
CIC utilizes an LLM to quantify information loss, specifically measuring the `two objects exist' scenario in Chen \textit{et al.}~\cite{chen2024cat}.
VQAScore employs CLIP~\cite{radford2021learning}.
VIEScore measures semantic consistency, following the original paper's protocol.
Evaluations are conducted using five images per prompt, totaling 6,000 images per method.
User study on perceptual quality and disentanglement (Sec.~\ref{sup_subsec:add_quanti}), details on evaluation metrics (Sec.~\ref{sup_sec:eval_metric_details}) and implementation considerations (Sec.~\ref{sup_sec:implementation_details}) are in Appendix.

\begin{figure}
    \centering
    \begin{subfigure}{0.32\linewidth}
        \includegraphics[width=\linewidth]{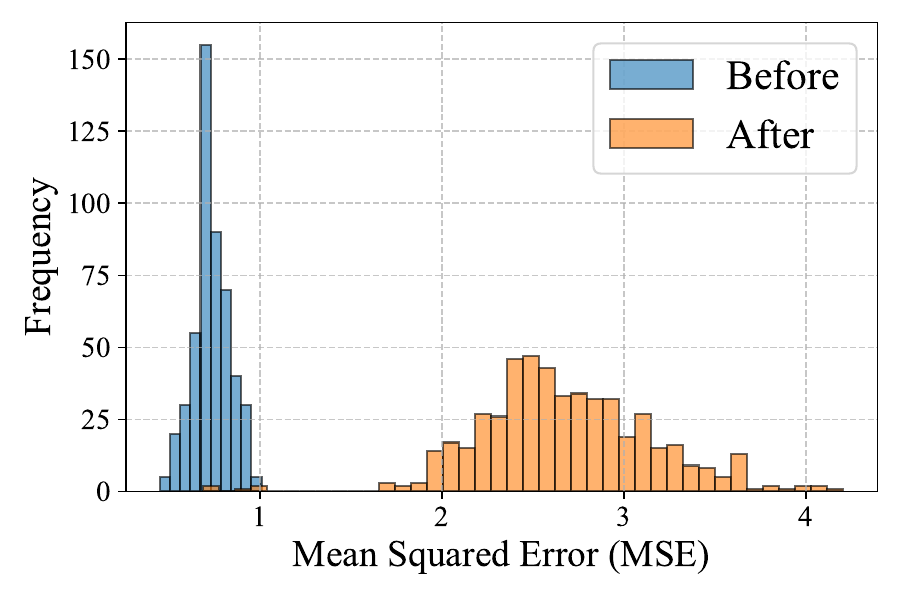}
    \end{subfigure}
    \begin{subfigure}{0.32\linewidth}
        \includegraphics[width=\linewidth]{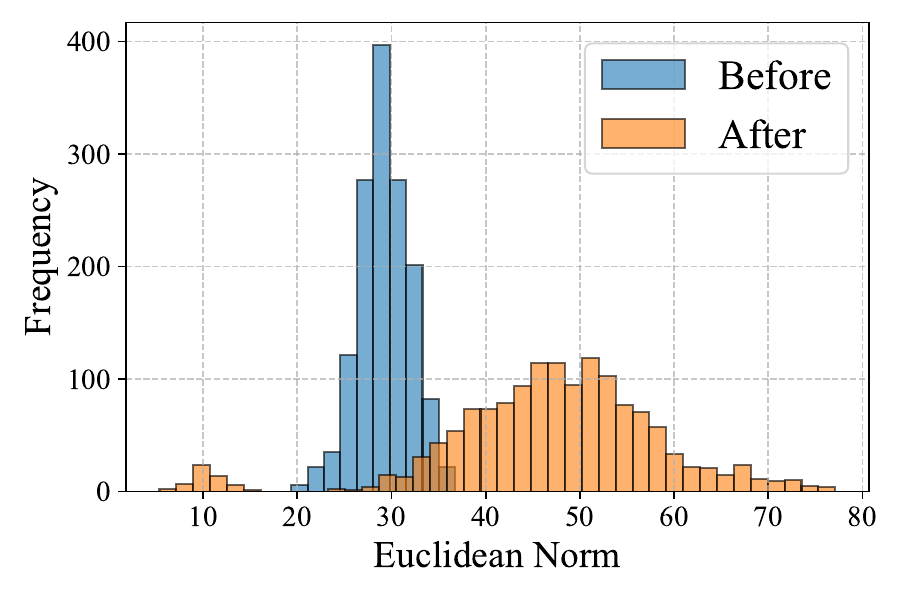}
    \end{subfigure}
    \begin{subfigure}{0.32\linewidth}
        \includegraphics[width=\linewidth]{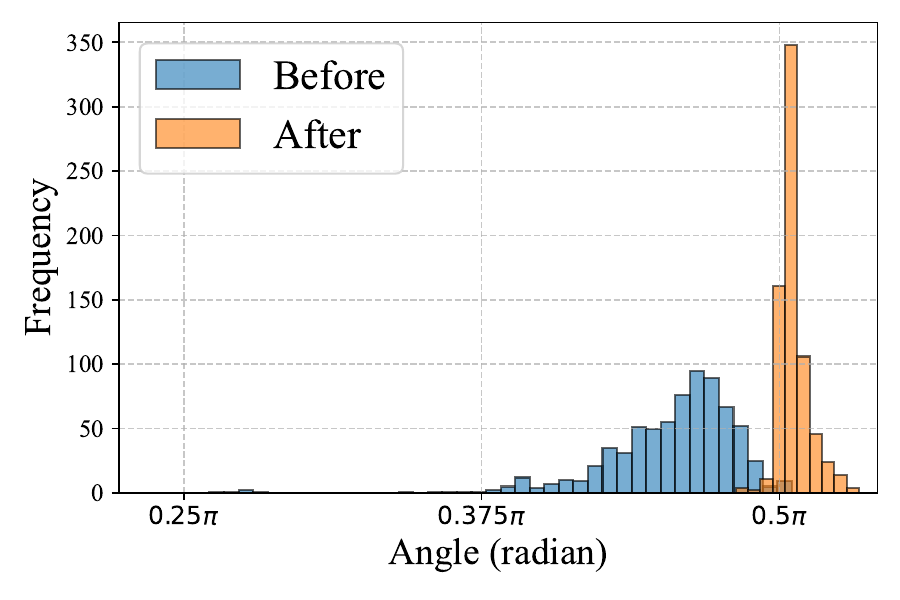}
    \end{subfigure}
    \vspace{-1.0em}
    \caption{\textbf{Analysis of the geometrical properties of text tokens before and after applying TokeBi.} Using the T2I-CompBench dataset, the mean squared error (MSE) between object tokens, the norm of each token, and the angle between the tokens are measured for both before and after applying TokeBi. 
    Note that CAPO in TokeBi orthogonalizes inter-NP tokens, resulting in angles concentrated around $\pi/2$ in the third figure.
    Wilcoxon’s signed rank test shows significant increases in MSE ($p$-value$=4.1\times 10^{-119}$), norm ($p$-value$=4.6\times 10^{-213}$), and angle ($p$-value$=1.9\times 10^{-119}$).
    }
    \label{fig:analysis}
    \vspace{-1.5em}
\end{figure}

\subsection{Comparison}
\label{subsec:comparison}

\paragraph{Quantitative comparison.}
We conduct a comprehensive quantitative comparison of various semantic binding methods using SDXL~\cite{podell2023sdxl} as the base model. Tab.~\ref{tab:2x2_quant_table} presents the results obtained from images generated by each method with various metrics. As shown, TokeBi shows superior performance across all datasets and metrics, indicating a \textbf{120.7\%} improvement in the GenEval score compared to SDXL, clearly validating its improved capabilities. We also evaluate the methods on more challenging 3$\times$3 prompts (Tab.~\ref{tab:3x3_quant_table}). The results confirm that TokeBi continues to exhibit strong performance even with complex attribute compositions, highlighting its robustness in managing intricate multi-attribute prompts. 
Furthermore, note that TokeBi demonstrates its competitiveness against existing \emph{training requiring} methods~\cite{feng2024ranni,hu2024ella,jiang2024comat, zarei2024understanding}, showcasing comparable and, on average, superior BLIP-VQA~\cite{huang2023t2i} scores (Sec.~\ref{sup_subsec:add_quanti}).
Additional experiments, including computational cost, (Sec.~\ref{sup_sec:efficiency}), alternative base models (Sec.~\ref{sup_subsec:add_quanti}), and user study results (Sec.~\ref{sup_subsec:add_quanti}), are in Appendix.

\vspace{-0.5em}
\paragraph{Qualitative comparison.}
We conduct qualitative comparisons between TokeBi and existing semantic binding methods across multiple T2I models, including SDXL and PlayG-v2, as shown in Fig.~\ref{fig:qualitative_comparison}.
TokeBi outperforms in semantic binding, both for simple and complex scenarios with multiple attributes and objects.
We also extend our experiments to PixArt-$\Sigma$, which integrates the \emph{non-causal} text encoder T5 and DiT architecture, with results shown in Fig.~\ref{fig:quali_pixart}.
These additional results further validate TokeBi's generalizability and robustness across diverse model architectures.
Additional qualitative results, including \emph{uncurated samples}, are provided in Appendix Sec.~\ref{sup_subsec:add_quali}-\ref{sup_subsec:uncurated}.

\subsection{Discussion}
To validate the effectiveness of TokeBi, we performed additional analysis and ablation studies with a subset of T2I-CompBench using SDXL. Additional ablation studies, including qualitative results for each configuration in Tab.~\ref{tab:ablation}, are provided in Appendix (Sec.~\ref{sup_sec:ablation}).
The limitations (Sec.~\ref{sup_sec:limitations}) and broader impacts (Sec.~\ref{sup_sec:broader_impacts}) are also detailed in Appendix.

\vspace{-0.7em}
\paragraph{Effectiveness of TokeBi on text token geometry.}
We propose TokeBi to enhance semantic binding, grounded in exhaustive analysis of how text token geometry influences CA maps. To assess the impact of TokeBi on token geometry, we examine changes in MSE, angles, and norms in Fig.~\ref{fig:analysis}. All metrics show increasing trends, consistent with our theory and supporting TokeBi’s efficacy.

\vspace{-0.6em}
\paragraph{Effectiveness of CAPO.}
\vspace{-0.2em}

\begin{wraptable}{r}{0.46\textwidth}
\centering
\vspace{-0.8em}
\caption{\textbf{Ablation study on the T2I-CompBench benchmark.} 
Ablation studies on CAPO and ATM confirm their strong individual contributions to the task.
The latent vectors are optimized only when ATM is applied.
}
\label{tab:ablation}
\renewcommand{\arraystretch}{1.0} 
\setlength{\tabcolsep}{0.25em} 
\resizebox{0.46\textwidth}{!}{%
\begin{tabular}{c|c|cc|cccc}
\toprule
\multirow{2}{*}{Conf.} & \multirow{2}{*}{\tabincell{c}{CAPO}} & \multicolumn{2}{c|}{\tabincell{c}{ATM}} & \multicolumn{4}{c}{BLIP-VQA~\cite{huang2023t2i} $\uparrow$} \\
& & $\mathcal{L}_{Ent.}$. & $\mathcal{L}_{Bhat.}$ & Color & Texture & Shape & Avg. \\ 
\midrule
A & \xmarkg & \xmarkg & \xmarkg & 0.5616 & 0.6446 & 0.5712 & 0.5925 \\
B & \cmarkg & \xmarkg & \xmarkg & 0.5900 & 0.6954 & 0.5910 & 0.6255 \\
C & \xmarkg & \cmarkg & \xmarkg & 0.7346 & 0.7796 & 0.6323 & 0.7155 \\ 
D & \xmarkg & \cmarkg & \cmarkg & \rednum{0.7689} & \lightorangenum{0.7798} & \orangenum{0.6472} & \lightorangenum{0.7320} \\ 
E & \cmarkg & \cmarkg & \xmarkg & \orangenum{0.7641} & \orangenum{0.8000} & \lightorangenum{0.6415} & \orangenum{0.7352} \\ 
F & \cmarkg & \cmarkg & \cmarkg & \lightorangenum{0.7576} & \rednum{0.8041} & \rednum{0.6866} & \rednum{0.7494} \\
\bottomrule
\end{tabular}
}
\vspace{-1em}
\end{wraptable}

To validate CAPO's effectiveness, we compare semantic binding performance with and without CAPO, as summarized in Tab.~\ref{tab:ablation}. The significant improvement from Conf. A (baseline) to Conf. B (with CAPO only) confirms CAPO's capability to reduce entanglements and strengthen semantic binding. Further enhancements observed between Conf. D (without CAPO) and Conf. F (Ours) reinforce CAPO's efficacy.

\vspace{-0.8em}
\paragraph{Effectiveness of ATM with our loss design.}
In Tab.~\ref{tab:ablation}, comparing Conf. A (baseline) and Conf. C ($\mathcal{L}_{\text{Ent.}}$ only) shows that entropy loss significantly increases performance. Adding the $\mathcal{L}_{\text{Bhat.}}$, which forms our total loss design (Conf. D), further improves performance. Optimizing the mixing matrix and latent space with our loss after applying CAPO (Conf. F, Ours) yields the best results.

\section{Conclusion}
\label{sec:conclusion}
We propose TokeBi, a training-free text embedding-aware attention regularization method for strong semantic binding in T2I generation involving multiple objects and attributes.
Our theoretical and empirical analyses identified that the geometric properties of text token embeddings (\textit{i.e.}, norm and angular distance) are crucial factors for explaining semantic binding.
Our theoretical findings led to TokeBi's Causality-Aware Projection Out (CAPO) and Adaptive Token Mixing (ATM) with a unique loss design that condenses intra-NP attention and separates inter-NP attention.
Extensive experiments demonstrate TokeBi’s superior performance across diverse datasets and metrics, underscoring its effectiveness and scalability in T2I synthesis with strong semantic binding.

\section*{Acknowledgements}
This work was supported in part by Institute of Information \& communications Technology Planning \& Evaluation (IITP) grant funded by the Korea government (MSIT) [NO. RS-2021-II211343, Artificial Intelligence Graduate School Program (Seoul National University)] and National Research Foundation of Korea (NRF) grant funded by the Korea government (MSIT) (No. NRF-2022M3C1A309202211). Also, the authors acknowledged the financial support from the BK21 FOUR program of the Education and Research Program for Future ICT Pioneers, Seoul National University.

{\small
 \bibliographystyle{plain}
 \bibliography{neurips_2025}
}


\appendix

\renewcommand{\thefigure}{S\arabic{figure}}
\renewcommand{\thetable}{S\arabic{table}}
\renewcommand{\thesection}{S\arabic{section}}
\renewcommand{\theequation}{S\arabic{equation}}

\clearpage
\setcounter{section}{0}
\setcounter{figure}{0}
\setcounter{table}{0}
\setcounter{theorem}{0}
\setcounter{remark}{0}
\setcounter{equation}{0}
\setcounter{proposition}{0}


\section*{\LARGE{Appendix}}

\section{Proofs}
\label{sup_sec:proofs}
\subsection{Proof of Proposition 1}
\label{sup_subsec:proof_proposition_1}
\renewcommand{\thetheorem}{1}
\begin{proposition}[KL Divergence Monotonicity in Distance]
    Let $Q=[q_1; \cdots; q_N]$ where $q_a$ is the $a$-th query token corresponds to $a$-th token embedding $t_a$. Then, increasing the distance between text token embeddings $\|t_i - t_j\|^2$ monotonically increases the Kullback-Leibler divergences $D_{KL}(A_i\|A_j)$.   
\end{proposition}

\begin{proof}
    Due to the complex representations from the softmax, we adopt a perturbation approach when $t_i \approx t_j$. Specifically, we examine how the Kullback--Leibler (KL) divergence $D_{\mathrm{KL}}(A_i \| A_j)$ changes with respect to $\|\delta t\|^2$ under a second-order approximation, where $t_j = t_i + \delta t$.
    For ease of notation, we denote $P_{ai}$ as the attention value for $a$-th query token $q_a$ and $i$-th text token $t_i$:
    \begin{equation}
        P_{ai} = \frac{\exp(q_aW_K^\top t_i^\top)}{Z_a}, \quad
        Z_a=\sum_{l=1}^{L}\exp(q_aW_K^\top t_l^\top).
        \label{eq:supp_def_P_Z}
    \end{equation}
    We also define the normalized attention map $A_{i}$ across the query tokens as:
    \begin{equation}
        A_i=\frac{P_i}{\|P_i\|_1},
        \quad A_{ai}=\frac{P_{ai}}{S_i}, 
        \quad S_i = \sum_{b=1}^{N}P_{bi} = \sum_{b=1}^N \frac{\exp(q_bW_K^\top t_i^\top)}{Z_b}.
        \label{eq:supp_def_A_S}
    \end{equation}
    The Kullback-Leibler (KL) divergence $D_{KL}$ of $A_i$ and $A_j$ is then defined as:
    \begin{equation}
        D_{KL}(A_i\|A_j)
        = \sum_{a=1}^{N}A_{ai}\log\frac{A_{ai}}{A_{aj}} \\
        = \mathbb{E}_{A_{ai}} \left[ \log\frac{A_{ai}}{A_{aj}} \right].
        \label{eq:def_kl}
    \end{equation}
    Since $A_{aj}$ is dependent on $\delta t$ and $A_{aj} = A_{ai}$ for $\delta t = 0$, we derive $D_{\mathrm{KL}}(A_i \| A_j)$ using a second-order approximation with respect to $t$ as:
    \begin{align}
        D_{KL}(A_i\|A_j) 
        & = \mathbb{E}_{A_{ai}} \left[ \log \frac{A_{ai}}{A_{aj}} \right] \\
        & \approx \mathbb{E}_{A_{ai}} \left[ \log A_{ai} - \left(\log A_{ai} + \nabla_{t} \log A_{ai} \delta t^{\top} + \frac{1}{2} \delta t \nabla_{t}^2 \log A_{ai} \delta t^T \right) \right] \\
        & = \mathbb{E}_{A_{ai}} \left[ -\frac{1}{2} \delta t \nabla_{t}^2 \log A_{ai} \delta t^T \right],
        \label{eq:supp_kl_interm}
    \end{align}
    where we use the fact that $\mathbb{E}_{A_{ai}} \left[ \nabla_{t} \log A_{ai} \right] = 0$. Thus, we need to derive the explicit form of $\mathbb{E}_{A_{ai}} \left[ -\frac{1}{2} \delta t \nabla_{t}^2 \log A_{ai} \delta t^T \right]$. Since $ \log A_{ai} = \log P_{ai} - \log S_i$, $\nabla_{t} \log P_{ai}$ and $\nabla_{t} \log S_i$ are:
    \begin{align}
        \nabla_{t} \log P_{ai} = \nabla_{t} \left( q_a W_K^\top t_i^\top \right) = q_a W_K^\top,
    \end{align}
    \begin{align}
        \nabla_{t} \log S_i = \frac{1}{S_i} \nabla_{t} S_i = \frac{1}{S_i} \sum_{b=1}^{N} \frac{1}{Z_b} \exp(q_b W_K^\top t_i^\top) q_b W_K^\top = \sum_{b=1}^{N} \frac{P_{bi}}{S_i} q_b W_K^\top.
    \end{align}
    Therefore, we can represent $\nabla_{t} \log A_{ai}$ as:
    \begin{align}
        \nabla_{t} \log A_{ai} = q_a W_K^\top - \sum_{b=1}^{N} A_{bi} q_b W_K^\top.
        \label{eq:supp_first_order}
    \end{align}
    We now derive the explicit form of the Hessian $-\frac{1}{2} \delta t \nabla_{t}^2 \log A_{ai} \delta t^T$. From Eq.~\ref{eq:supp_first_order}:
    \begin{align}
        \nabla^2_{t} \log A_{ai} = - \nabla_{t} \left( \sum_{b=1}^{N} A_{bi} q_b W_K^\top \right), \quad
        A_{bi} = \frac{P_{bi}}{S_i} = \frac{1}{Z_b} \cdot \frac{\exp(q_b W_K^\top t_i^\top)}{S_i}.
        \label{eq:supp_hessian_interm}
    \end{align}
    Therefore, we can represent $\nabla_{t} A_{bi}$ as:
    \begin{align}
        \nabla_{t} A_{bi} = A_{bi} \left( q_b W_K^\top - \sum_{c=1}^{N} A_{ci} q_c W_K^\top \right).
        \label{eq:supp_gradient_A}
    \end{align}
    By substituting Eq.~\ref{eq:supp_gradient_A} into Eq.~\ref{eq:supp_hessian_interm}, we can derive $\nabla^2_{t} \log A_{ai}$ as follows:
    \begin{align}
        \nabla^2_{t} \log A_{ai} &= - \sum_{b=1}^{N} A_{bi} \left( q_b W_K^\top - \sum_{c=1}^{N} A_{ci} q_c W_K^\top \right)^\top \left( q_b W_K^\top - \sum_{c=1}^{N} A_{ci} q_c W_K^\top \right) \nonumber \\
        &= W_K \Sigma_{A_{i}} W_K^\top,
    \end{align}
    where $\Sigma_{A_{i}} = \sum_{b=1}^{N} A_{bi} (q_b - \bar{q}_{A_i})^\top (q_b - \bar{q}_{A_i})$ is a positive-definite matrix and $\bar{q}_{A_i} = \sum_{b=1}^{N} A_{bi} q_b$. Since it is constant across query index $a$, we finally derive $D_{KL}(A_i\|A_j)$ as:
    \begin{align}
        D_{KL}(A_i\|A_j) 
        & = \mathbb{E}_{A_{ai}} \left[ -\frac{1}{2} \delta t \nabla_{t}^2 \log A_{ai} \delta t^T \right] \\
        & = \frac{1}{2} \delta t W_K \Sigma_{A_{i}} W_K^\top \delta t^\top
        \label{eq:supp_final_kl}
    \end{align}
    We now express the distance between $t_i$ and $t_j$ with $\delta t$,
    \begin{equation}
        \|t_i-t_j\|_2^2=\delta t \delta t ^\top.
    \end{equation}
    From Eq.~\ref{eq:supp_final_kl}, we have shown: 
    \begin{equation}
         D_{KL}(A_i\|A_j) \propto\delta t W_K \Sigma_A W_K^\top\delta t^\top,
    \end{equation}
    Since $W_K \Sigma_{A_{i}} W_K^\top$ is positive definite, $D_{KL}(A_i \| A_j)$ increases as $\|t_i - t_j\|_2^2$ increases. Beyond small perturbations, $D_{KL}(A_i\|A_j)$ remains convex in $A_j$ and continues to increase as $A_j$ deviates further from $A_i$, driven primarily by the growth in $\|t_i - t_j\|_2^2$, aligning with the quadratic approximation.
\end{proof}

\subsection{Proof of Proposition 2}
\label{sup_subsec:proof_proposition_2}

\renewcommand{\thetheorem}{2}
\begin{proposition}
Let two token embeddings $t_i,\;t_j\in \mathbb{R}^d$ follow the assumptions 1) $\|t_i\|\approx\|t_j\|=r$ and 2) $\cos\theta_{ij} < 0.5$ where $\cos\theta_{ij}=(t_it_j^\top/(\|t_i\|\|t_j\|))$. With scalar values $\lambda_i,\;\lambda_j > 1$, the following holds.
\[
     \|t_i-\;t_j\|_2^2 < \|\lambda_i t_i-\lambda_jt_j\|_2^2.
\]
\end{proposition}
\begin{proof}
    Then with the assumption 1),
\begin{equation}
    \begin{aligned}
        &\|\lambda_i t_i - \lambda_j t_j\|_2^2 - \|t_i - t_j\|_2^2\\
        &\quad= r^2 \big( \lambda_i^2 + \lambda_j^2 + 2(1 - \lambda_i \lambda_j) \cos\theta_{ij} - 2 \big).
    \end{aligned}
\end{equation}
    Since $\lambda_i,\;\lambda_j > 1$ and with the assumption 2),
\begin{equation}
    \begin{aligned}
    &r^2 \big( \lambda_i^2 + \lambda_j^2 + 2(1 - \lambda_i \lambda_j) \cos\theta_{ij} - 2 \big) \\
    &\quad > r^2 \big( \lambda_i^2 + \lambda_j^2 - \lambda_i \lambda_j - 1 \big) \\
    &\quad > 0.
    \end{aligned}
\end{equation}
    Therefore, we have proven the following inequality.  
\begin{equation}
    \|\lambda_i t_i - \lambda_j t_j\|_2^2 - \|t_i - t_j\|_2^2 > 0.
\end{equation}
\end{proof}

\subsection{Proof of Remark 1}
\label{sup_subsec:proof_remark_1}

\renewcommand{\theremark}{1}
\begin{remark}[Norm of token embedding add \& sub]
Suppose that two token embeddings \( t_i, t_j \in \mathbb{R}^d \) 
follow \( t_i, t_j \sim \mathcal{N}(\mu_d, \Sigma_d) \) and they satisfy $\|\mu_d\|_2 \ll \|t_i\|_2, \|t_j\|_2$. Then:
    \[
        \mathbb{E}[\|t_i\|_2],\ \mathbb{E}[\|t_j\|_2] \le \mathbb{E}[\|t_i\pm t_j\|_2].
    \]
\end{remark}
\begin{proof} Expanding the squared norm,
    \begin{equation}
        \|t_i \pm t_j\|_2^2 = \|t_i\|_2^2 + \|t_j\|_2^2 \pm 2 t_i \cdot t_j.
    \end{equation}
    Taking expectations and using linearity,
    \begin{equation}
        \mathbb{E}[\|t_i \pm t_j\|_2^2] = \mathbb{E}[\|t_i\|_2^2] + \mathbb{E}[\|t_j\|_2^2] \pm 2 \mathbb{E}[t_i \cdot t_j].
    \end{equation}
    Since \( t_i \) and \( t_j \) follow $\mathcal{N}(\mu_d, \Sigma_d)$, their inner product satisfies \( \mathbb{E}[t_i \cdot t_j] = \|\mu_d\|^2_2 \) and $\|\mu_d\|_2^2 \ll \|t_i\|^2_2, \|t_j\|^2_2$, simplifying to:
    \begin{align}
        \mathbb{E}[\|t_i \pm t_j\|_2^2] &= \mathbb{E}[\|t_i\|_2^2] + \mathbb{E}[\|t_j\|_2^2] \pm 2\|\mu_d\|^2_2 \\
        &\approx \mathbb{E}[\|t_i\|_2^2] + \mathbb{E}[\|t_j\|_2^2].
    \end{align}
    Applying Jensen’s inequality to the square root function,
    \begin{equation}
        \mathbb{E}[\|t_i \pm t_j\|_2] \geq \sqrt{\mathbb{E}[\|t_i\|_2^2] + \mathbb{E}[\|t_j\|_2^2]}.
    \end{equation}
    Since \( \mathbb{E}[\|t_i\|_2] \leq \sqrt{\mathbb{E}[\|t_i\|_2^2]} \) by Jensen's inequality, it follows that:
    \begin{equation}
        \mathbb{E}[\|t_i\|_2], \mathbb{E}[\|t_j\|_2] \leq \mathbb{E}[\|t_i \pm t_j\|_2].
    \end{equation}
    Equality holds when either \( t_i = 0 \) or \( t_j = 0 \).
\end{proof}

\section{Additional Analysis}
\label{sup_sec:additional_analysis}

\subsection{Verification of Assumptions}
\label{sup_subsec:verification_of_assumptions}

\begin{figure*}[h]
  \centering
  \begin{subfigure}{0.32\linewidth}
    \includegraphics[width=\linewidth]{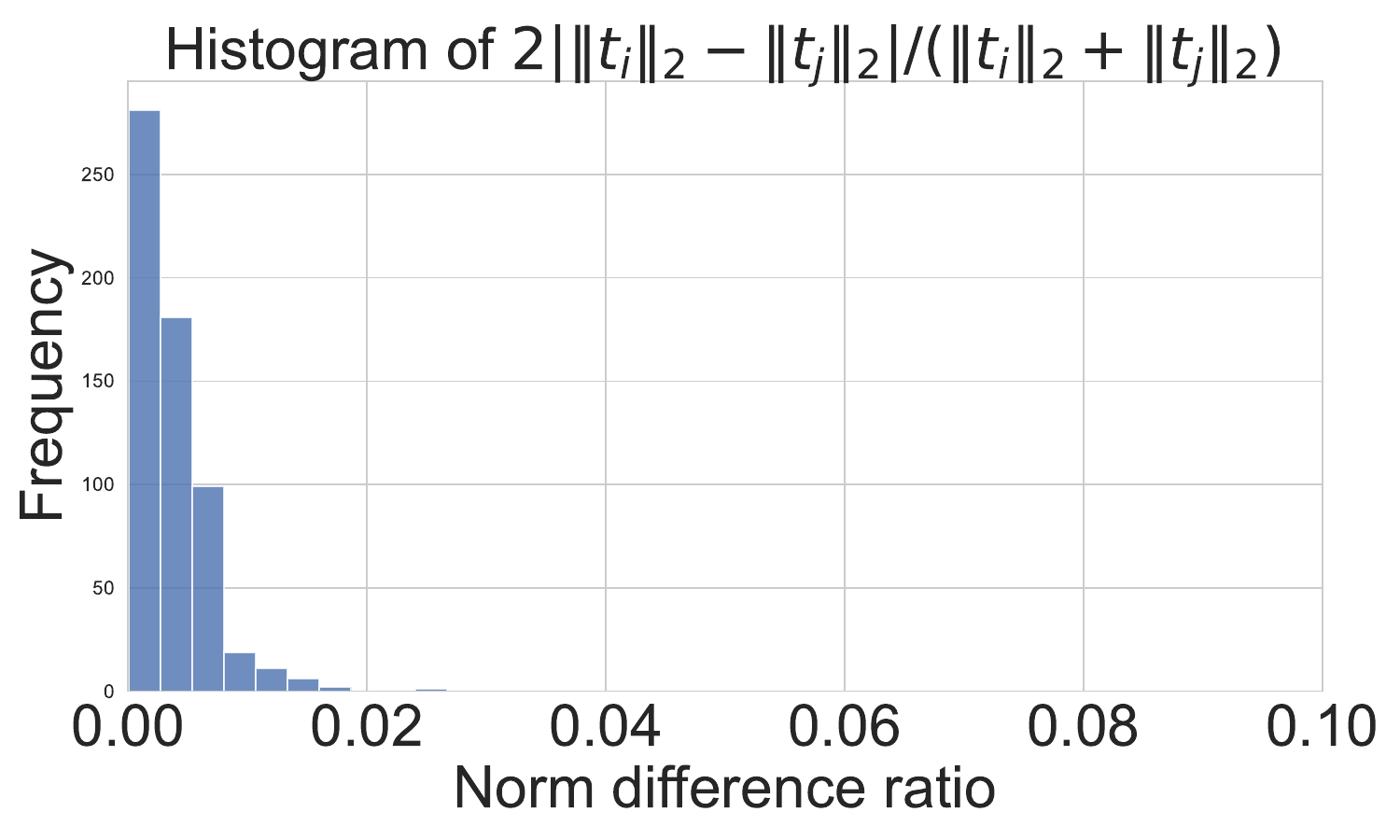}
    \caption{Histogram of normalized token difference.}
    \label{sup_fig:assumption_diff_norm}
  \end{subfigure}
  \begin{subfigure}{0.32\linewidth}
    \includegraphics[width=\linewidth]{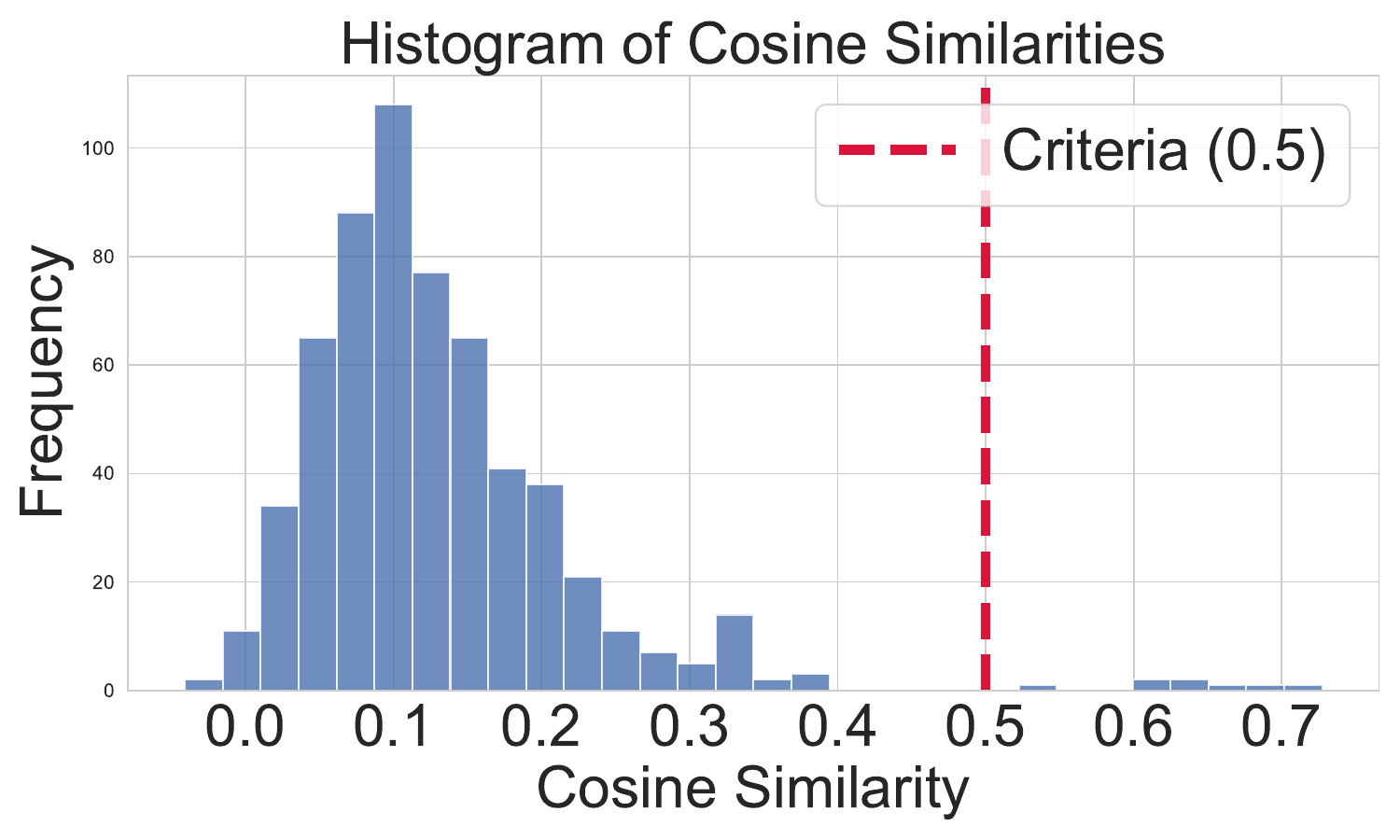}
    \caption{Histogram of $\cos\theta_{ij}$ between tokens.}
    \label{sup_fig:assumption_cossim}
  \end{subfigure}
  \begin{subfigure}{0.32\linewidth}
    \includegraphics[width=\linewidth]{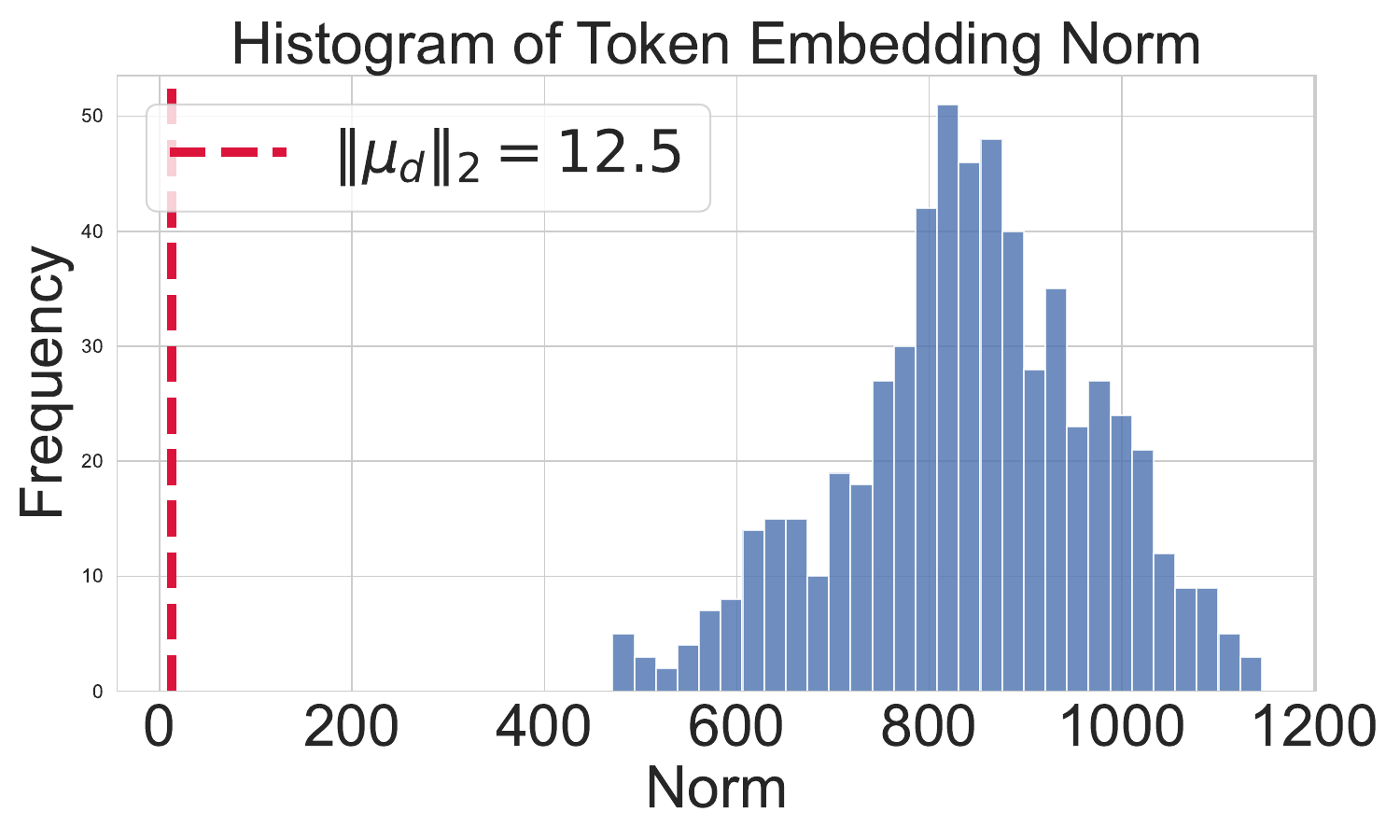}
    \caption{Histogram of token embedding and $\|\mu_d\|_2$.}
    \label{sup_fig:assumption_mu_d}
  \end{subfigure}
  \caption{\textbf{(a)} Histogram of the norm differences and norm ratios. Most values are concentrated around 0, and the largest value is below 0.02, indicating that norm differences between token embedding pairs are extremely small. \textbf{(b)} Histogram of cosine similarity between object token embeddings located in different noun phrases (NPs) within the text prompt set for BLIP-VQA. It can be observed that most token embedding pairs have similarities lower than 0.5. \textbf{(c)} Comparison between the histogram of norms of token embeddings and $\|\mu_d\|_2$. While $\|\mu_d\|_2$ is relatively small (12.5), most token embeddings exhibit norms greater than 400, confirming that the assumption $\|\mu_d\|_2 \ll \|t_i\|_2, \|t_j\|_2$ holds true for the CLIP text encoder.}
  \label{sup_fig:verification_of_assumptions}
  \vspace{-1em}
\end{figure*}

\paragraph{Assumption of norm difference in Proposition 2.}
In Proposition 2, we assumed that the norms of two token embeddings, $t_i$ and $t_j$, are approximately similar. To empirically validate this assumption, we conducted experiments on 600 prompts sampled from the T2I Compbench dataset used in BLIP-VQA. Specifically, we quantified the relative norm difference by computing the ratio $2|\|t_i\|_2 - \|t_j\|_2|/(\|t_i\|_2 + \|t_j\|_2)$ for each object token embedding pair present in these prompts. The results are depicted in Fig.~\ref{sup_fig:assumption_diff_norm}. As shown in the figure, most computed ratios are close to zero, indicating that the differences in norm between embedding pairs are minimal relative to their magnitudes. These results strongly support the validity of our assumption regarding the similarity of the embedding norms.

\paragraph{Assumption of cosine similarity in Proposition 2.}
In Proposition 2, we assumed that $\cos\theta_{ij}=\frac{t_i t_j^\top}{|t_i||t_j|} < 0.5$. In this section, we present experimental results validating the feasibility of this assumption. Fig.~\ref{sup_fig:assumption_cossim} illustrates a histogram of $\cos\theta_{ij}$ values computed between object token embeddings $t_i$ and $t_j$, corresponding to different NPs across 600 prompts from the BLIP-VQA dataset. As seen in the figure, the majority of token embedding pairs satisfy the stated condition. The few prompts that violate this condition typically correspond to cases where both tokens represent the same object category (\textit{e.g.}, ``a plastic chair and a wooden chair''). Since such cases are rare, these results demonstrate the validity of our assumption.

\paragraph{Assumption of norm in Remark 1}
Text encoders utilized in TokeBi enforce a Gaussian distribution through a layer normalization module before producing final embeddings. In the case of the T5 encoder~\cite{raffel2020exploring}, due to the absence of bias terms in layer norm, two token embeddings $t_i$ and $t_j$ naturally satisfy $\mathbb{E}[t_i \cdot t_j]=0$, making the remark hold, without requiring specific constraints on $\mu_d$. However, the CLIP text encoder~\cite{radford2021learning} incorporates bias terms during the layer normalization step, causing $\mu_d$ to deviate from the zero vector. Hence, we analyze whether the embeddings extracted from the CLIP text encoder satisfy the assumption stated in Remark 1, namely, $\|\mu_d\|_2 \ll \|t_i\|^2, \|t_j\|^2$. The results of this analysis in 600 prompts from the T2I Compbench dataset are presented in Fig.~\ref{sup_fig:assumption_mu_d}. As shown in the figure, the norm $\|\mu_d\|_2$ is approximately $12.5$, significantly smaller compared to the norms of individual embeddings. This result supports the compatibility of the CLIP text encoder embeddings with our assumption.

\subsection{Token Embedding Scaling in Baselines}
\label{subsec:token_embedding_scaling_in_baselines}

\paragraph{Prompt-to-prompt.}
Prompt-to-prompt~\cite{hertz2023prompt} (P2P) performs image editing by controlling the degree of attribute expression in the generated image through cross-attention re-weighting. This can be formulated as follows:
\begin{align}
    \text{AttentionRe}&\text{-weight}= \alpha_i\cdot\text{softmax}\Big(\frac{QK^\top}{\sqrt{d_k}}\Big)V \nonumber \\
    &\approx \text{softmax}\Big(\frac{Q(\alpha_i\cdot K)^\top}{\sqrt{d_k}}\Big)(\alpha_i\cdot V),
\end{align}
where $\alpha_i$ is a vector whose elements are set to a user-specified scalar value at the token index $i$ intended for re-weighting, with all other elements set to 1. This approach is equivalent to scaling the value matrix $V$. Since the vectors used as inputs for key and value projections remain identical as text embeddings, and the projections for keys and values are linear, this operation can ultimately be interpreted as scaling the original text embeddings.

\paragraph{Magnet.}
Magnet~\cite{zhuang2024magnet} obtains an embedding vector $\hat{e}_i$, used for Text-to-Image (T2I) synthesis, by linearly combining the original embedding $e_i$ with positive ($v^{pos}_i$) and negative embedding vectors ($v^{neg}_i$) derived from a pre-defined dataset. This procedure can be formulated as follows:
\begin{equation}
    \hat{e}_i = e_i + \alpha_i\cdot v^{pos}_i-\beta_i\cdot v^{neg}_i.
\end{equation}
According to Remark 1, this approach yields results exactly equivalent to scaling the embedding of the $i$-th token, thereby improving semantic binding performance.


\paragraph{ToMe.}
One of the key approaches of ToMe~\cite{hu2024token} is leveraging semantic additivity to create a single token embedding representing a noun phrase (NP) by linearly combining tokens corresponding to attributes and objects within that NP. Given an NP token vector set $\mathcal{U}=\{u_1, u_2, \dots, u_n\}$, where $u_1$ denotes the object token embedding, ToMe computes the merged token embedding $u_{mer.}$ as follows:
\begin{equation}
  u_{mer.} = \alpha\cdot u_1 +\beta\cdot\sum_{i=2}^nu_i , 
\end{equation}
where $\alpha$ and $\beta$ are scalar values greater than 1, fixed as $\alpha=1.1$ and $\beta=1.2$ in their implementation. According to Remark 1, this also results in an effect of increasing the norm of the token embedding.

\begin{figure*}[!t]
  \centering
  \begin{subfigure}{0.38\linewidth}
    \includegraphics[width=\linewidth]{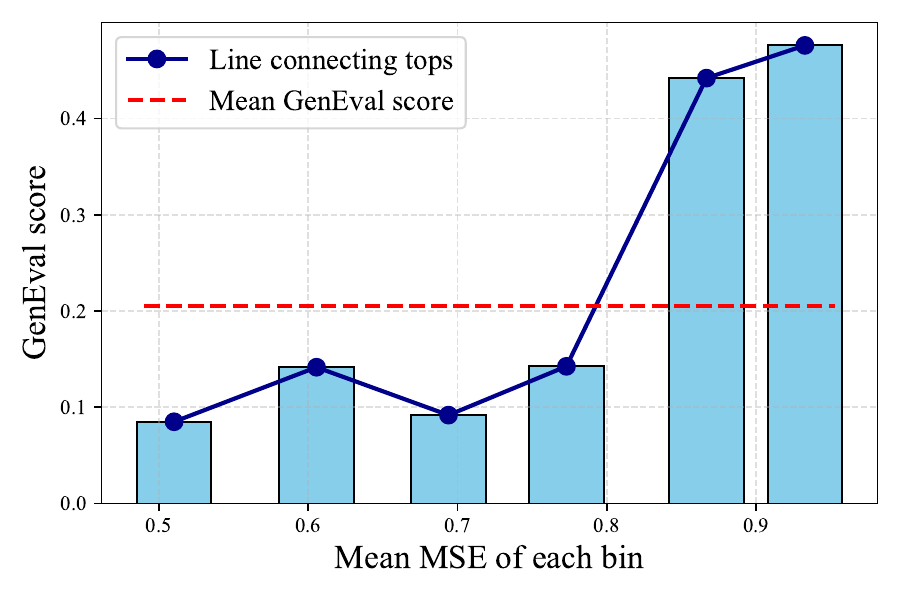}
    \caption{MSE vs GenEval score.}
    \label{sup_fig:mse_vs_geneval}
  \end{subfigure}
  \begin{subfigure}{0.6\linewidth}
    \includegraphics[width=\linewidth]{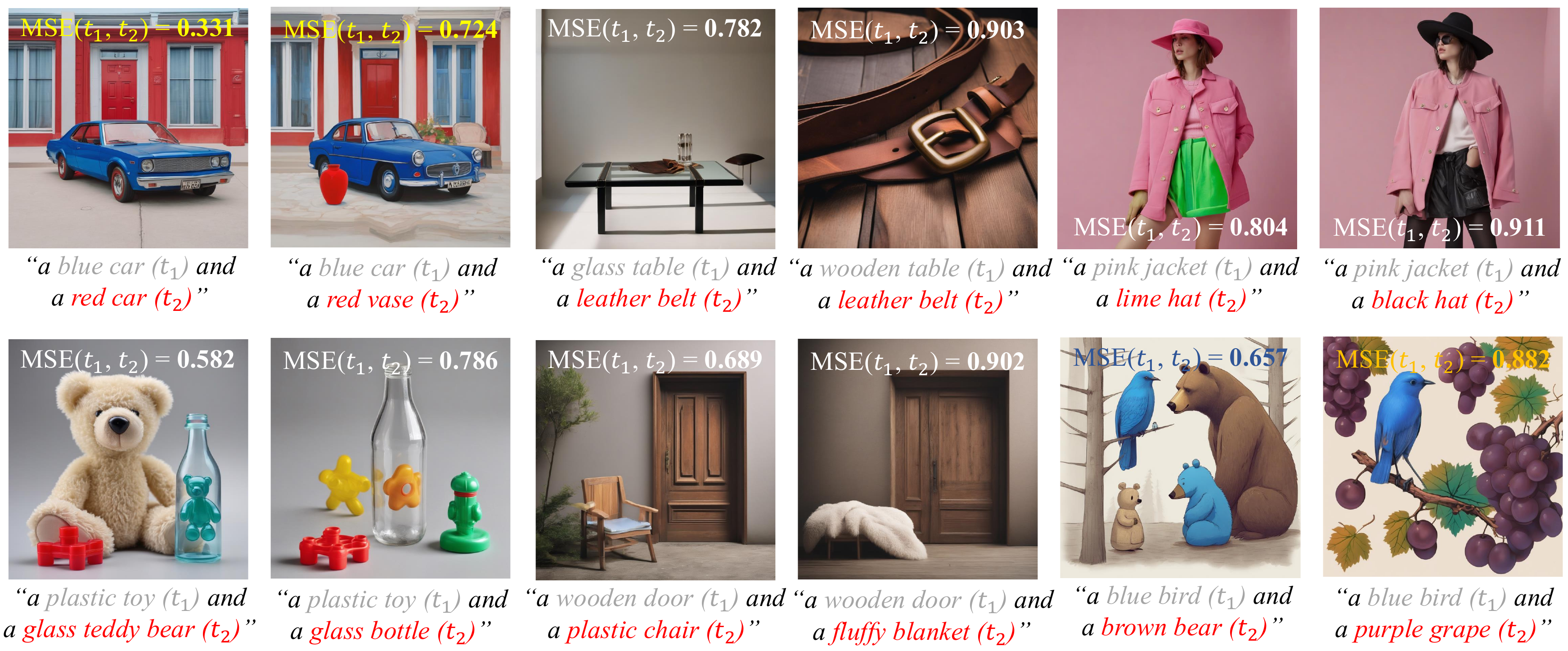}
    \caption{Results on relationship between MSE and semantic binding.}
    \label{sup_fig:mse_analysis_sub}
  \end{subfigure}
  \caption{\textbf{(a)} Relationship between Mean-Squared Error (MSE) and GenEval score. We observe a clear trend in which higher MSE values are associated with higher GenEval scores qualitatively and quantitatively. \textbf{(b)} Additional qualitative results illustrating the impact of MSE on semantic binding performance. Consistent with the main manuscript, it can be observed that smaller MSE between token embeddings correlate with increased difficulty in achieving proper semantic binding.}
  \label{sup_fig:mse_analysis}
  \vspace{-1em}
\end{figure*}

\subsection{Additional Analysis Results}
\label{sup_subsec:additional_analysis_results}

\paragraph{Effect of MSE in semantic binding.}


In the main paper, we theoretically and empirically demonstrated the impact of the distance between tokens belonging to distinct noun phrases (NPs) on semantic binding. In this section, we extend these experiments to investigate whether similar trends and outcomes hold true for a broader set of prompts as shown in Fig.~\ref{sup_fig:mse_analysis}. 

We examine the statistical relationship between mean-squared error (MSE) and semantic binding performance, we generated five images for each of 100 GenEval color attribute prompts and computed the GenEval score for each image. Since the GenEval score for the color attribute task is discretized into five values (0, 0.25, 0.5, 0.75, 1), we grouped the data into six equal-sized bins based on MSE. We then computed the average MSE and the corresponding average GenEval score for each bin to analyze the trend, as illustrated in Fig.~\ref{sup_fig:mse_vs_geneval}. Beyond this qualitative analysis, we assessed statistical significance using Spearman’s rank correlation test, which yielded a coefficient of $\rho=0.95$ and a $p$-value of $0.005$, indicating a statistically significance ($p < 0.05$).

Furthermore, we present more results showing relationship between MSE and semantic binding in Fig.~\ref{sup_fig:mse_analysis_sub}. A higher MSE between two tokens clearly correlates with more distinct cross-attention maps and improved semantic binding. 

These results align well with Proposition 1 introduced in the main text, further substantiating the motivation behind Causality-Aware Projection Out (CAPO) and Adaptive Token Mixing (ATM).

\begin{figure}[!b]
\vspace{-1.0em}
\begin{center}
\centerline{\includegraphics[width=\linewidth]{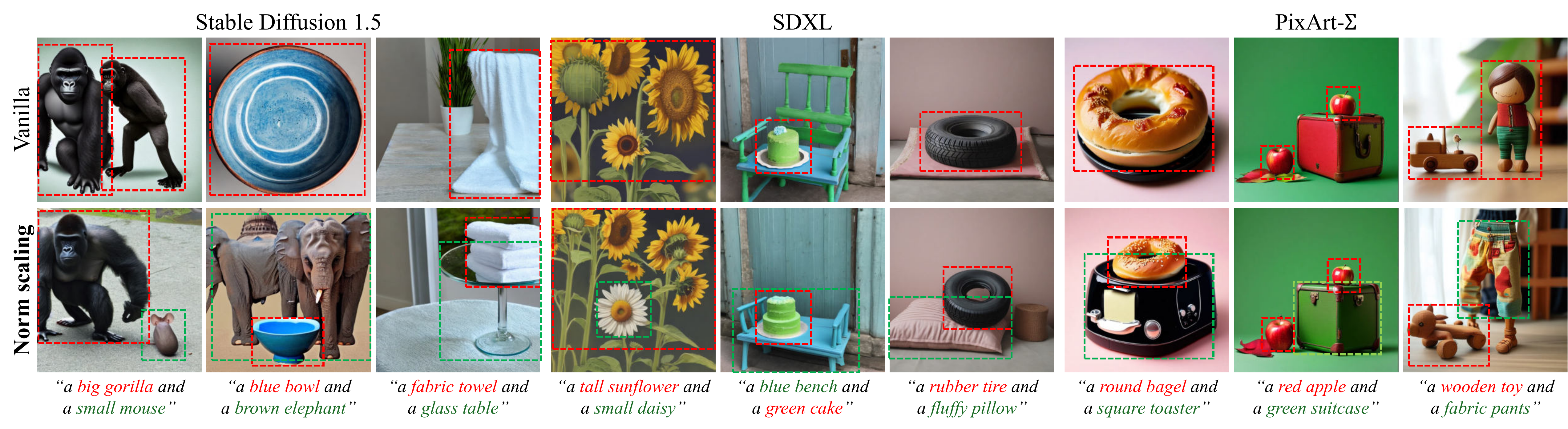}}
\vspace{-0.5em}
\caption{\textbf{Additional qualitative results further illustrate the effectiveness of norm scaling on semantic binding.} Beyond the examples presented in the main manuscript, we observe consistent effects across various models, including Stable Diffusion 1.5, SDXL, and PixArt-$\Sigma$, and across diverse prompts. Notably, even in cases of semantic leakage (\textit{e.g.}, ``a blue bench,'' ``a green suitcase'') norm scaling effectively mitigates the leakage, reinforcing its role in improving semantic binding.}
\label{sup_fig:analysis_norm}
\end{center}
\end{figure}

\paragraph{Norm scaling.}
We have empirically and theoretically demonstrated that scaling the token norm improves underrepresented semantics and enhances semantic binding in the main manuscript. In this section, we provide further empirical evidence supporting this finding. As illustrated in the Fig.~\ref{sup_fig:analysis_norm}, this phenomenon consistently occurs irrespective of the architectures of the text encoder and denoising network. This observation, reinforced by our theoretical analysis, confirms once again that the phenomenon inherently arises from the fundamental properties of the cross-attention mechanism.

\paragraph{Cross-attention maps of TokeBi.}

\begin{wrapfigure}[27]{r}{0.49\textwidth}
\vspace{-1.0em}
\centering
\includegraphics[width=\linewidth]{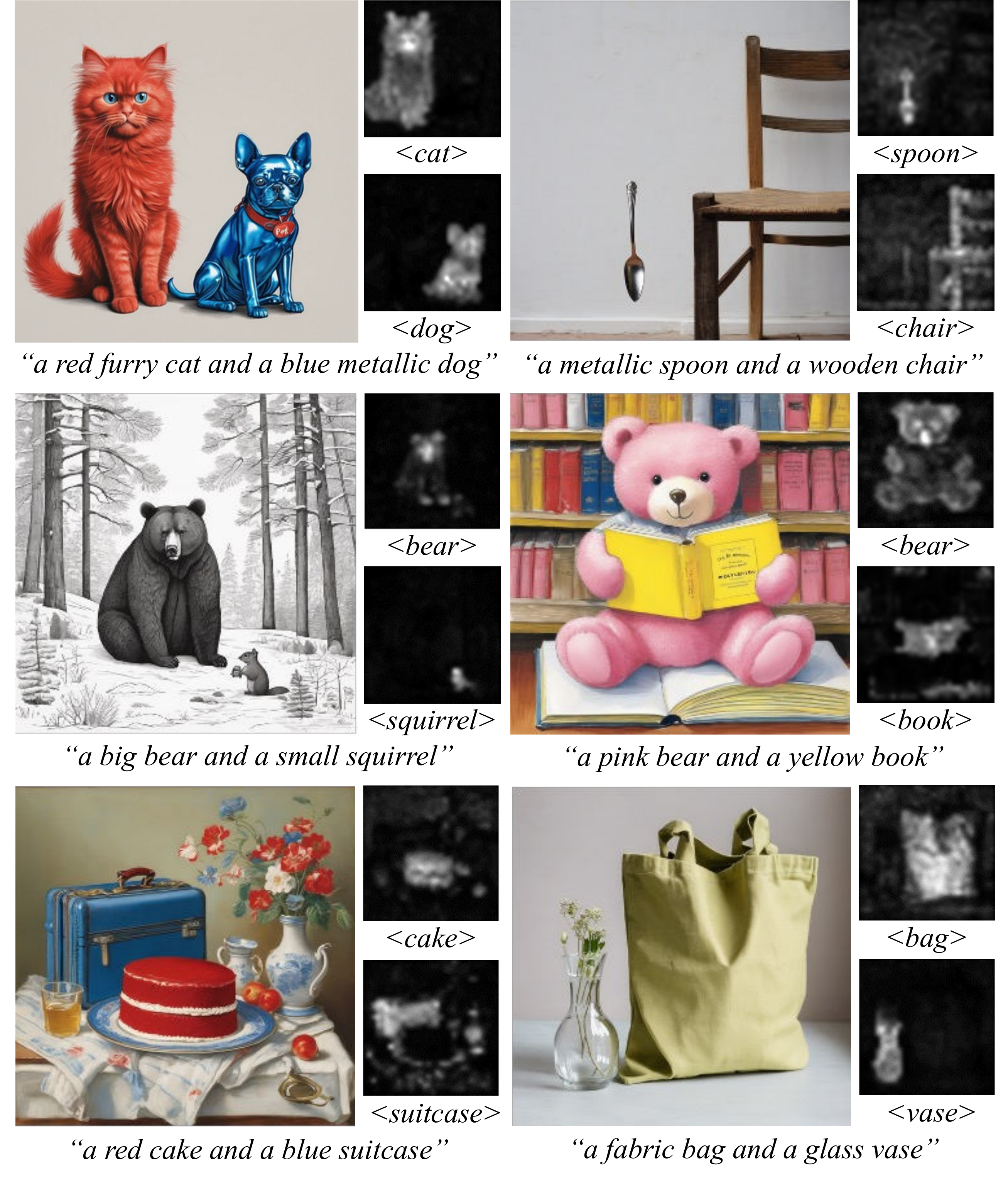}
    \vspace{-1.7em}
    \caption{\textbf{Visualization of images generated by TokeBi with various input prompts, along with their corresponding cross-attention (CA) maps for object tokens within each prompt.} Consistent with our discussion, CA maps for images generated by TokeBi show minimal overlap and distinct separation, confirming that TokeBi effectively achieves clear semantic binding.}
    \label{sup_fig:ca_map}
    \vspace{-2.0em}
\end{wrapfigure}

We developed our method aiming to ensure that tokens belonging to different noun phrases (NPs) possess distinct cross-attention (CA) maps during the denoising process. Here, we qualitatively demonstrate whether tokens corresponding to each object in images generated by TokeBi indeed exhibit distinct attention maps. As shown in Fig.~\ref{sup_fig:ca_map}, despite the presence of multiple objects and diverse attributes describing each object, tokens representing individual objects display mutually exclusive and clear attention maps. These results confirm the effectiveness of our methods.



\section{Efficiency}
\label{sup_sec:efficiency}
To evaluate the efficiency of TokeBi, we measured the inference time and peak memory usage of each method and present the results in Tab.~\ref{sup_tab:time_cost}. 
The evaluation was conducted on an Nvidia A100 GPU using the average of 10 prompts. 
We used the same guidance scale and sampling steps as in the main paper.  
In terms of inference time, TokeBi achieved an inference time of 22.3 seconds, which is the fastest among optimization-based approaches (A\&E, SynGen, and ToMe). For peak memory usage, it recorded 14.1 GB, ranking third with a minimal difference from SDXL (10.7 GB), while maintaining the best performance in semantic binding.  
These results demonstrate that our method not only outperforms existing approaches but also efficiently handles the semantic binding task.

\begin{wraptable}[15]{r}{0.43\textwidth}
\vspace{-1.7em}
\caption{\textbf{Computation costs of TokeBi compared to baseline methods.} Sampling time was measured exclusively during the inference steps following the parsing stage, and peak memory usage was measured using \texttt{torch.cuda.max\_memory\_allocated()} to assess the precise memory usage.}
\vspace{-0.5em}
\centering
\resizebox{1.0\linewidth}{!}{%
\begin{tabular}{c|c|c|c}
\toprule
\multirow{2}{*}{Method} & \multirow{2}{*}{\tabincell{c}{Sampling\\Steps}} & \multirow{2}{*}{\tabincell{c}{Time $\downarrow$\\(sec.)}} & \multirow{2}{*}{\tabincell{c}{Peak Memory $\downarrow$\\(GB)}} \\
& & & \\
\midrule
SDXL~\cite{podell2023sdxl} & 50 & 6.00 & 10.7 \\ 
\midrule
ToMe~\cite{hu2024token} & 50 & 30.6 & 20.4 \\
StructureDiff$_{XL}$~\cite{feng2022training} & 50 & \orangenum{6.14} & \rednum{11.1} \\
A\&E$_{XL}$~\cite{chefer2023attend}& 50 & 37.0 & 29.3 \\ 
SynGen$_{XL}$~\cite{rassin2024linguistic} & 50 & 23.6 & 20.3 \\
Magnet~\cite{zhuang2024magnet} & 50 & \rednum{5.67} & \orangenum{11.1} \\
{TokeBi \textbf{(Ours)}} & 50 & \lightorangenum{22.3} & \lightorangenum{14.1} \\
\bottomrule
\end{tabular}
}
\vspace{-0.5em}
\label{sup_tab:time_cost}
\end{wraptable}
\section{Implementation Details}
\label{sup_sec:implementation_details}

\paragraph{Background of Diffusion Models.} 
Latent Diffusion Models (LDMs) perform denoising in the latent space, significantly reducing computational cost while preserving image quality. A pre-trained encoder $\phi$ maps an image $x$ to its latent representation $z = \phi(x)$, and a decoder $\psi$ reconstructs it, ensuring $\psi(z) \approx x$. The forward diffusion process gradually adds Gaussian noise to $z$, producing noisy latents $z_t$ over timesteps $t = 1, \dots, T$. A denoising network $\epsilon_{\theta}$ learns to remove this noise by minimizing $\|\epsilon_{\theta}(z_t, t) - \epsilon_t\|^2$, where $z_t = \sqrt{\bar{\alpha}_t}z_0 + \sqrt{1-\bar{\alpha}_t}\epsilon,\;\epsilon \sim \mathcal{N}(0,1)$, where $\bar{\alpha}_t$ is the cumulative product of $\alpha$ from $0$ to $t$ as defined in DDPM~\cite{ho2020denoising}. During inference, $z_T \sim \mathcal{N}(0,1)$ is progressively denoised into $z_0$ and decoded back into an image.  

TokeBi operates within the framework of Text-to-Image (T2I) generation and builds upon LDMs, leveraging a pre-trained CLIP text encoder $\mathcal{E}$ to process input prompts. Given a prompt $\mathcal{P}$, the encoder maps it to an embedding $c = \mathcal{E}(\mathcal{P})$, which conditions the diffusion process through cross-attention layers. The loss function $\|\epsilon_{\theta}(z_t, t, c) - \epsilon_t\|^2$ ensures semantic alignment between text and image while maintaining structural integrity in the latent space.

\paragraph{Base models.}
SDXL~\cite{podell2023sdxl}, PlayGround-v2 (PlayG-v2)~\cite{liplayground}, and PixArt-$\Sigma$~\cite{chen2024pixartsigma} are state-of-the-art AI models designed for high-quality image generation. SDXL (Stable Diffusion XL) is an enhanced version of Stable Diffusion~\cite{rombach2022high}, leveraging a Latent Diffusion Model (LDM) architecture with an improved Text-to-Image synthesis capability, enabling higher resolution outputs and more detailed visual representations. PlayGround-v2 also utilizes an LDM-based approach and incorporates a CLIP text encoder, which allows for effective conditioning on textual prompts. Both models inherit causality properties, meaning that later tokens retain the semantic information of preceding tokens~\cite{radford2021learning}. In contrast, PixArt-$\Sigma$ is based on a Diffusion Transformer (DiT)~\cite{peebles2023scalable} architecture and employs a T5-XXL~\cite{raffel2020exploring} text encoder, which differs from CLIP in that it follows a non-causal structure. This design choice allows PixArt-$\Sigma$ to generate high-resolution, photorealistic images with improved efficiency and semantic coherence. These models represent significant advancements in AI-driven image synthesis, catering to both artistic and functional applications.

\paragraph{Comparison methods.}
To assess the effectiveness of our method, we compare it against SOTA approaches in semantic binding tasks for T2I generation. The comparison is categorized into two groups:  
(i) Latent optimization-based methods, including Attend \& Excite (A\&E)~\cite{chefer2023attend} and SynGen~\cite{rassin2024linguistic};  
(ii) Token-based methods, including ToMe~\cite{hu2024token}, StructureDiffusion (StructureDiff)~\cite{feng2022training}, and Magnet~\cite{zhuang2024magnet}.

Among these, StructureDiff and A\&E were originally designed for Stable Diffusion (SD) 1.4, SynGen was developed for SD 1.5, and ToMe, and Magnet were implemented based on SDXL.
To ensure a fair comparison, we adapted each method’s implementation to be compatible with SDXL and PlayG-v2. A\&E was applied using subject token indices within the prompt, while ToMe incorporated subject token indices along with their corresponding adjective tokens and prompt anchors.
For consistency across all models, we applied SpaCy’s dependency parser to refine and align inputs with our framework while preserving each method’s original design. These adaptations ensured a fair benchmarking process and a reliable evaluation of semantic binding performance.
For ToMe~\footnote{\url{https://github.com/hutaihang/ToMe}} and Magnet~\footnote{\url{https://github.com/I2-Multimedia-Lab/Magnet}}, we utilized their official PyTorch implementations. For A\&E~\footnote{\url{https://huggingface.co/docs/diffusers/api/pipelines/attend_and_excite}}, we re-implemented the SDXL version based on the Huggingface implementation. Similarly, for StructureDiff~\footnote{\url{https://github.com/weixi-feng/Structured-Diffusion-Guidance?tab=readme-ov-file}} and SynGen~\footnote{\url{https://github.com/RoyiRa/Linguistic-Binding-in-Diffusion-Models}}, we implemented the SDXL version based on their official implementations.

Note that ToMe showed a significant gap between the performance reported in its original paper and our reproduced results. Using the official implementation, we verified consistency with the described methodology and found that one of ToMe’s key components—the End Token Substitution (ETS) module—consistently degraded performance.
With ETS enabled—under the full configuration reported in the With ETS enabled—using the full configuration reported in the ToMe paper—ToMe underperformed, scoring below the SDXL baseline across all metrics: BLIP-VQA 0.5404 (Color), 0.5324 (Texture), 0.4982 (Shape); GenEval 0.0175, CIC 0.518, VQAScore 0.6451, and VIEScore 6.132 (see SDXL metrics in the first row of Tab.~\ref{tab:2x2_quant_table}).
Despite the core role of ETS, the original paper did not provide an ablation study to assess its individual effect.
Therefore, we report ToMe’s best-case results—ToMe without the ETS component—in the comparison tables throughout this paper.
For fairness and robustness, while ToMe's original paper appears to use a single unstated fixed seed, we conducted five runs with different seeds and report the averaged metrics.
To expand the evaluation, we included additional metrics not reported in the ToMe's original paper.
Note that reproducibility concerns have already been raised by users on the official GitHub repository~\footnote{\url{https://github.com/hutaiHang/ToMe/issues/1}}.
We further evaluated ToMe with PlayG-v2 as base models, observing significantly lower performance with PlayG-v2 (Tab.~\ref{sup_tab:playG_quantitative}).
Importantly, although the ToMe paper presents PlayG-v2 results in its main table, all experiments were conducted solely with SDXL.
These findings highlight the limitations of ToMe and support the robustness and generalizability of our method, TokeBi, across diverse foundation models.

\paragraph{Other implementation considerations.}
To prevent excessive growth in token embedding norms within Adaptive Token Mixing (ATM), we applied appropriate clamping to the mixing matrix $\mathbf{M}$ during optimization.
Additionally, we empirically observed that jointly optimizing the end token (EOT) and padding tokens (PAD), which encapsulate global semantic and contextual information respectively~\cite{toker2025padding, hu2024token}, leads to better control of the cross-attention maps for word tokens within a prompt. So, we applied this optimization, whereas it was not used in PlayG-v2 or PixArt-$\Sigma$. 
For our total loss design, the lambda coefficient was set to 0.01 for SDXL and PlayG-v2, and 0.05 for PixArt-$\Sigma$.  
Regarding sampling hyperparameters, we used 50 steps with a guidance scale of 7.5 for SDXL, 50 steps with a guidance scale of 2.0 for PlayG-v2, and 20 steps with a guidance scale of 4.0 for PixArt-$\Sigma$.  
All experiments were conducted on a single Nvidia A100 GPU.

\section{Dataset}\label{sup_sec:dataset}
\begin{figure*}[p]
\begin{center}
\centerline{\includegraphics[width=0.9\linewidth]{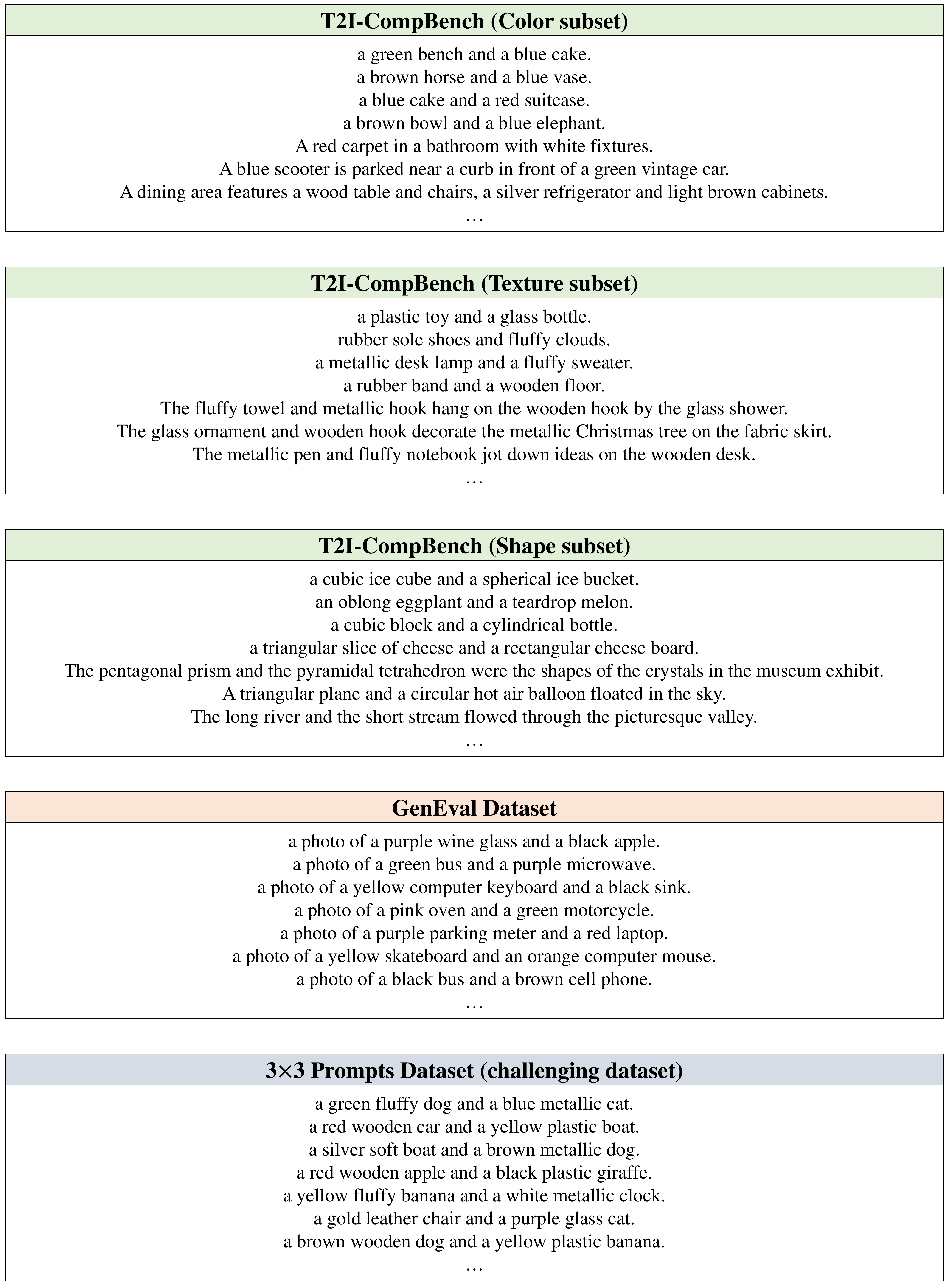}}
\caption{\textbf{Example prompts used in our experiments.} We utilize T2I-CompBench for BLIP-VQA, VQAScore, and VIEScore evaluations, the GenEval color-attribute dataset for GenEval score assessment, and 3$\times$3 prompts—generated from the adjectives and nouns in T2I-CompBench—to evaluate TokeBi's performance in challenging scenarios.}
\label{sup_fig:prompts_examples}
\end{center}
\end{figure*}

In our experiments, we use three datasets: 1) T2I-CompBench, 2) GenEval, and 3) the 3$\times$3 prompts dataset.
The sample prompts are provided in Fig.~\ref{sup_fig:prompts_examples}.

\paragraph{T2I-CompBench.} T2I-CompBench is a benchmark designed to evaluate compositional understanding in Text-to-Image (T2I) generation models. It consists of structured prompts that assess a model’s ability to accurately bind attributes to objects, ensuring proper semantic alignment in multi-object, multi-attribute scenarios. The dataset is categorized into three subsets—color, texture, and shape—each containing 300 prompts, including 240 relatively simple prompts and 60 longer, more complex ones. By providing diverse and challenging text prompts, T2I-CompBench serves as a rigorous testbed for measuring semantic binding and mitigating semantic neglect and attribute misassignment issues commonly observed in T2I models.  

For evaluation, each prompt is tested with five random seeds, generating a total of $300 \times 3$ subsets $\times$ 5 seeds $=$ 4500 images. The BLIP-VQA score is computed across the full prompts of the three attribute subsets in T2I-CompBench.

\paragraph{GenEval dataset.}
The GenEval dataset comprises 100 prompts (totaling 500 generated images with five seeds) designed for color alignment evaluation and is used to compute the GenEval metric proposed in the original study.

\paragraph{3$\times$3 prompts.} For the 3$\times$3 prompts dataset, we construct 200 prompts using words from T2I-CompBench, following the template “a/an \{adj.\} \{adj.\} \{noun\} and a/an \{adj.\} \{adj.\} \{noun\}.” These prompts are generated using an GPT-4o as a LLM. Each prompt is evaluated with five seeds, resulting in a total of 1000 generated images for assessment.


\section{Evaluation Metric Details}
\label{sup_sec:eval_metric_details}

\paragraph{BLIP-VQA.}
The BLIP-VQA metric in T2I-CompBench~\cite{huang2023t2i} uses BLIP~\cite{li2022blip}’s visual question answering module to assess whether a generated image accurately reflects the intended prompt. It queries the BLIP model with predefined questions about objects, attributes, and relationships, and measures alignment based on answer correctness. Compared to simpler metrics like CLIP-Score, BLIP-VQA offers finer insight into whether specific objects or attributes are missing or misrepresented, thus providing a more precise measure of compositional alignment. In our work, we also employ BLIP-VQA on T2I-CompBench subsets, following prior approaches~\cite{feng2024ranni, hu2024ella, jiang2024comat, hu2024token}.

\paragraph{GenEval.}
GenEval is a comprehensive evaluation metric designed to assess how effectively a Text-to-Image (T2I) generative model aligns its generated images with corresponding textual prompts. In this study, GenEval was utilized to evaluate whether image synthesis outputs produced by the compressed text encoder accurately reflect their intended textual descriptions. The GenEval metric consists of six distinct sub-metrics: Single Object Generation, assessing the model's capability to generate images from prompts containing a single object (\textit{e.g.}, \textit{`a photo of a giraffe'}); Two Objects Generation, evaluating image generation from prompts involving two distinct objects (\textit{e.g.}, \textit{`a photo of a knife and a stop sign'}); Counting, measuring accuracy in representing the specified quantity of objects (\textit{e.g.}, \textit{`a photo of three apples'}); Colors, verifying the accurate depiction of colors specified in the prompts (\textit{e.g.}, \textit{`a photo of a pink car'}); Position, testing the model's comprehension of spatial relationships described within prompts (\textit{e.g.}, \textit{`a photo of a sofa under a cup'}); and Color Attribution, assessing the correct assignment of specified colors to multiple objects (\textit{e.g.}, \textit{`a photo of a black car and a green parking meter'}). In our evaluation, we specifically employed the ``Color Attribution'' metric to measure semantic binding performance in image generation, with random seeds.

\paragraph{CIC.} 
The evaluation metric proposed by Chen \etal~\cite{chen2024cat}, referred to as CIC (named after the paper title ``A Cat Is A Cat''), quantifies information loss and object mixture issues in Text-to-Image (T2I) diffusion models more accurately than existing methods, such as CLIP-based similarity scores. Unlike the CLIP score, CLIP-BLIP score, and bounding box overlap-based measures, which struggle to precisely assess object presence and mixture, CIC leverages OWL-ViT, an open-vocabulary object detection model, for more robust evaluation. It classifies generated images based on whether objects are fully present, mixed, or missing. In this paper, we specifically apply CIC to assess TokeBi’s object neglect issue, restricting our evaluation to the `$2$ objects exist' case as a hard metric and reporting the results in Tab.~\ref{tab:2x2_quant_table}, Tab.~\ref{tab:3x3_quant_table}, and Tab.~\ref{sup_tab:playG_quantitative}.

\paragraph{VQAScore.} 
VQAScore~\cite{lin2024evaluating} is a metric for evaluating Text-to-Visual generation by leveraging visual-question-answering (VQA) models to assess image-text alignment. Unlike CLIPScore, which struggles with complex compositional prompts, VQAScore computes alignment by estimating the probability of a ``Yes'' response to a structured question like ``Does this figure show {text}?''. 
This approach, implemented with open-source VQA models, achieves state-of-the-art performance across multiple benchmarks, surpassing proprietary models like GPT-4o. Additionally, the study introduces GenAI-Bench, a benchmark with 1,600 compositional prompts and 15,000+ human ratings, further validating VQAScore’s reliability.

\paragraph{VIEScore.} 
VIEScore \cite{ku2024viescore} proposes a framework called Visual Instruction-guided Explainable Score for evaluating conditional image synthesis. It introduces a general assessment methodology and compares its alignment with existing metrics and human judgments across tasks like Text-to-Image generation, image editing, and subject-driven image generation. Leveraging the reasoning capabilities of Multimodal Large Language Models (MLLMs), VIEScore enables training-free evaluation. TokeBi's VIEScore metric employs the ``Text-Guided Image Generation'' prompt template from the Semantic Consistency (SC) Rating Prompt Template using GPT-4o, sourced from the original VIEScore paper’s Appendix.

\begin{figure}[!t]
\vspace{-1.0em}
\centering
\includegraphics[width=0.8\linewidth]{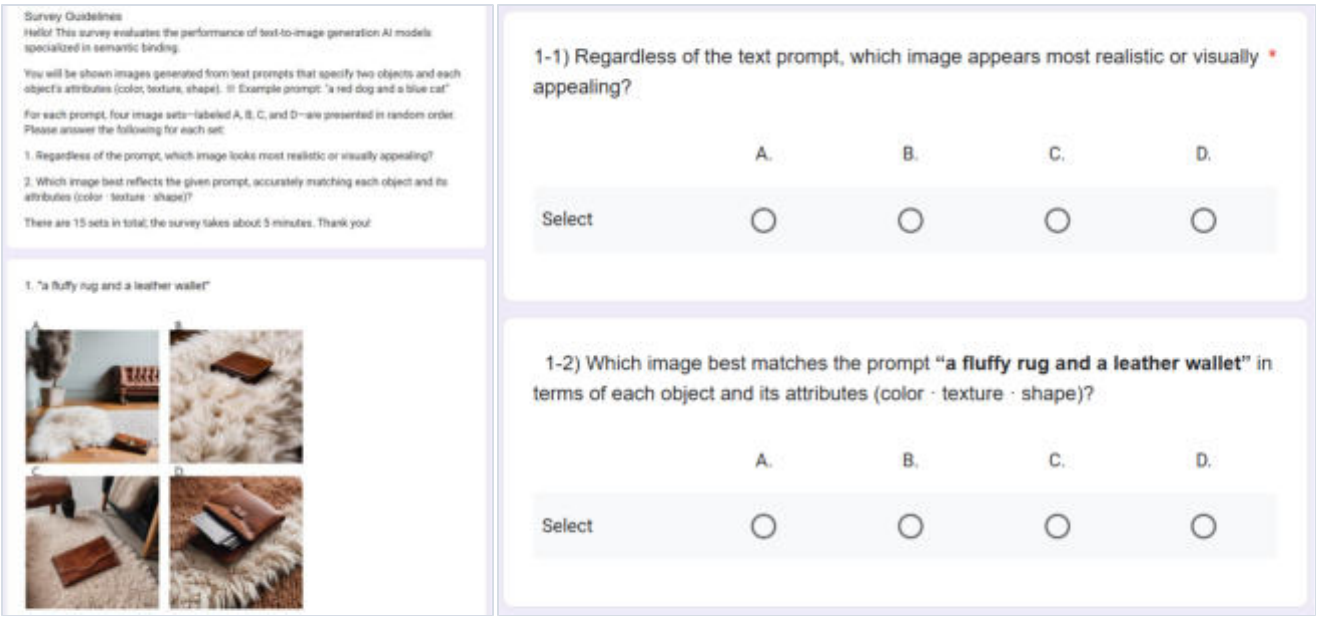}
    \vspace{-0.5em}
    \caption{A screenshot of the user study for human evaluation on \romannumeral 1) perceptual quality and \romannumeral 2) semantic alignment accuracy of objects and their attributes in the text prompt. Four methods (SDXL, Magnet, ToMe, and TokeBi) were compared anonymously with randomized image order.}
    \label{sup_fig:user_study_screenshot}
    \vspace{-1.0em}
\end{figure}

\paragraph{Human evaluation.} 
In addition to the automatic evaluation metrics presented in the main paper, we conducted a user study involving 143 participants to further validate our method.
From each of the three subsets of T2I-CompBench and from our challenging 3$\times$3 dataset, we randomly selected 3 prompts per dataset, totaling 12 prompts.
Participants evaluated \romannumeral 1) \emph{perceptual quality} and \romannumeral 2) \emph{semantic alignment accuracy} between the text prompt and the generated images.
For each image set, participants answered two questions: 1) Regardless of the prompt, which image appears most realistic or visually appealing? 2) Which image best aligns with the prompt in terms of object attributes (color, texture, shape)?
Our method was evaluated anonymously alongside SDXL, Magnet, and ToMe, with the image order randomized. A screenshot of the user study interface is shown in Fig.~\ref{sup_fig:user_study_screenshot}. Participants selected their top choice for each question, and we report the resulting selection ratios. The user study results are presented in Tab.~\ref{sup_tab:user_study_results} in Sec.~\ref{sup_subsec:add_quanti}.

\section{Additional Ablation Studies}
\label{sup_sec:ablation}
\begin{figure}[p]
\begin{center}
\centerline{\includegraphics[width=0.95\linewidth]{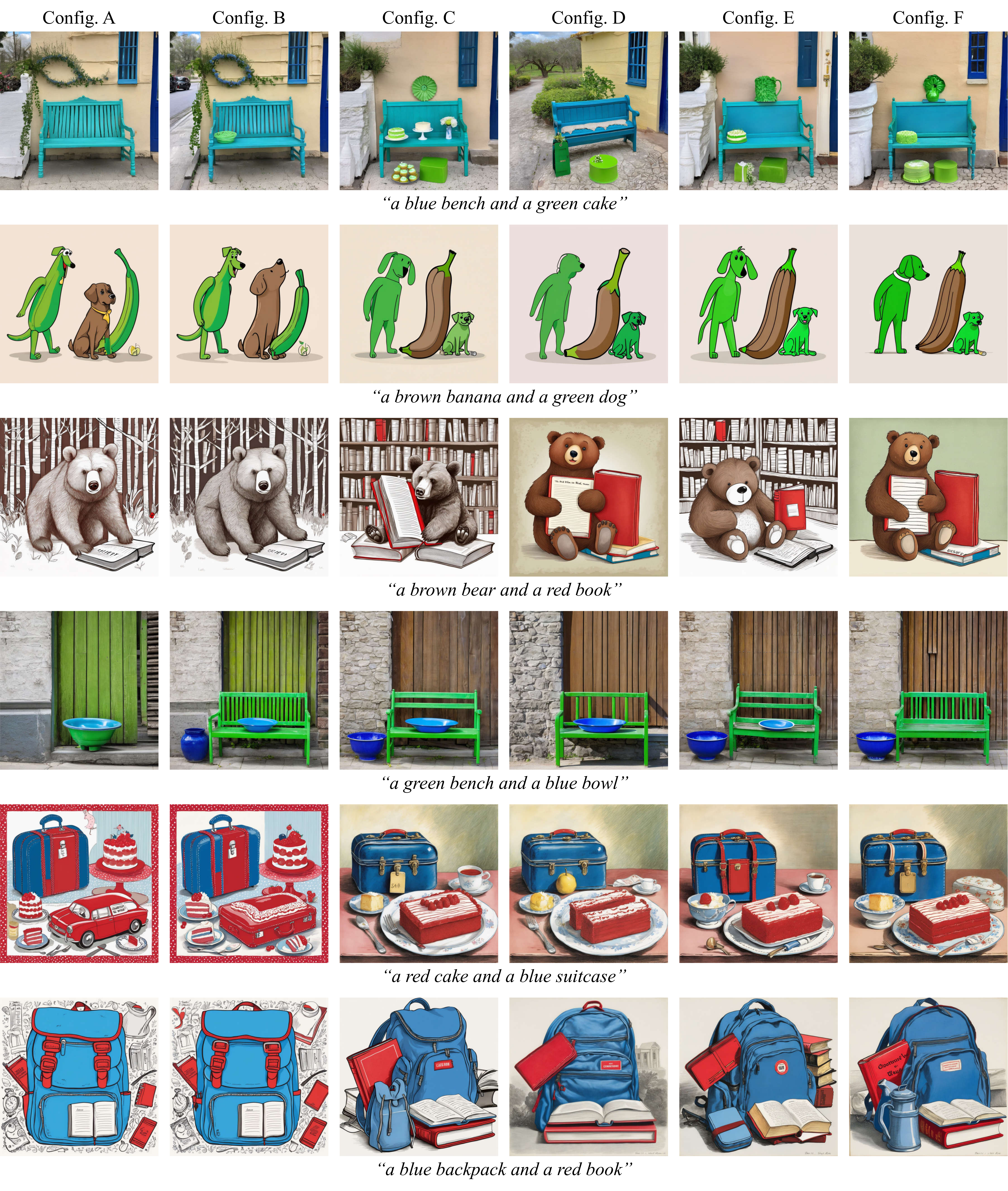}}
\caption{\textbf{Qualitative results of the ablation study for each configuration on the ``Color'' subset of T2I-CompBench.} With CAPO, the frequency of generating irrelevant or missing objects decreases significantly. Furthermore, as each component is incrementally added, images become increasingly aligned with the text prompt, demonstrating improved attribute binding. Ultimately, configuration F exhibits the best overall performance.}
\label{sup_fig:ablation_color}
\end{center}
\end{figure}

\begin{figure*}[p]
\begin{center}
\centerline{\includegraphics[width=0.95\linewidth]{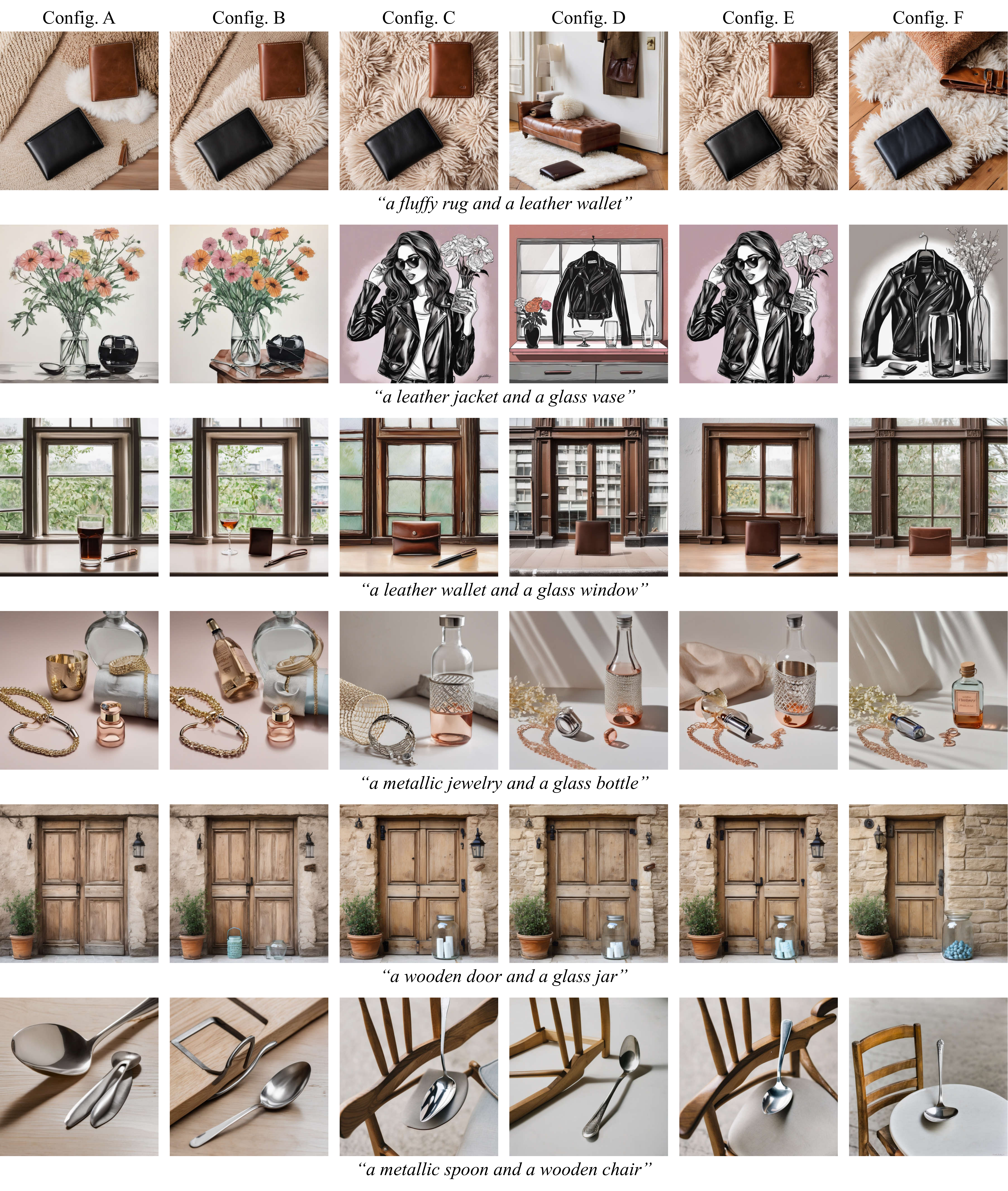}}
\caption{\textbf{Qualitative results of the ablation study for each configuration on the ``Texture'' subset of T2I-CompBench.} With CAPO, the frequency of generating irrelevant or missing objects decreases significantly. Furthermore, as each component is incrementally added, images become increasingly aligned with the text prompt, demonstrating improved attribute binding. Ultimately, configuration F exhibits the best overall performance.}
\label{sup_fig:ablation_texture}
\end{center}
\end{figure*}

\begin{figure}[p]
\begin{center}
\centerline{\includegraphics[width=0.95\linewidth]{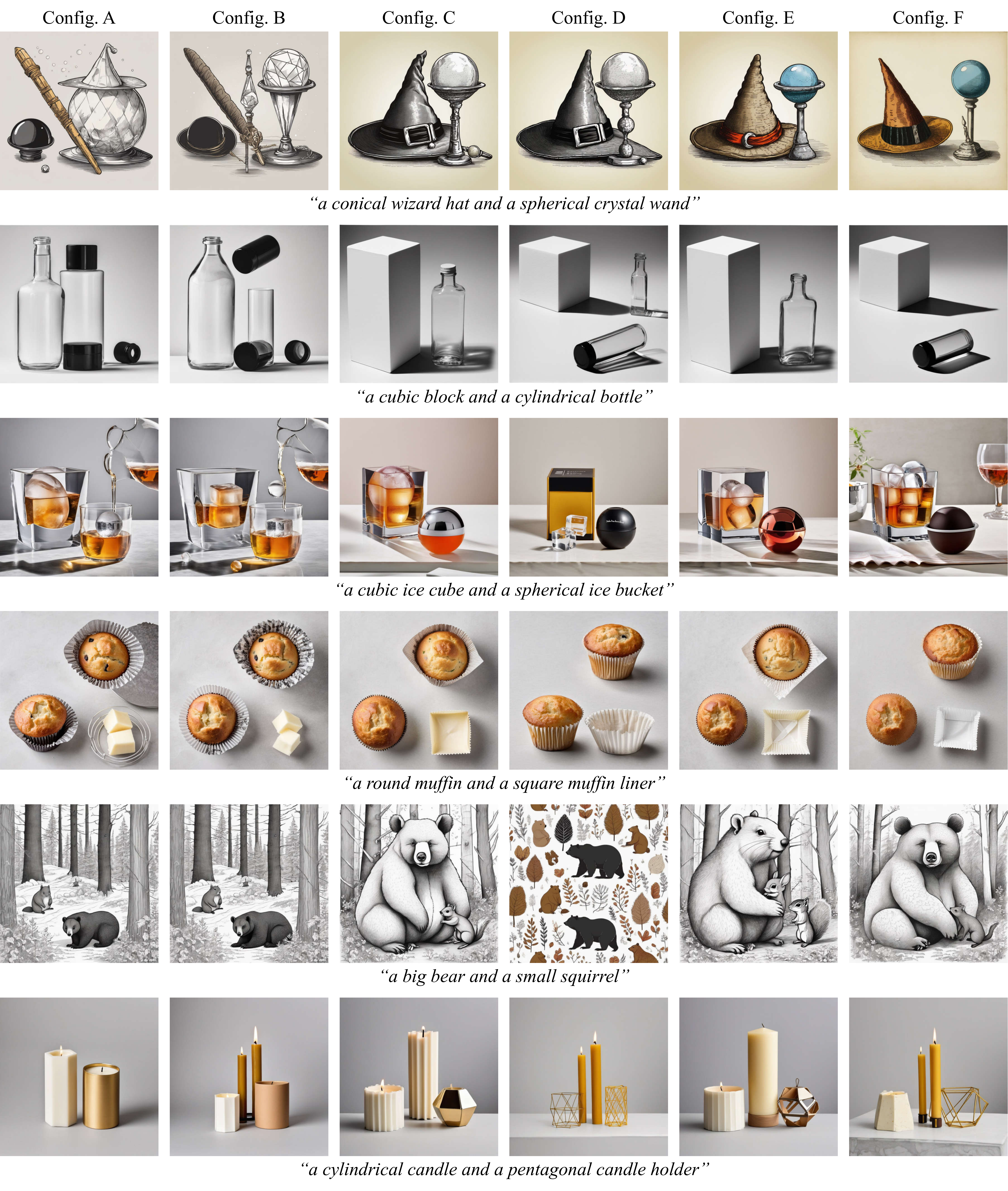}}
\caption{\textbf{Qualitative results of the ablation study for each configuration on the ``Shape'' subset of T2I-CompBench.} With CAPO, the frequency of generating irrelevant or missing objects decreases significantly. Furthermore, as each component is incrementally added, images become increasingly aligned with the text prompt, demonstrating improved attribute binding. Ultimately, configuration F exhibits the best overall performance.}
\label{sup_fig:ablation_shape}
\end{center}
\end{figure}

\paragraph{Additional qualitative results.}
In addition to the quantitative results presented in the main text, we also report qualitative results for each configuration. We provide qualitative results for color, texture, and shape from the T2I-CompBench, as illustrated in the Fig.~\ref{sup_fig:ablation_color}, Fig.~\ref{sup_fig:ablation_texture}, and Fig.~\ref{sup_fig:ablation_shape}. As shown in the results, each component of TokeBi contributes to the performance of semantic binding.

\begin{figure}[h]
\begin{center}
\centerline{\includegraphics[width=\linewidth]{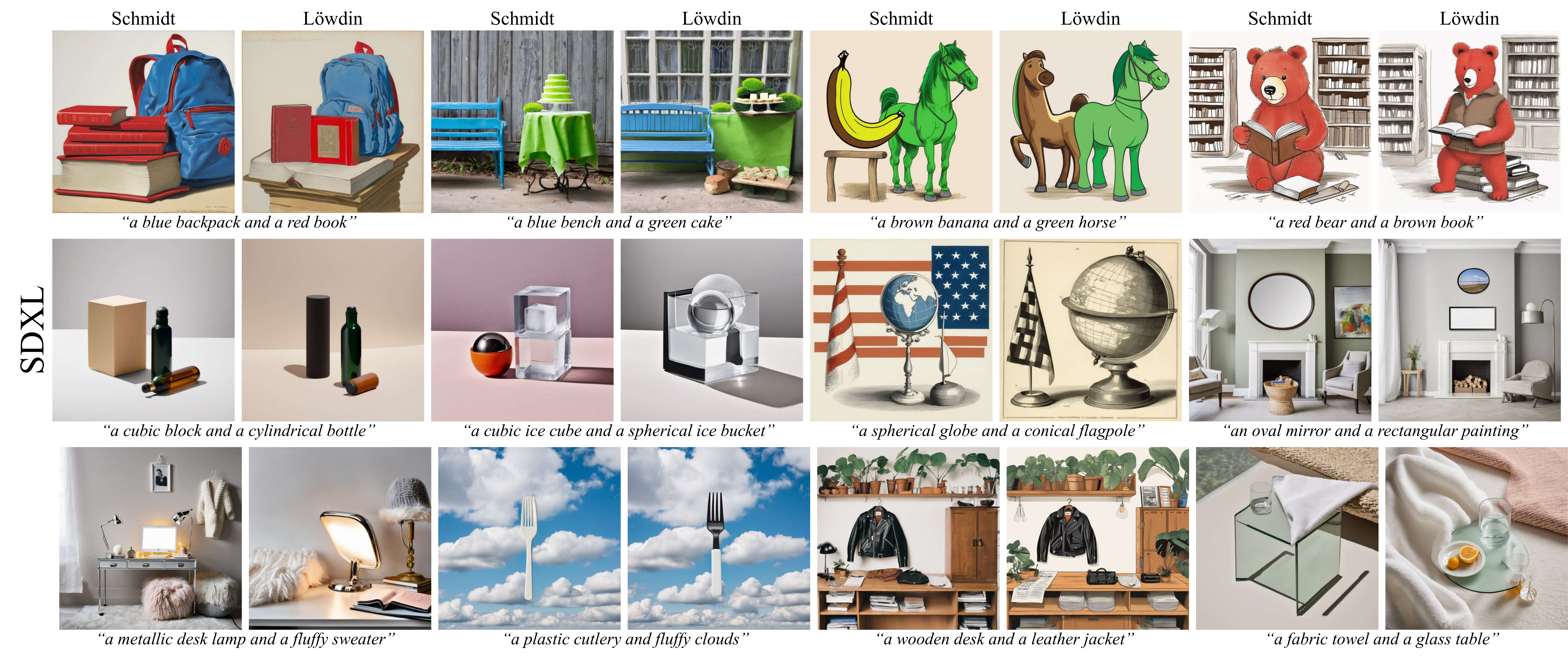}}
\caption{\textbf{Qualitative results illustrating the effectiveness of causality awareness in CAPO using the SDXL model.} As SDXL utilizes a causal text encoder (CLIP text encoder), CAPO applies Schmidt orthogonalization. The results presented here show outcomes when L\"owdin orthogonalization, which is causality-unaware, is applied within TokeBi. As demonstrated, the absence of causality awareness frequently leads to object neglect or worse semantic binding performance.}
\label{sup_fig:capo_ablation_sdxl}
\end{center}

\begin{center}
\centerline{\includegraphics[width=\linewidth]{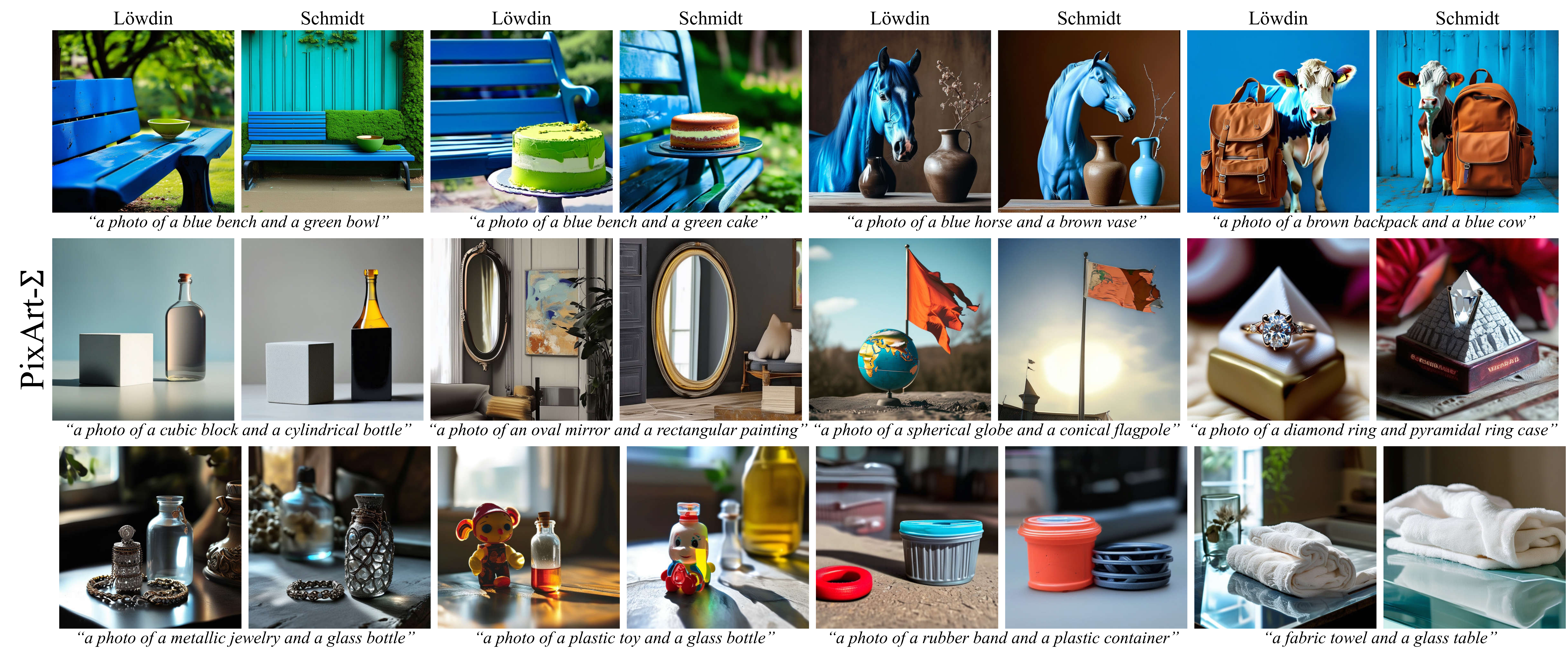}}
\caption{\textbf{Qualitative results illustrating the effectiveness of causality awareness in CAPO using the PixArt-$\mathbf{\Sigma}$ model.} Since PixArt-$\Sigma$ employs a non-causal text encoder (T5 text encoder), CAPO utilizes L\"owdin orthogonalization. The presented results show cases where Schmidt orthogonalization—which assumes causality—is incorrectly applied within TokeBi. As shown, neglecting causality-awareness leads to object neglect or inadequate semantic binding.}
\label{sup_fig:capo_ablation_pixart}
\end{center}
\end{figure}

\paragraph{Ablation study on a causality awareness.}
Through Causality-Aware Projection Out (CAPO), we increased the angular distance by employing asymmetric orthogonalization via Schmidt orthogonalization for causal text encoders, thus accounting for the unidirectional nature of information flow within these encoders. Conversely, for non-causal text encoders, we increased the angular distance by applying symmetric orthogonalization through L\"owdin orthogonalization, effectively considering the bidirectional exchange of information. To verify the effectiveness of these causality-aware orthogonalization methods, we applied L\"owdin orthogonalization to SDXL~\cite{podell2023sdxl}, which employs a causal text encoder (CLIP~\cite{radford2021learning}), and Schmidt orthogonalization to PixArt-$\Sigma$~\cite{chen2024pixart}, which uses a non-causal text encoder (T5-XXL~\cite{raffel2020exploring}). The corresponding results are presented in the Fig.~\ref{sup_fig:capo_ablation_sdxl} for SDXL, and Fig.~\ref{sup_fig:capo_ablation_pixart} for PixArt-$\Sigma$, confirming the validity of our orthogonalization strategies based on text encoder causality. In these experiments, all other components of TokeBi were kept unchanged, except for the orthogonalization methods.

\begin{figure}[t]
    \centering
    \includegraphics[width=\linewidth]{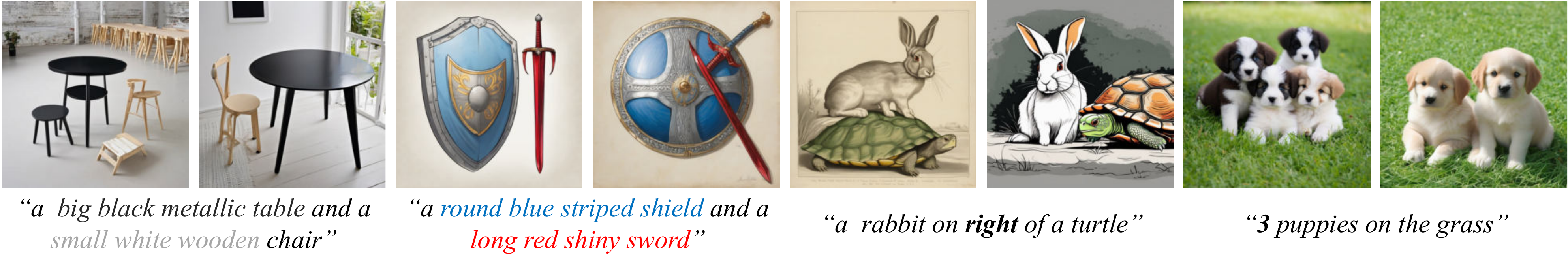}
    \vspace{-2em}
    \caption{\textbf{Qualitative results demonstrating the limitations of TokeBi.} The four generated images on the left illustrate cases where TokeBi struggles when multiple attributes describe a single object. The four synthesized images on the right highlight failure cases inherited from the base diffusion model, specifically in handling counting and spatial relationships.}
    \label{sup_fig:limitation}
    \vspace{-0.5em}
\end{figure}

\section{Limitations}
\label{sup_sec:limitations}
While TokeBi provides an efficient and effective solution to the attribute binding problem, it has inherent limitations. Fig.~\ref{sup_fig:limitation} presents failure cases that illustrate key challenges. Although our method effectively controls attention, generated images may still exhibit missing attributes or objects and attribute leakage for prompts with a large number of attributes (columns 1-4). This is likely due to the increased complexity of the loss landscape as the number of cross-attention maps to be controlled increases, making optimization more challenging.
Another limitation is TokeBi’s occasional misalignment of positional or numerical relationships (columns 5-8), where the generated concepts are semantically correct but incorrectly positioned or counted. This limitation stems from the inherent difficulty of the pretrained diffusion models in precisely capturing spatial and numerical relationships within complex scenes. As our focus lies primarily on the semantic binding problem, these inherent limitations of diffusion models persist in our approach.
These limitations, while varying in severity, are well-documented challenges in the semantic binding domain~\cite{rassin2024linguistic,zhuang2024magnet}. We leave these challenges as future work.

\section{Broader Impacts} 
\label{sup_sec:broader_impacts}
Our research explores strong semantic binding in Text-to-Image (T2I) generation by leveraging the geometric properties of text token embeddings. Like other T2I methods, it poses risks of misuse, such as generating misleading images that may spread misinformation or infringe on personal copyrights and intellectual property. Responsible and mature use of T2I synthesis requires continued attention from academia, industry, and users alike.

\begin{table*}[!t]
\caption{\textbf{Additional semantic binding results are reported on diverse T2I benchmarks using PlayG-v2~\cite{liplayground} as the base model.} As prior methods did not use PlayG-v2, we re-implemented them with it (\textit{{method}}$_{PGv2}$). Top-three scores are highlighted in \textit{red}, \textit{orange}, and \textit{yellow}. TokeBi, built on PlayG-v2, outperforms most baselines without additional training.}
\vspace{-0.5em}
\centering
\resizebox{0.8\linewidth}{!}{%
\begin{tabular}{c|c|ccc|c|c}
\toprule
\multirow{2}{*}{Method} & \multirow{2}{*}{\tabincell{c}{Base\\Model}} & \multicolumn{3}{c|}{BLIP-VQA~\cite{huang2023t2i} $\uparrow$} & \multirow{2}{*}{\tabincell{c}{CIC~\cite{chen2024cat} $\uparrow$\\`2 Obj. exist'}} & \multirow{2}{*}{\tabincell{c}{VQAScore~\cite{lin2024evaluating}} $\uparrow$} \\ 
& & Color & Texture & Shape & & \\ 
\midrule
PlayG-v2~\cite{liplayground} & - & 0.5743 & 0.5832 & 0.4849 & 0.463 & 0.7964 \\
\midrule
StructureDiff$_{PGv2}$~\cite{feng2022training} & \multirow{6}{*}{PlayG-v2} & 0.4820 & 0.5146 & 0.4234 & 0.376 & 0.6762 \\ 
A\&E$_{PGv2}$~\cite{chefer2023attend} & & 0.5637 & \lightorangenum{0.5859} & \orangenum{0.4886} & \lightorangenum{0.471} &\orangenum{0.7971} \\ 
SynGen$_{PGv2}$~\cite{rassin2024linguistic} & & \lightorangenum{0.5792} & 0.5625 & 0.4661 & 0.438 & \lightorangenum{0.7924} \\ 
Magnet$_{PGv2}$~\cite{zhuang2024magnet} & & \rednum{0.6630} & \orangenum{0.5864} & \lightorangenum{0.4849} & \orangenum{0.488} & {0.7911} \\
ToMe$_{PGv2}$~\cite{jiang2024comat} & & 0.3688 & 0.4111 & 0.3214 & 0.248 & 0.6001 \\
TokeBi \textbf{(Ours)} & & \orangenum{0.6464} & \rednum{0.6152} & \rednum{0.5181} & \rednum{0.499} & \rednum{0.8078} \\
\bottomrule
\end{tabular}
}
\vspace{-0.2em}
\label{sup_tab:playG_quantitative}
\vspace{-0.5em}
\end{table*}

\section{Additional Results}
\label{sup_sup_sec:add_results}
\subsection{Additional Quantitative Results}
\label{sup_subsec:add_quanti}
\paragraph{Other base model.}

\begin{wraptable}[8]{tr}{0.46\textwidth}
\vspace{-1.2em}
\caption{\textbf{Performance comparison of TokeBi and its base model PixArt-$\Sigma$.} The best scores are marked in \textit{red}. TokeBi outperforms PixArt-$\Sigma$ on average.}
\label{sup_tab:pixart_quantitative}
\vspace{-0.5em}
\centering
\renewcommand{\arraystretch}{1.0} 
\setlength{\tabcolsep}{1.0em} 
\resizebox{\linewidth}{!}{%
\begin{tabular}{c|cccc}
\toprule
\multirow{2}{*}{Method} & \multicolumn{4}{c}{BLIP-VQA~\cite{huang2023t2i} $\uparrow$} \\
& Color & Texture & Shape & Avg. \\
\midrule
PixArt-$\Sigma$~\cite{chen2024pixartsigma} & 0.6167 & 0.5854 & 0.4604 & 0.5542 \\
{TokeBi \textbf{(Ours)}} & \rednum{0.6664} & \rednum{0.5962} & \rednum{0.4871} & \rednum{0.5832} \\
\bottomrule
\end{tabular}
}
\end{wraptable}
In the main manuscript, we measured the semantic binding performance of various methods across multiple metrics, using SDXL as the base model. In this section, we examine quantitative results obtained by applying TokeBi and previous works to the PlayGround-v2 model.  
We also applied TokeBi to the PixArt-$\Sigma$\cite{chen2024pixartsigma} base model.
The experimental outcomes presented in the Tab.~\ref{sup_tab:playG_quantitative} and Tab.~\ref{sup_tab:pixart_quantitative} demonstrate that TokeBi maintains superior performance across different base models.

\paragraph{Comparison with training requiring methods.}

\begin{wraptable}[15]{tr}{0.5\textwidth}
\vspace{-1.2em}
\caption{\textbf{Performance comparison between TokeBi and training-based methods.} Results for training-based baselines (marked with *) are taken from their original papers due to closed-source implementations. Despite being entirely training-free, TokeBi matches or exceeds the performance of methods demanding significant training resources.}
\vspace{-0.5em}
\centering
\resizebox{\linewidth}{!}{%
\begin{tabular}{c|c|cccc}
\toprule
\multirow{2}{*}{Method} & \multirow{2}{*}{\tabincell{c}{Training\\free}} & \multicolumn{4}{c}{BLIP-VQA~\cite{huang2023t2i} $\uparrow$} \\
& & Color & Texture & Shape & Avg. \\
\midrule
SDXL~\cite{podell2023sdxl} & \cmarkg & 0.6121 & 0.5568 & 0.4983 & 0.5557 \\ 
\midrule
Ranni*~\cite{feng2024ranni} & \xmarkg & 0.6893 & 0.6325 & 0.4934 & 0.6051  \\ 
WiCLIP*~\cite{zarei2024understanding} & \xmarkg & \orangenum{0.7801} & 0.6557 & 0.5166 & 0.6058 \\
ELLA*~\cite{hu2024ella} & \xmarkg & 0.7260 & \rednum{0.6686} & \rednum{0.5634} & \lightorangenum{0.6527} \\ 
CoMat*~\cite{jiang2024comat} & \xmarkg & \rednum{0.7872} & \lightorangenum{0.6468} & \lightorangenum{0.5329} & \orangenum{0.6541}  \\
\midrule
{TokeBi \textbf{(Ours)}} & \cmarkg & \lightorangenum{0.7610} & \orangenum{0.6581} & \orangenum{0.5472} & \rednum{0.6555} \\
\bottomrule
\end{tabular}
}
\vspace{-0.3em}
\label{sup_tab:comparison_train}
\end{wraptable}
TokeBi performs semantic binding through an optimization process at inference time without requiring any training. While prior approaches have aimed to enhance semantic binding performance through extensive training~\cite{feng2024ranni, hu2024ella, jiang2024comat, zarei2024understanding}, we validate the effectiveness of TokeBi by comparing it against these training-requiring methods. The results, presented in Tab.~\ref{sup_tab:comparison_train}, include training-based methods implemented using SDXL~\cite{podell2023sdxl} as the base model. Since all models are closed-source, we directly report numerical results from the respective papers.
Notably, while WiCLIP~\cite{zarei2024understanding} does not provide SDXL experimental results in the main paper, such results were disclosed during the review process~\footnote{\url{https://openreview.net/forum?id=QVBeBPsmy0}}, and we include them in our report. Despite being entirely training-free, TokeBi achieves performance on par with or even surpassing training-dependent methods. In particular, TokeBi establishes a new state-of-the-art in the average performance across three subsets of BLIP-VQA.

\begin{wraptable}[10]{h!}{0.49\textwidth}
\vspace{-1.2em}
\caption{\textbf{User study on perceptual and semantic quality.} TokeBi ranks highest in human-rated perceptual quality and semantic alignment. Scores are normalized to 100.}
\vspace{-0.5em}
\centering
\renewcommand{\arraystretch}{1.0} 
\setlength{\tabcolsep}{1.2em} 
\resizebox{\linewidth}{!}{%
\begin{tabular}{c|c|c}
\toprule
\multirow{2}{*}{Method} & \multirow{2}{*}{\tabincell{c}{Perceptual\\quality}} & \multirow{2}{*}{\tabincell{c}{Text-image \\ semantic alignment}} \\
& & \\
\midrule
SDXL~\cite{podell2023sdxl} & 16.61 & 9.09 \\ 
Magnet~\cite{zhuang2024magnet} & \lightorangenum{21.33} & \orangenum{15.56} \\
ToMe~\cite{hu2024token} & \orangenum{22.73} & \lightorangenum{11.19} \\
{TokeBi \textbf{(Ours)}} & \rednum{39.34} & \rednum{64.16} \\
\bottomrule
\end{tabular}
}
\vspace{-0.5em}
\label{sup_tab:user_study_results}
\end{wraptable}

\paragraph{User study for perceptual quality and exact semantic alignment.}
We conducted a user study on perceptual quality and exact semantic alignment between the text prompt and the generated image, following the protocol and interface shown in Sec.~\ref{sup_sec:eval_metric_details}.
Participants evaluated image sets generated by SDXL, Magnet, ToMe, and TokeBi (ours), selecting the most appropriate image for each question. We report the overall selection ratios across methods.
As summarized in Tab.~\ref{sup_tab:user_study_results}, our method, TokeBi, consistently demonstrated the highest perceptual quality and the most accurate semantic alignment.

\subsection{Additional Qualitative Results}
\label{sup_subsec:add_quali}
\begin{figure}[!t]
\begin{center}
\centerline{\includegraphics[width=\linewidth]{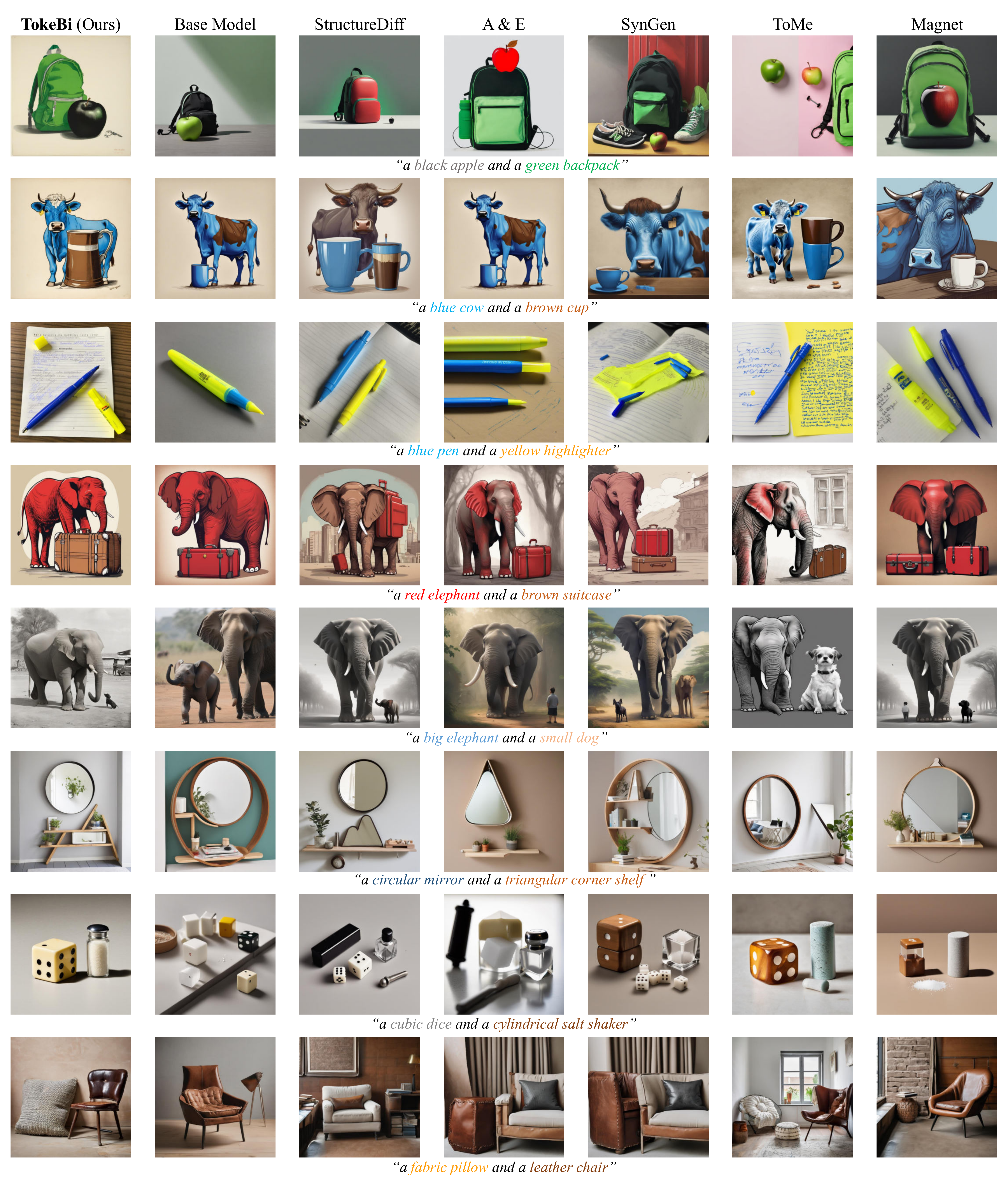}}
\vspace{-0.5em}
\caption{\textbf{Qualitative comparison of TokeBi against other baseline methods using the SDXL as a base model on the T2I-CompBench dataset.} The prompts used in this comparison consist of relatively simple cases, each containing two objects with a single descriptive attribute per object, covering diverse examples across color, texture, and shape. This comparison confirms that TokeBi achieves superior semantic binding performance, not only quantitatively but also qualitatively.}
\label{sup_fig:quali_1}
\vspace{-1.0em}
\end{center}
\end{figure}

\begin{figure}[!t]
\begin{center}
\centerline{\includegraphics[width=\linewidth]{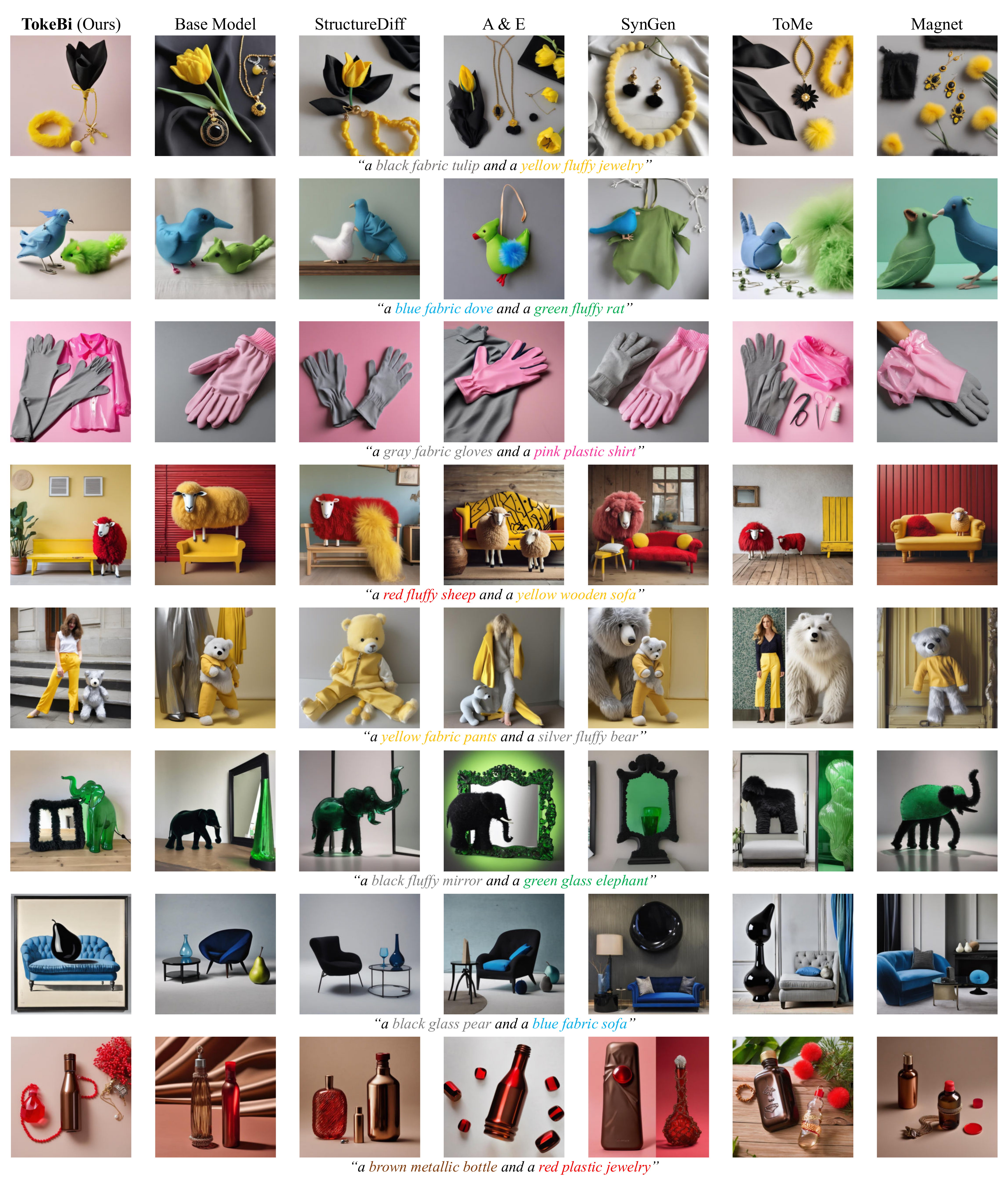}}
\vspace{-0.5em}
\caption{\textbf{Qualitative comparison of TokeBi against other baseline methods using the SDXL as a base model on our $3\times 3$ dataset.} The prompts used in this comparison consist of relatively simple cases, each containing two objects with a single descriptive attribute per object, covering diverse examples across color, texture, and shape. This comparison confirms that TokeBi achieves superior semantic binding performance, not only quantitatively but also qualitatively.}
\label{sup_fig:quali_2}
\vspace{-1.0em}
\end{center}
\end{figure}

\begin{figure}[!t]
\begin{center}
\centerline{\includegraphics[width=\linewidth]{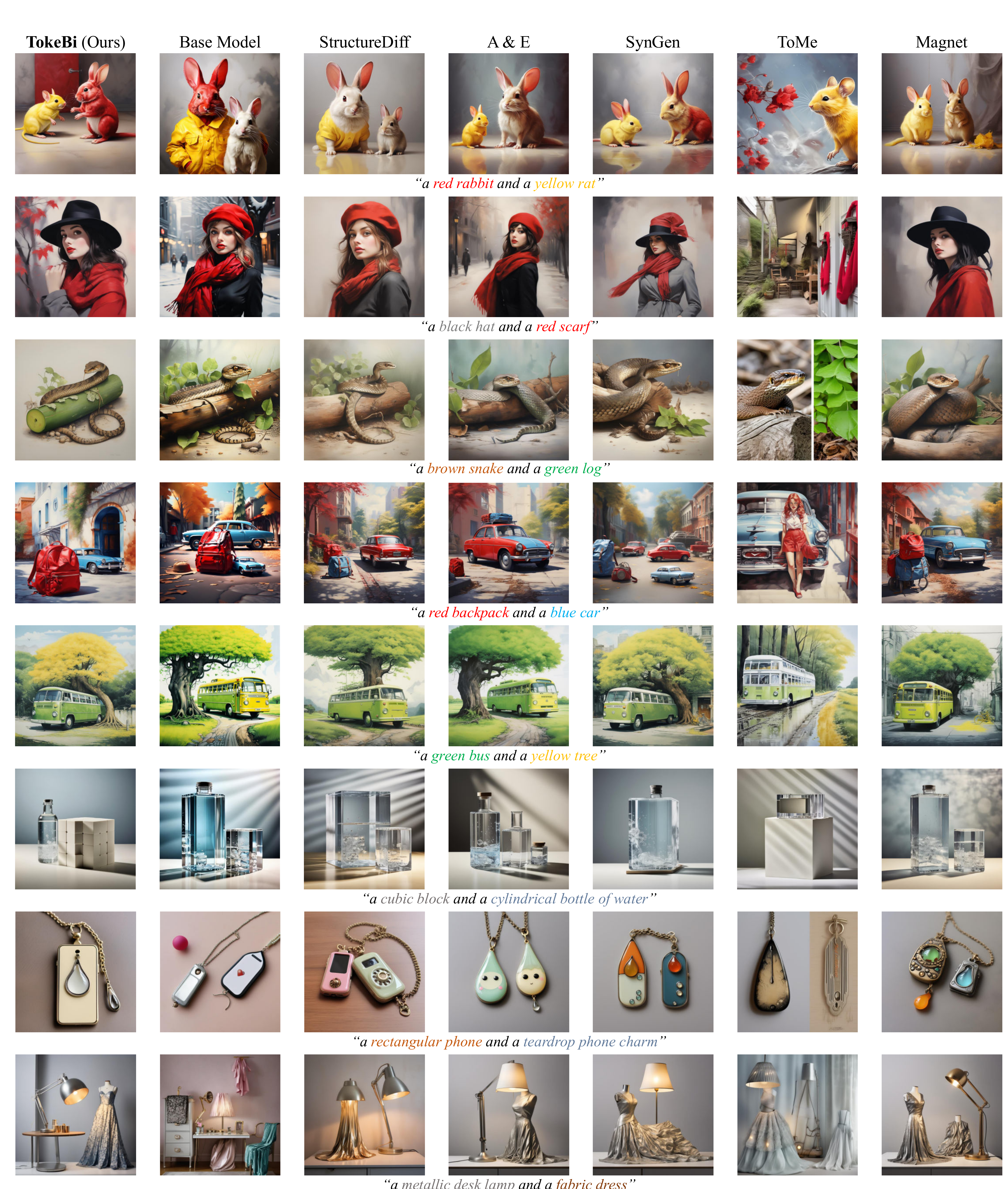}}
\vspace{-0.5em}
\caption{\textbf{Qualitative comparison of TokeBi against other baseline methods using the PlayGround-v2 as a base model on the T2I-CompBench dataset.} The prompts used in this comparison consist of relatively simple cases, each containing two objects with a single descriptive attribute per object, covering diverse examples across color, texture, and shape. This comparison confirms that TokeBi achieves superior semantic binding performance, not only quantitatively but also qualitatively.}
\label{sup_fig:quali_3}
\vspace{-1.0em}
\end{center}
\end{figure}

\begin{figure}[!t]
\begin{center}
\centerline{\includegraphics[width=\linewidth]{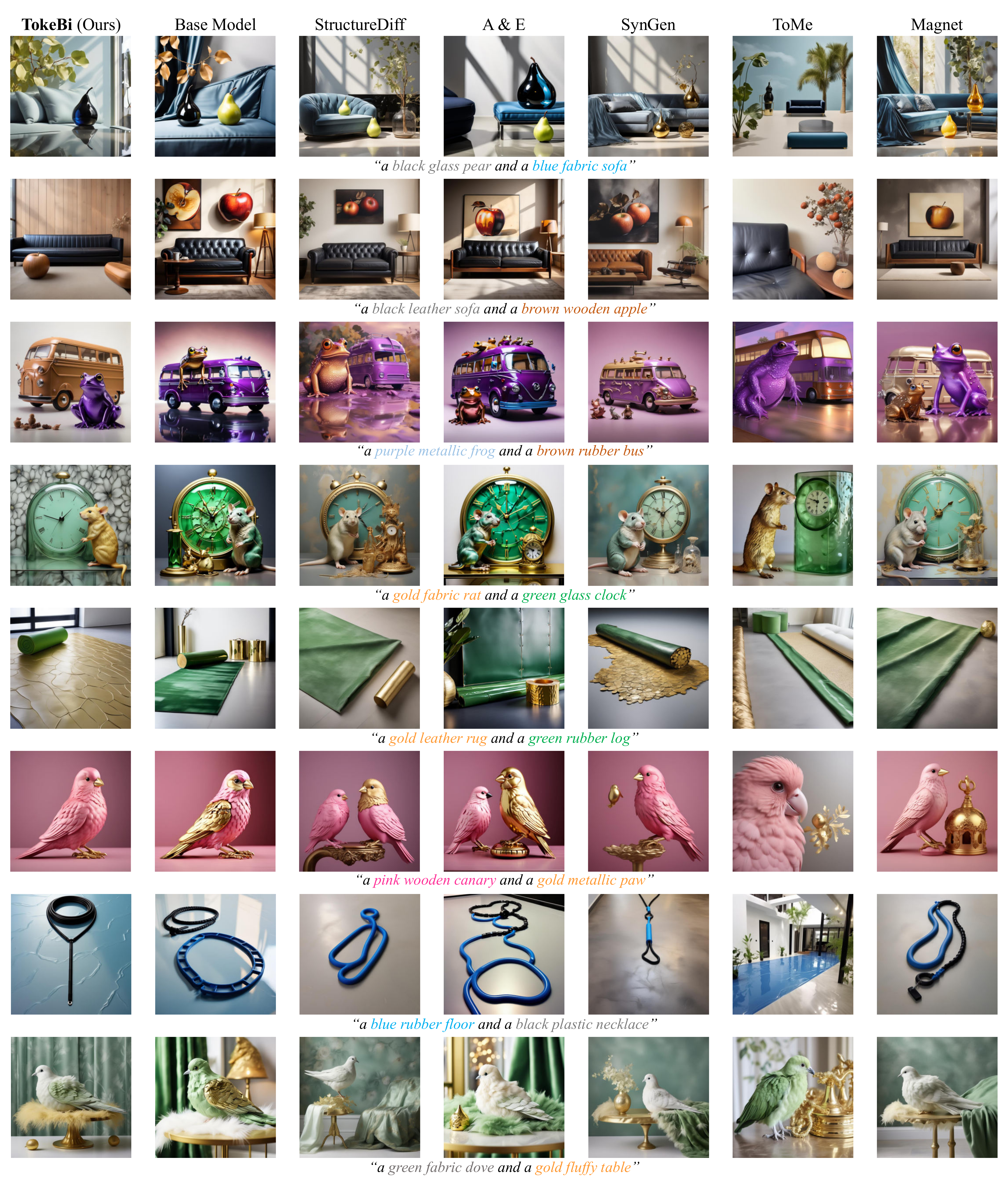}}
\vspace{-0.5em}
\caption{\textbf{Qualitative comparison of TokeBi against other baseline methods using the PlayGround-v2 as a base model on our $3\times 3$ dataset.} The prompts used in this comparison consist of relatively simple cases, each containing two objects with a single descriptive attribute per object, covering diverse examples across color, texture, and shape. This comparison confirms that TokeBi achieves superior semantic binding performance, not only quantitatively but also qualitatively.}
\label{sup_fig:quali_4}
\vspace{-1.0em}
\end{center}
\end{figure}

\begin{figure}[!t]
\begin{center}
\centerline{\includegraphics[width=\linewidth]{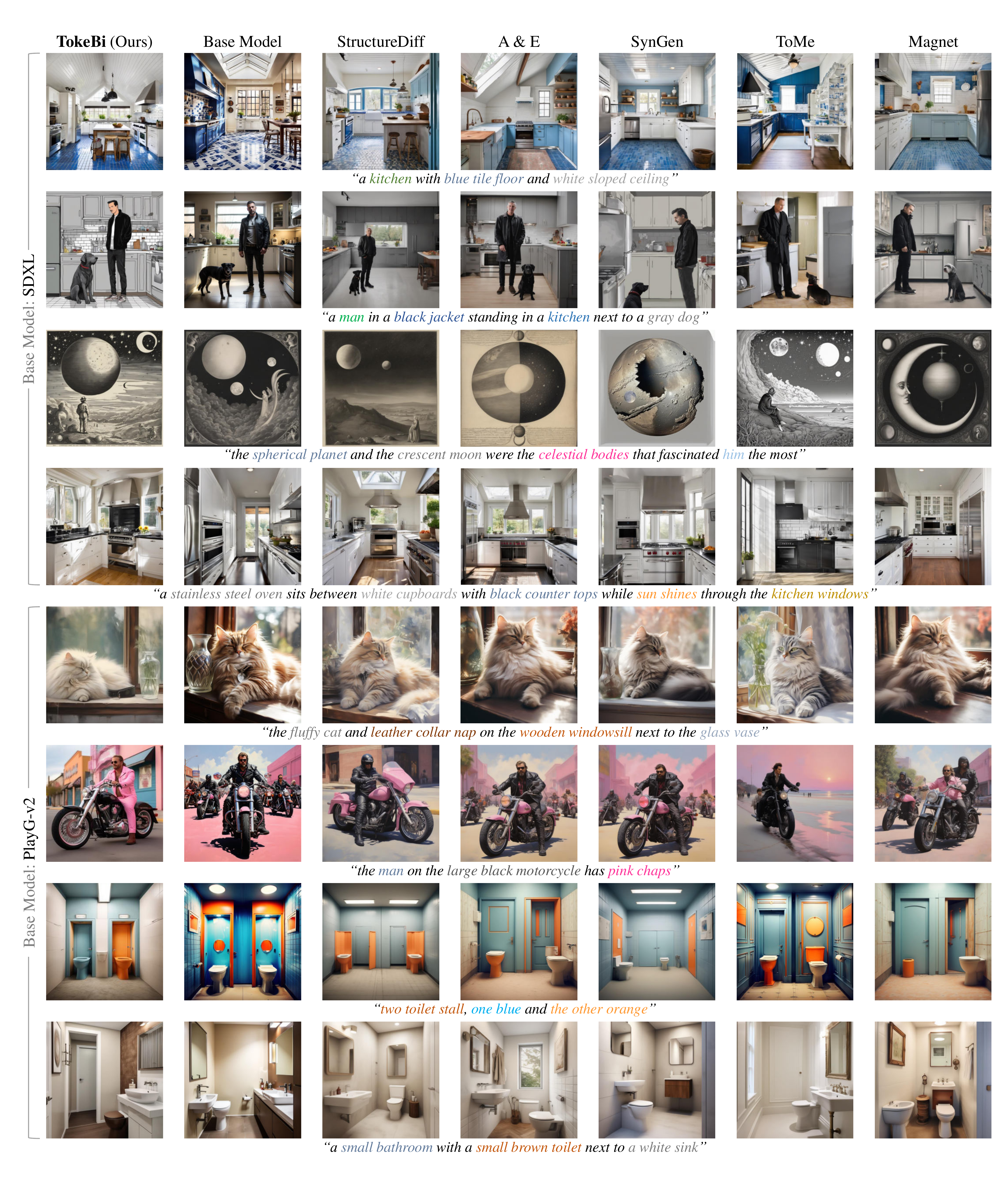}}
\vspace{-0.5em}
\caption{\textbf{Qualitative comparison of TokeBi against other baseline methods using the SDXL and PlayGround-v2 as a base models on the T2I-CompBench dataset.} The prompts used in this comparison consist of relatively complex cases, containing multiple objects, attributes and environment, covering diverse examples across color, texture, and shape. This comparison confirms that TokeBi achieves superior semantic binding performance, not only quantitatively but also qualitatively.}
\label{sup_fig:quali_5}
\vspace{-1.0em}
\end{center}
\end{figure}

\begin{figure}[!t]
\begin{center}
\centerline{\includegraphics[width=\linewidth]{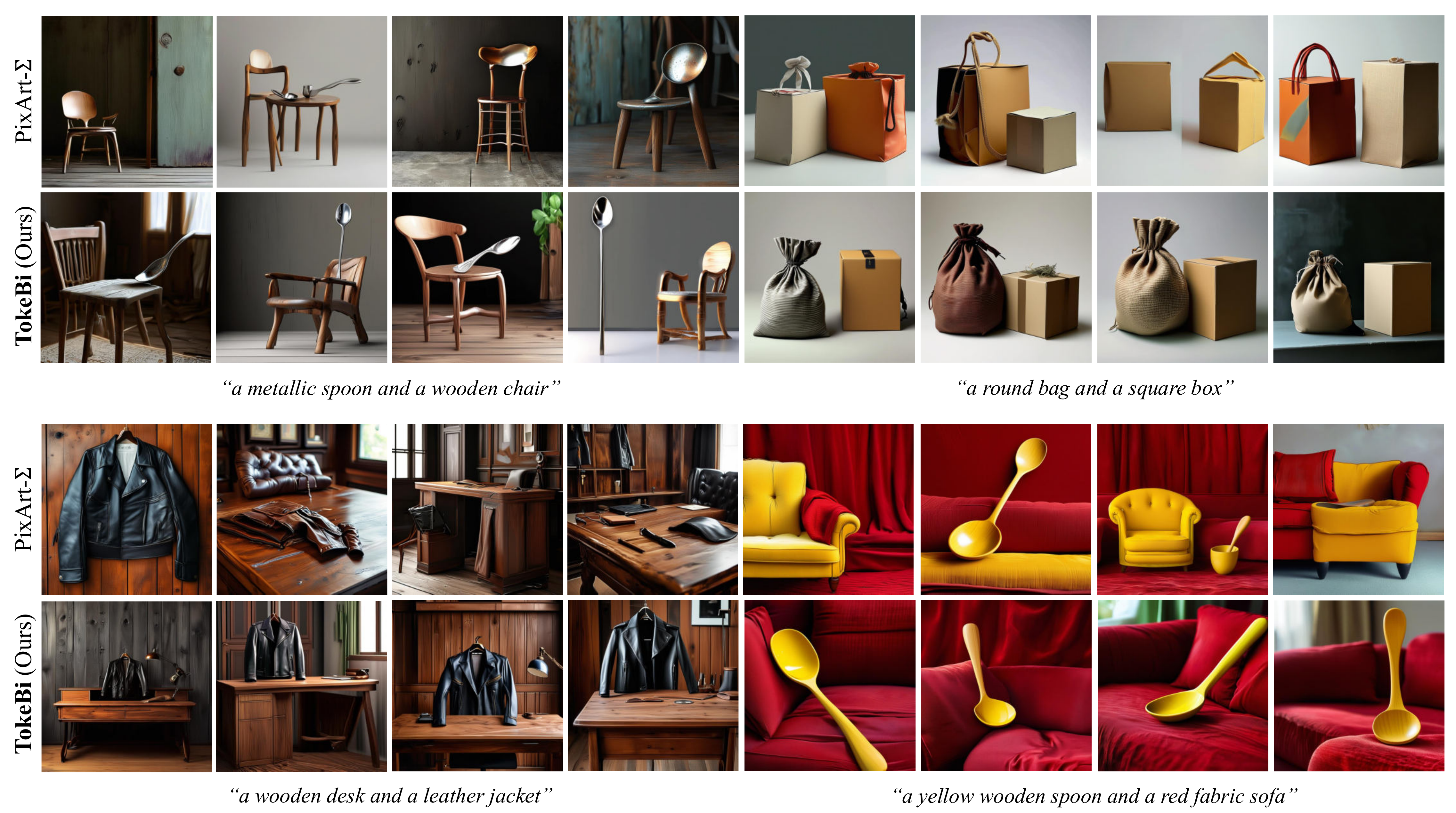}}
\caption{\textbf{Qualitative results of applying TokeBi using PixArt-$\mathbf{\Sigma}$ as the base model.} TokeBi demonstrates strong performance not only on simple prompts but also on challenging prompts such as the $3\times 3$ configuration. These results show that TokeBi performs effectively even with PixArt-$\Sigma$, which employs a non-causal text encoder and the DiT architecture.}
\label{sup_fig:pixart}
\end{center}
\end{figure}

\begin{figure}[!t]
\begin{center}
\centerline{\includegraphics[width=\linewidth]{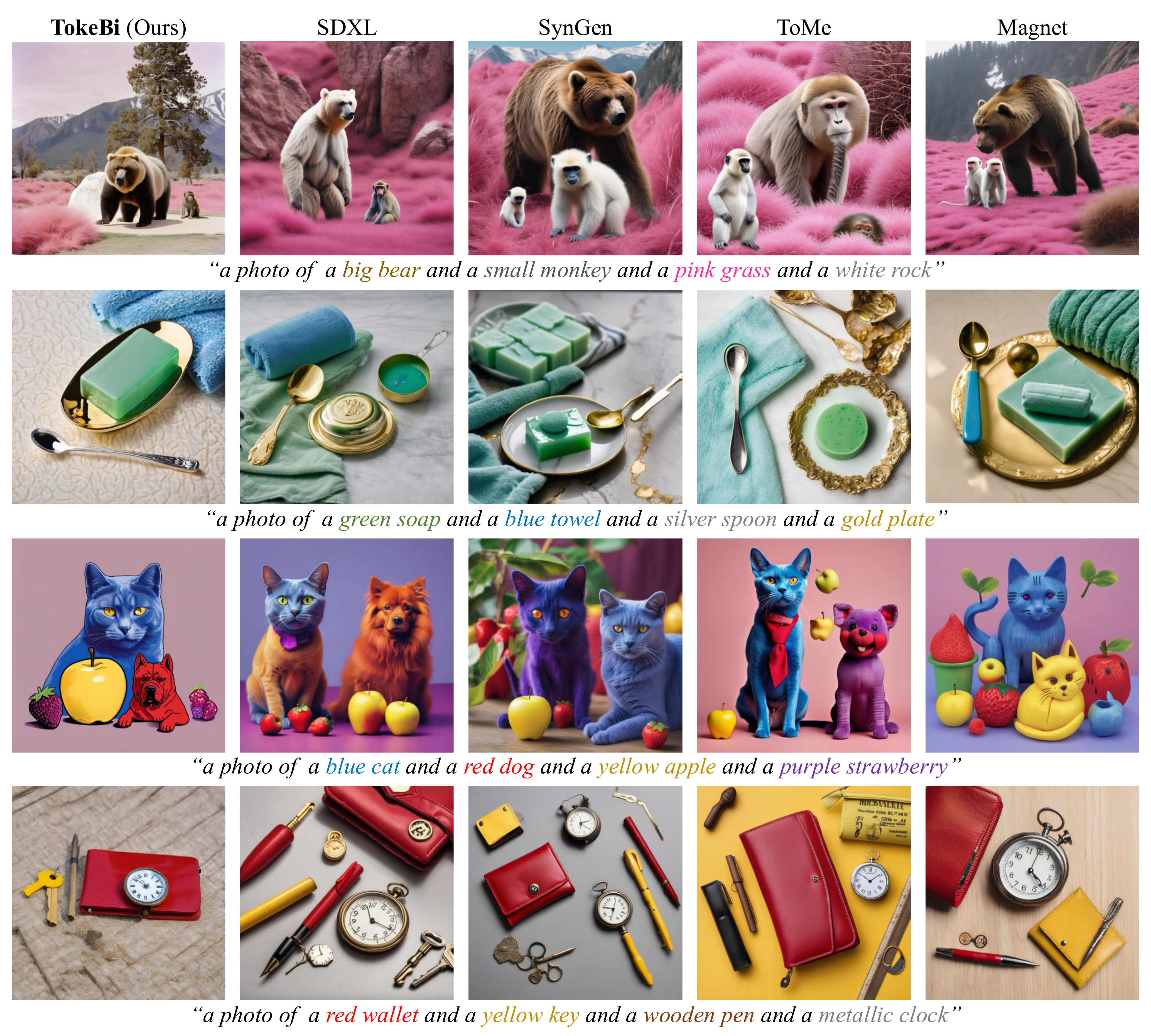}}
\caption{\textbf{Qualitative comparisons TokeBi against other baseline methods using the SDXL on highly complex prompts involving four distinct objects, each described by individual attributes.} Despite the complexity arising from multiple objects and diverse attributes, TokeBi consistently achieves excellent semantic binding performance, accurately reflecting each object and its associated attributes in the generated images.}
\label{sup_fig:2222}
\end{center}
\end{figure}

We provide additional qualitative comparisons on both the T2I CompBench dataset and our proposed $3\times 3$ dataset. These experiments were conducted under the same settings as in the main manuscript for SDXL and PlayG-v2, with results illustrated in the Fig.~\ref{sup_fig:quali_1}, Fig.~\ref{sup_fig:quali_2}, Fig.~\ref{sup_fig:quali_3}, Fig.~\ref{sup_fig:quali_4}, and Fig.~\ref{sup_fig:quali_5}. The presented outcomes demonstrate that TokeBi achieves superior performance compared to various alternative methods. Additionally, we provide further qualitative results for PixArt-$\Sigma$ with a broader set of prompts, as shown in the Fig.~\ref{sup_fig:pixart}. These results clearly indicate that TokeBi robustly operates across diverse denoising model architectures (U-Net vs. DiT) and varying types of text encoder causality.
Additionally, we conducted experiments using prompts containing four objects, each with their own attributes, and these results are presented in Fig.~\ref{sup_fig:2222}. Note that ours is the first work to report results for four objects without spatial guidance. While existing methods typically experience decreased semantic binding performance as the number of objects increases, TokeBi demonstrates excellent semantic binding capability even under these challenging conditions.

\subsection{Uncurated Samples}
\label{sup_subsec:uncurated}
\begin{figure}[!t]
\begin{center}
\centerline{\includegraphics[width=0.6\linewidth]{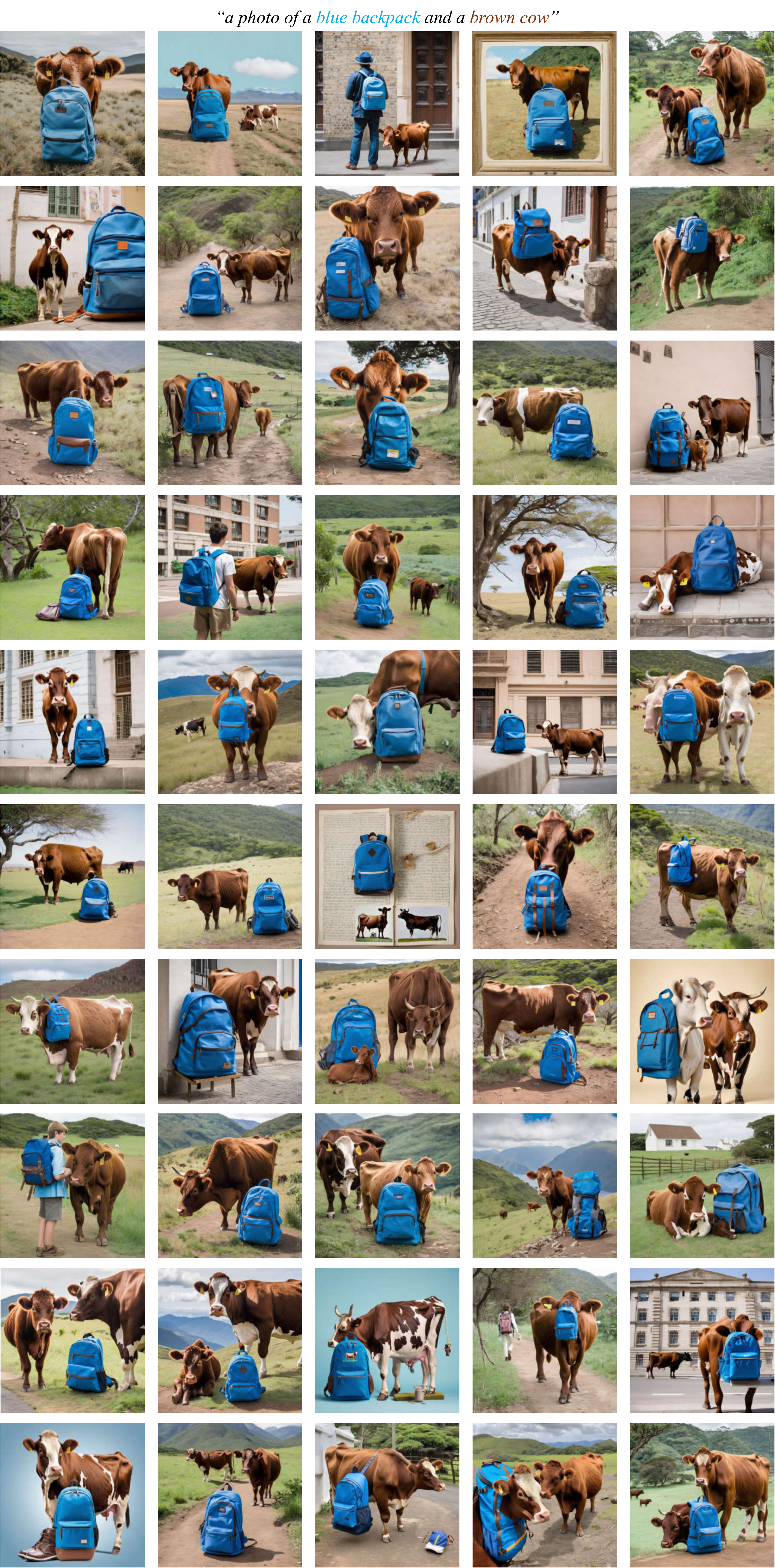}}
\caption{\textbf{50 random, uncurated images generated by TokeBi.} The images were generated from the input prompt \textit{``a photo of a blue backpack and a brown cow''}. Note that we generated 50 images regardless of their quality and directly report the results. These results demonstrate that TokeBi robustly performs semantic binding across diverse generations.}
\label{sup_fig:uncurated_1}
\end{center}
\end{figure}

\begin{figure}[!t]
\begin{center}
\centerline{\includegraphics[width=0.6\linewidth]{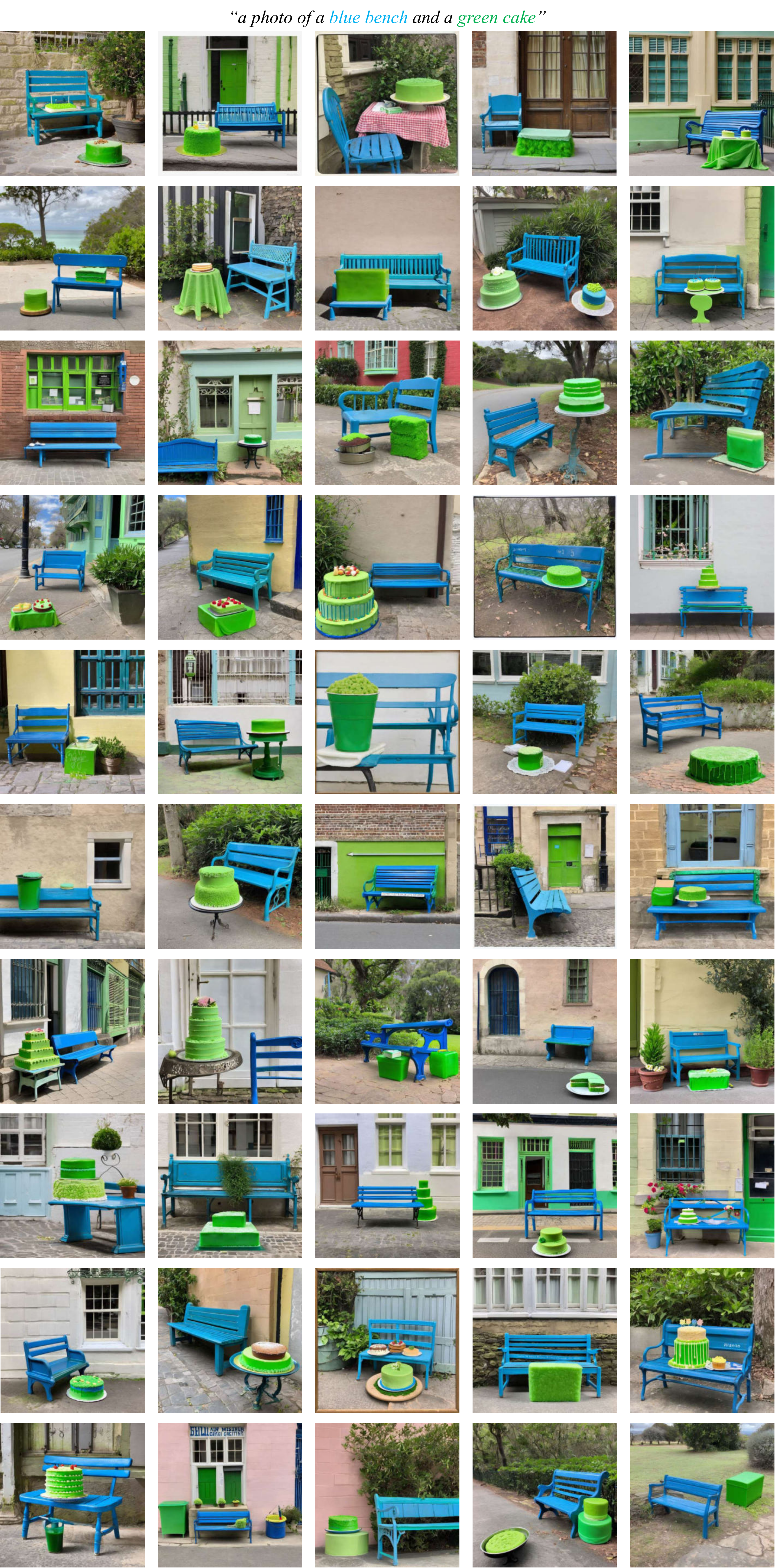}}
\caption{\textbf{50 random, uncurated images generated by TokeBi.} The images were generated from the input prompt \textit{``a photo of a blue bench and a green cake''}. Note that we generated 50 images regardless of their quality and directly report the results. These results demonstrate that TokeBi robustly performs semantic binding across diverse generations.}
\label{sup_fig:uncurated_2}
\end{center}
\end{figure}

\begin{figure}[!t]
\begin{center}
\centerline{\includegraphics[width=0.6\linewidth]{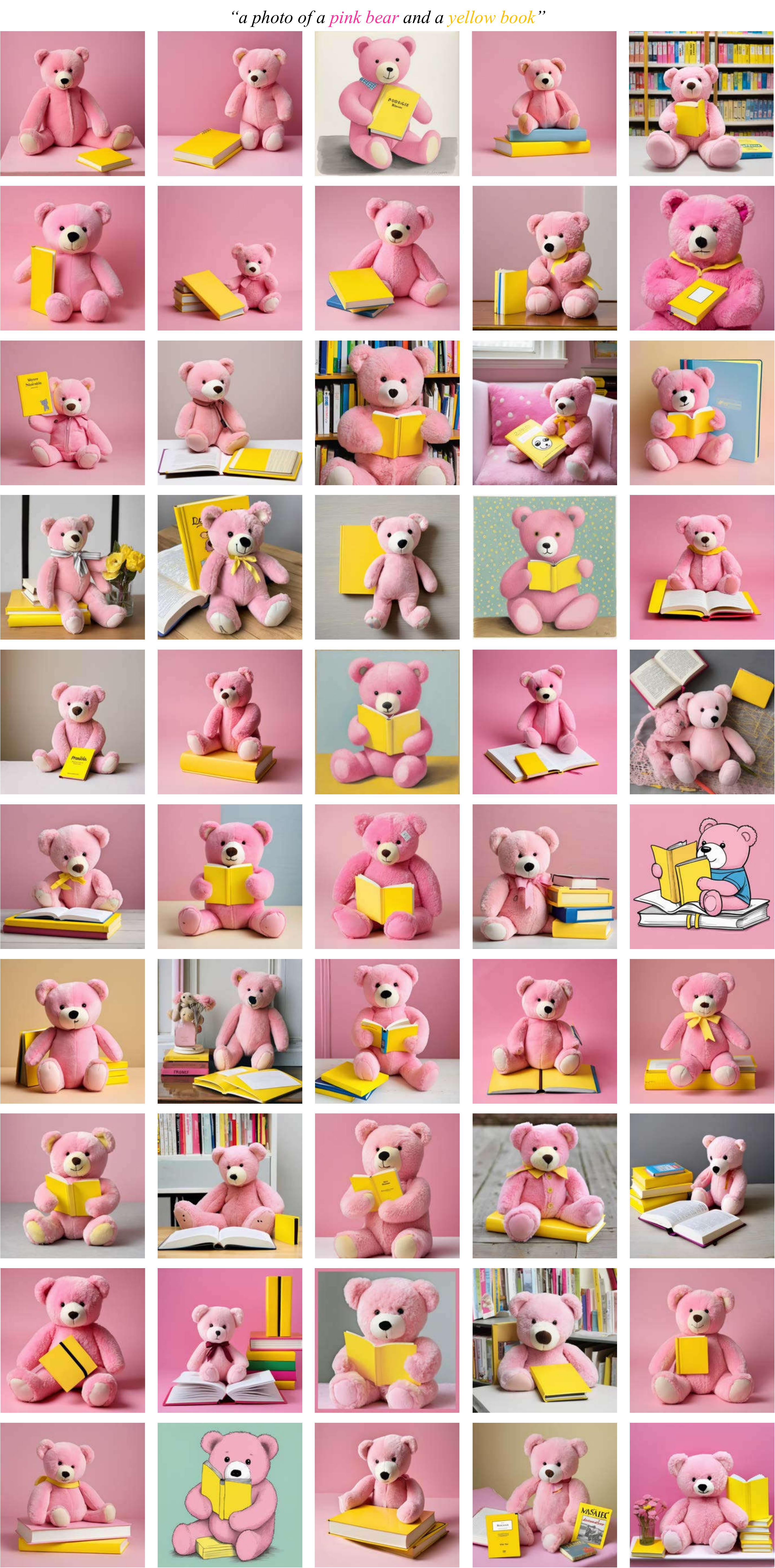}}
\caption{\textbf{50 random, uncurated images generated by TokeBi.} The images were generated from the input prompt \textit{``a photo of a pink bear and a yellow book''}. Note that we generated 50 images regardless of their quality and directly report the results. These results demonstrate that TokeBi robustly performs semantic binding across diverse generations.}
\label{sup_fig:uncurated_3}
\end{center}
\end{figure}

\begin{figure}[!t]
\begin{center}
\centerline{\includegraphics[width=0.6\linewidth]{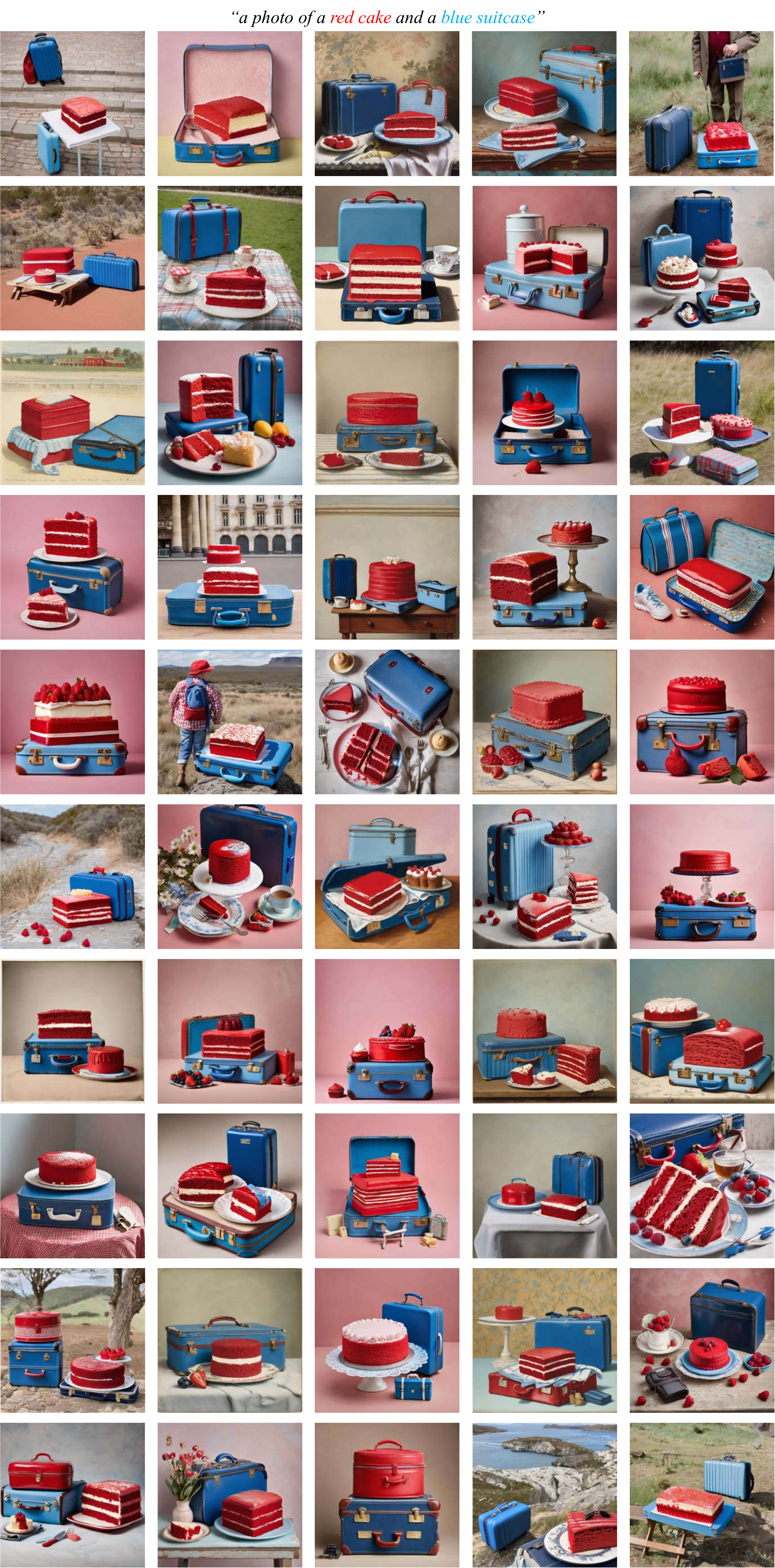}}
\caption{\textbf{50 random, uncurated images generated by TokeBi.} The images were generated from the input prompt \textit{``a photo of a red cake and a blue suitcase''}. Note that we generated 50 images regardless of their quality and directly report the results. These results demonstrate that TokeBi robustly performs semantic binding across diverse generations.}
\label{sup_fig:uncurated_4}
\end{center}
\end{figure}

\begin{figure}[!t]
\begin{center}
\centerline{\includegraphics[width=0.6\linewidth]{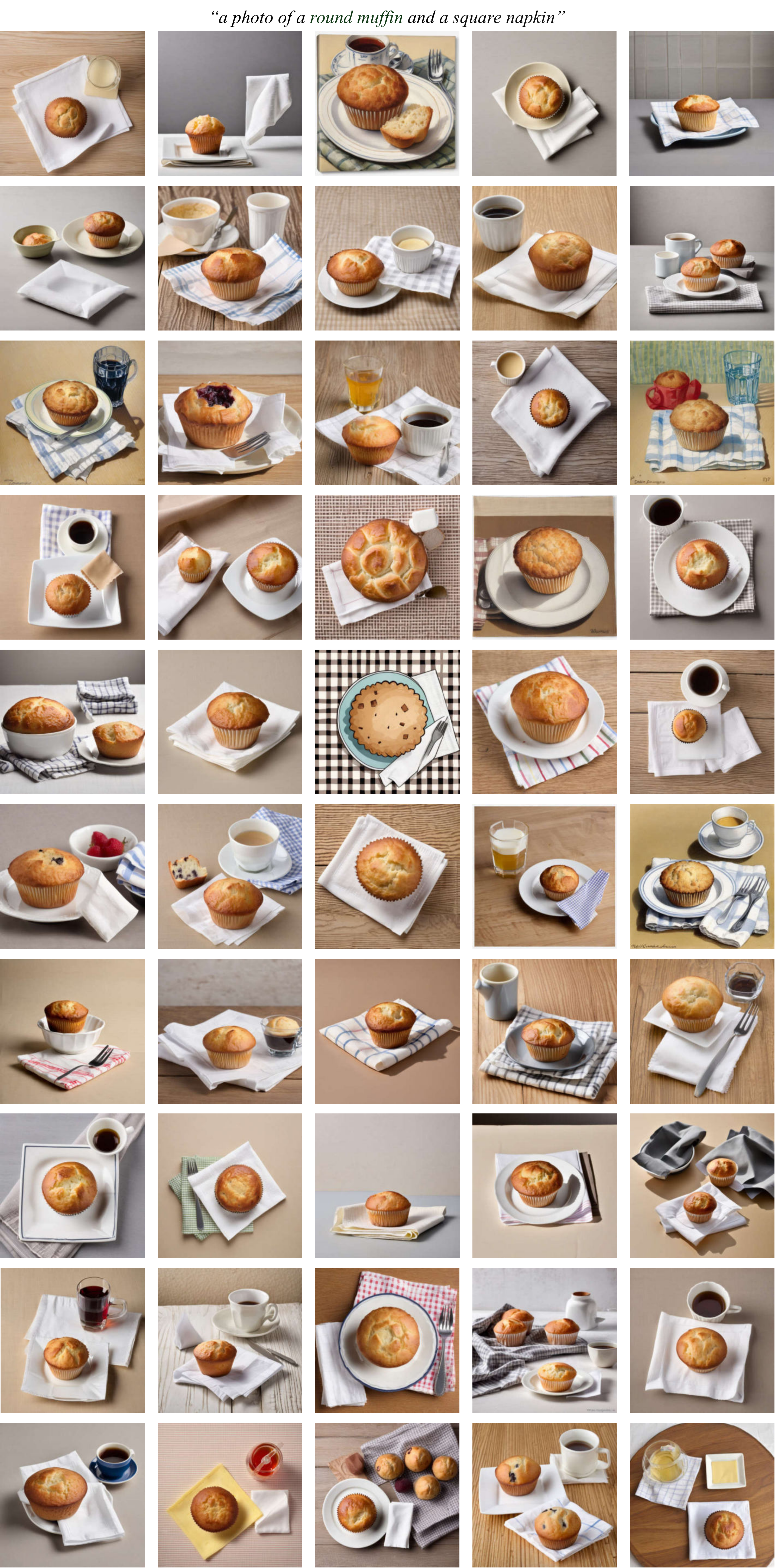}}
\caption{\textbf{50 random, uncurated images generated by TokeBi.} The images were generated from the input prompt ``a photo of a round muffin and a square napkin''. Note that we generated 50 images regardless of their quality and directly report the results. These results demonstrate that TokeBi robustly performs semantic binding across diverse generations.}
\label{sup_fig:uncurated_5}
\end{center}
\end{figure}

\begin{figure}[!t]
\begin{center}
\centerline{\includegraphics[width=0.6\linewidth]{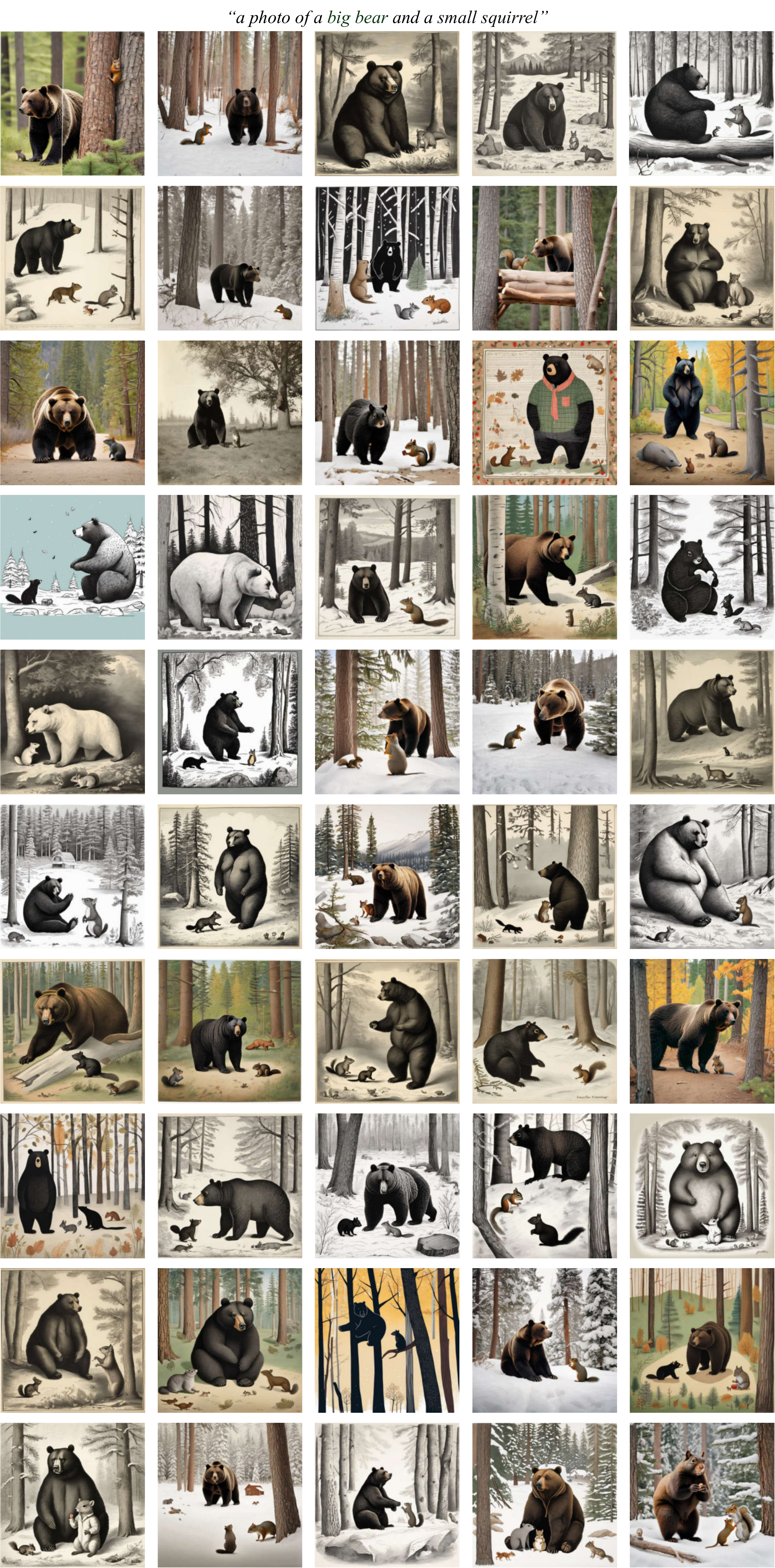}}
\caption{\textbf{50 random, uncurated images generated by TokeBi.} The images were generated from the input prompt \textit{``a photo of a big bear and a small squirrel''}. Note that we generated 50 images regardless of their quality and directly report the results. These results demonstrate that TokeBi robustly performs semantic binding across diverse generations.}
\label{sup_fig:uncurated_6}
\end{center}
\end{figure}

\begin{figure}[!t]
\begin{center}
\centerline{\includegraphics[width=0.6\linewidth]{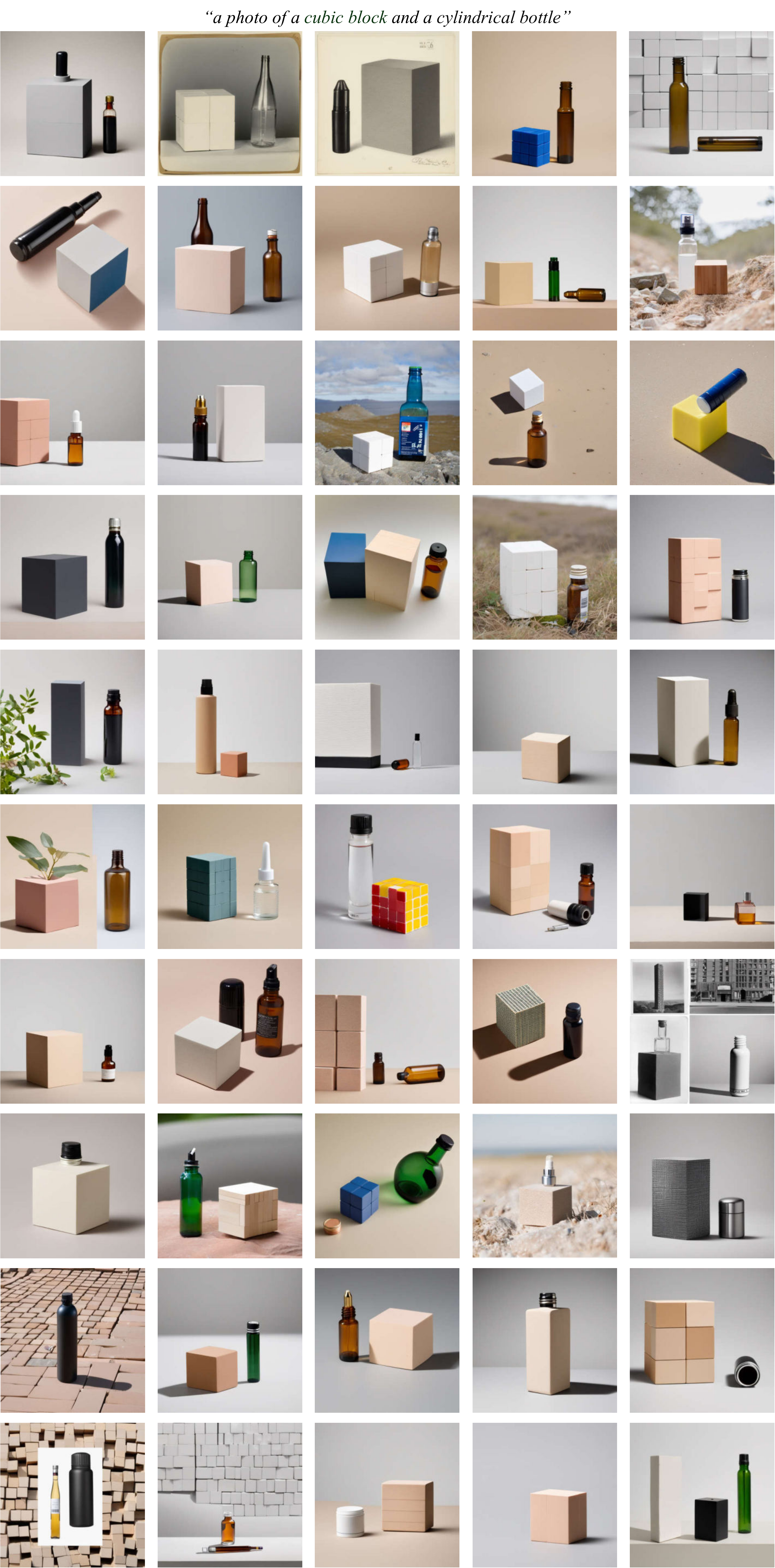}}
\caption{\textbf{50 random, uncurated images generated by TokeBi.} The images were generated from the input prompt \textit{``a photo of a cubic block and a cylindrical bottle''}. Note that we generated 50 images regardless of their quality and directly report the results. These results demonstrate that TokeBi robustly performs semantic binding across diverse generations.}
\label{sup_fig:uncurated_7}
\end{center}
\end{figure}

\begin{figure}[!t]
\begin{center}
\centerline{\includegraphics[width=0.6\linewidth]{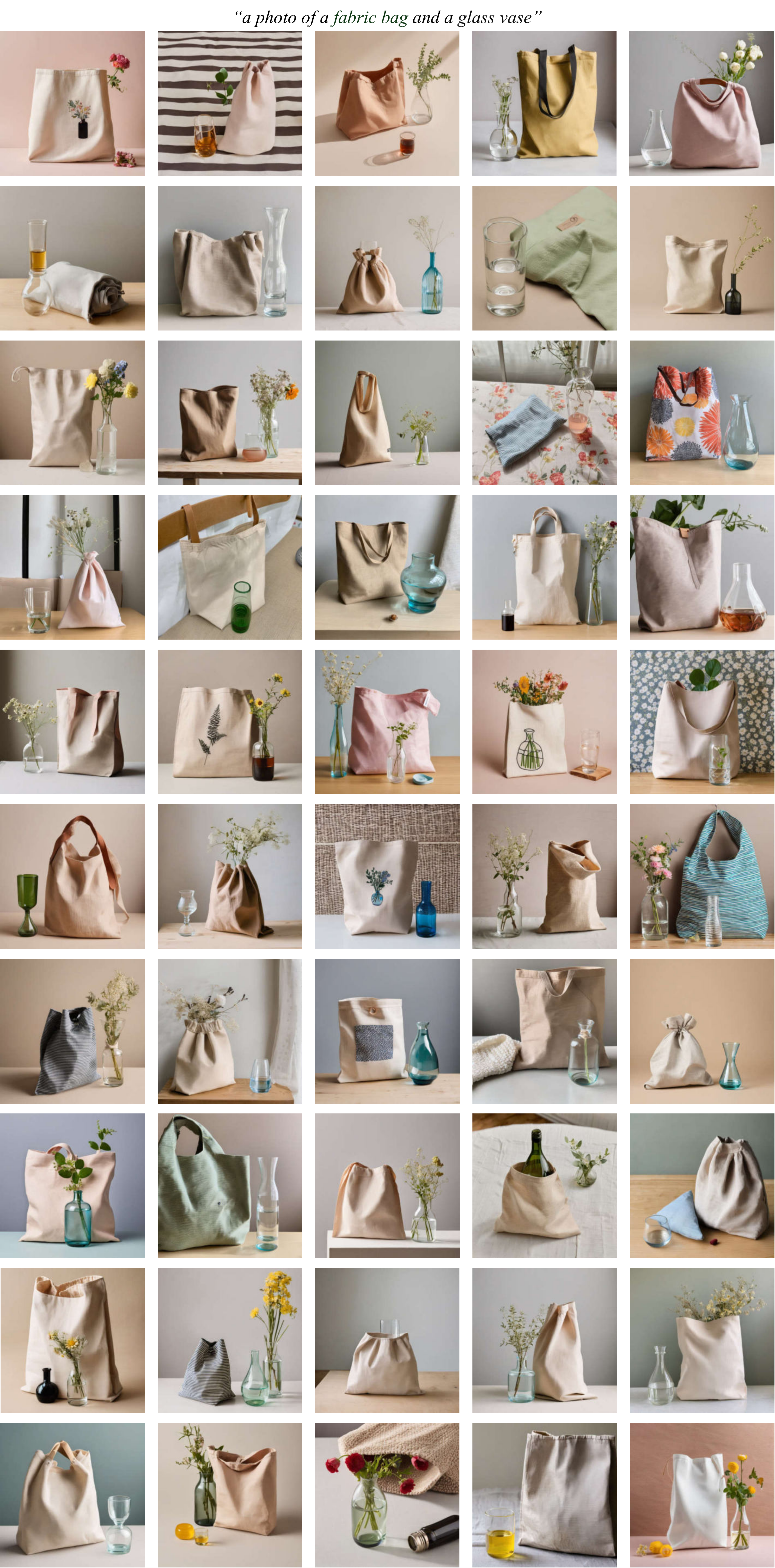}}
\caption{\textbf{50 random, uncurated images generated by TokeBi.} The images were generated from the input prompt \textit{``a photo of a fabric bag and a glass vase''}. Note that we generated 50 images regardless of their quality and directly report the results. These results demonstrate that TokeBi robustly performs semantic binding across diverse generations.}
\label{sup_fig:uncurated_8}
\end{center}
\end{figure}

\begin{figure}[!t]
\begin{center}
\centerline{\includegraphics[width=0.6\linewidth]{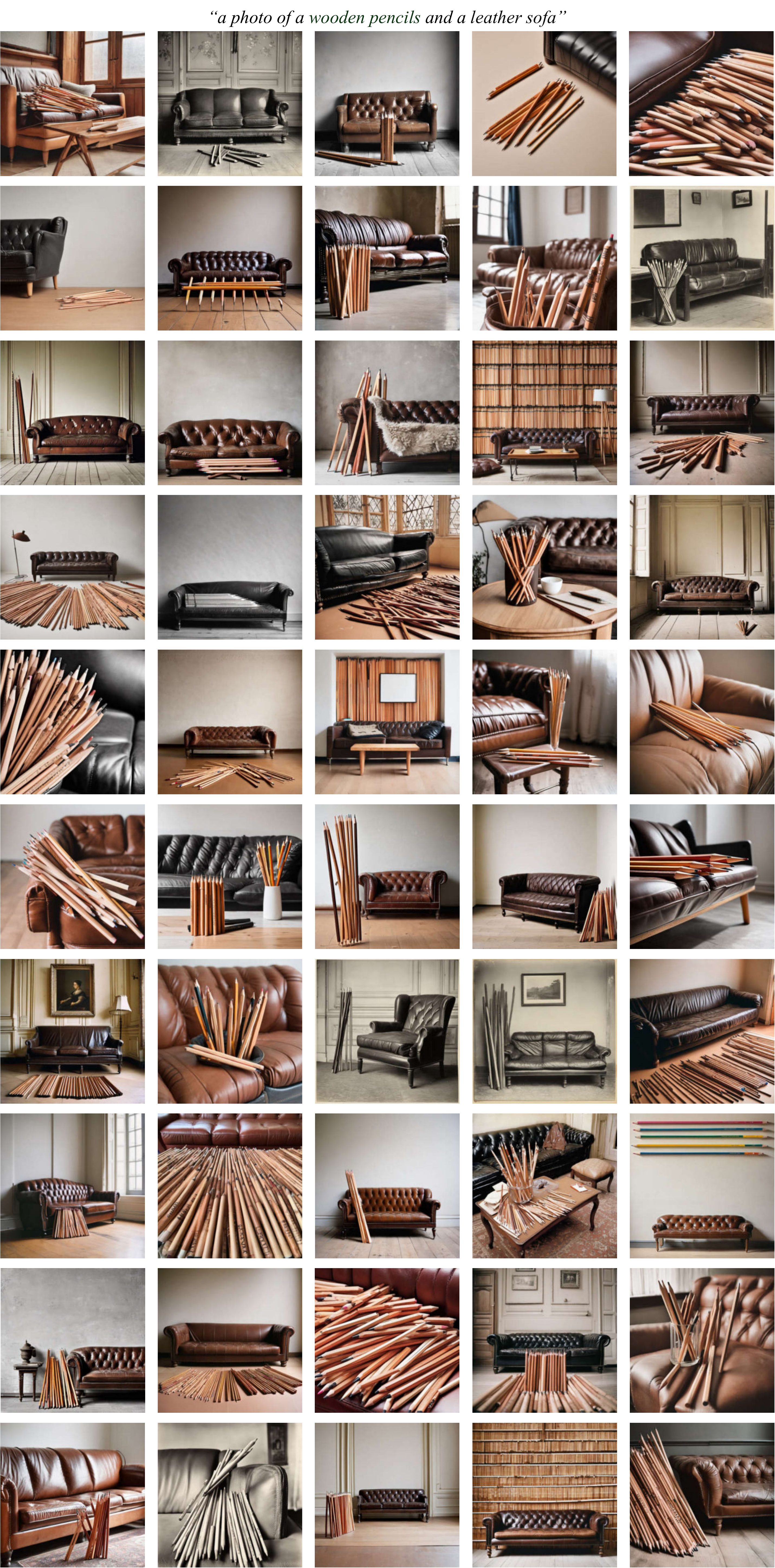}}
\caption{\textbf{50 random, uncurated images generated by TokeBi.} The images were generated from the input prompt \textit{``a photo of wooden pencils and a leather sofa''}. Note that we generated 50 images regardless of their quality and directly report the results. These results demonstrate that TokeBi robustly performs semantic binding across diverse generations.}
\label{sup_fig:uncurated_9}
\end{center}
\end{figure}

\begin{figure}[!t]
\begin{center}
\centerline{\includegraphics[width=0.6\linewidth]{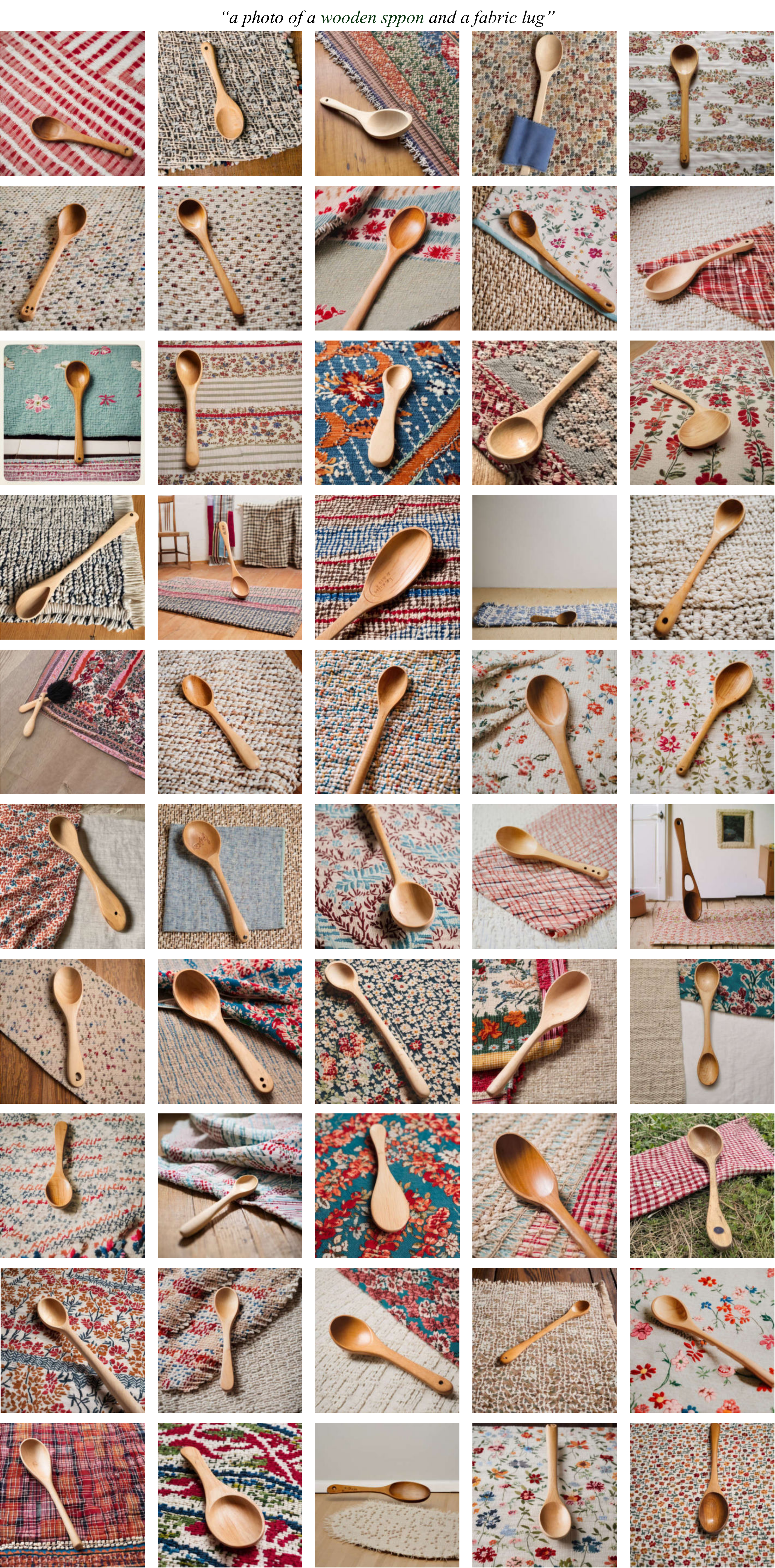}}
\caption{\textbf{50 random, uncurated images generated by TokeBi.} The images were generated from the input prompt \textit{``a photo of a wooden spoon and a fabric rug''}. Note that we generated 50 images regardless of their quality and directly report the results. These results demonstrate that TokeBi robustly performs semantic binding across diverse generations.}
\label{sup_fig:uncurated_10}
\end{center}
\end{figure}

\begin{figure}[!t]
\begin{center}
\centerline{\includegraphics[width=0.6\linewidth]{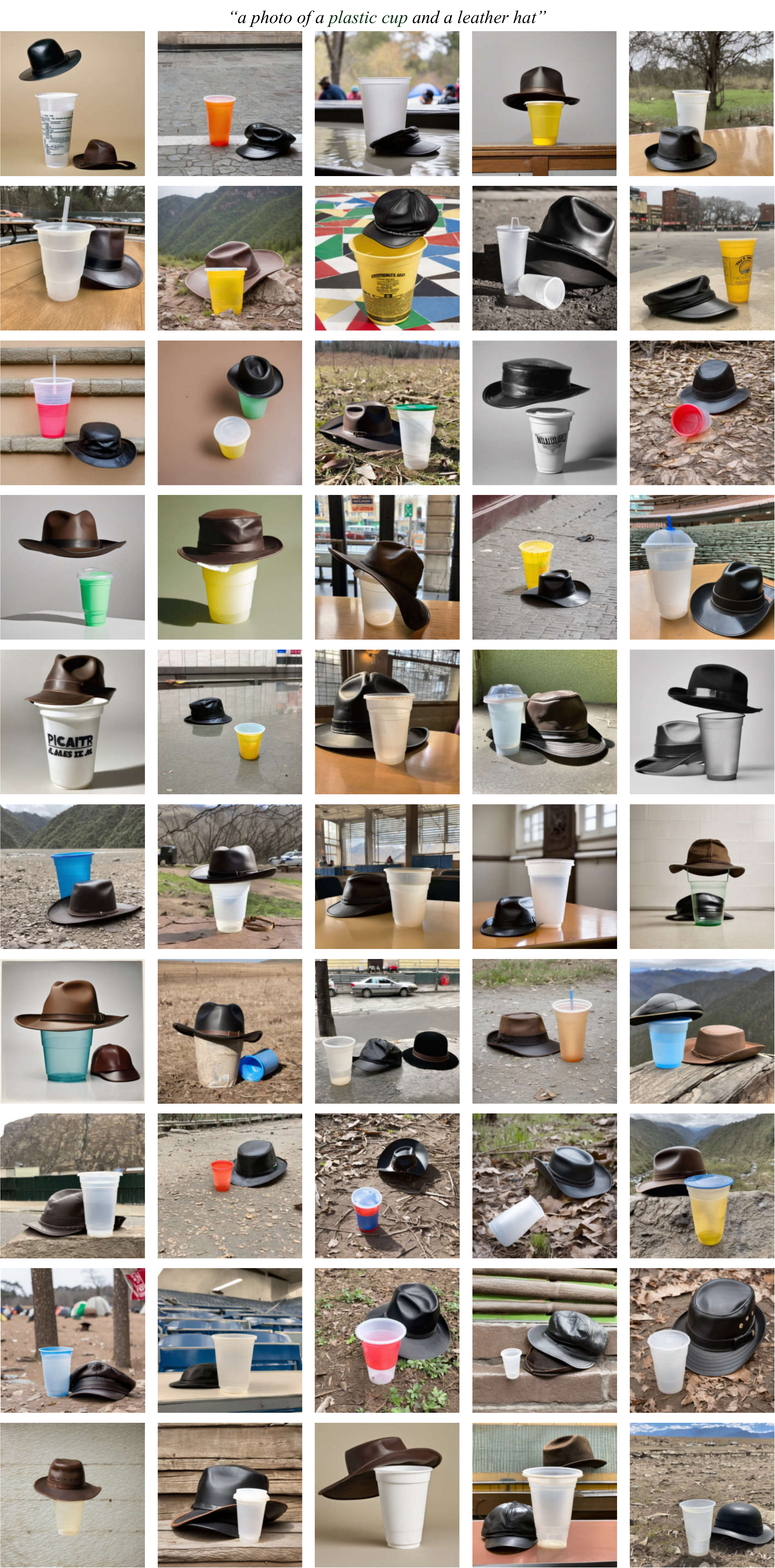}}
\caption{\textbf{50 random, uncurated images generated by TokeBi.} The images were generated from the input prompt \textit{``a photo of a plastic cup and a leather hat''}. Note that we generated 50 images regardless of their quality and directly report the results. These results demonstrate that TokeBi robustly performs semantic binding across diverse generations.}
\label{sup_fig:uncurated_11}
\end{center}
\end{figure}

In T2I synthesis, the random seed significantly affects not only the quality of generated images but also several critical aspects of semantic binding performance. Although we have quantitatively verified the robustness of TokeBi through evaluations involving thousands of images, we further provide qualitative evidence of TokeBi's robustness by presenting uncurated samples. Specifically, for each of the 11 prompts, we performed 50 purely random image generations without selecting any manual seed sets, and the resulting images are presented in the Fig.~\ref{sup_fig:uncurated_1}, Fig.~\ref{sup_fig:uncurated_2}, Fig.~\ref{sup_fig:uncurated_3}, Fig.~\ref{sup_fig:uncurated_4}, Fig.~\ref{sup_fig:uncurated_5}, Fig.~\ref{sup_fig:uncurated_6}, Fig.~\ref{sup_fig:uncurated_7}, Fig.~\ref{sup_fig:uncurated_8}, Fig.~\ref{sup_fig:uncurated_9}, Fig.~\ref{sup_fig:uncurated_10}, and Fig.~\ref{sup_fig:uncurated_11}. These results clearly demonstrate that TokeBi achieves robust semantic binding across diverse input prompts.

\end{document}